\documentclass[final,1p,times,authoryear]{elsarticle}

\usepackage{times} 
\usepackage{amsmath} 
\usepackage{amsfonts}
\usepackage{amssymb}  
\usepackage{bm}
\usepackage{multirow}
\usepackage{float}
\usepackage{graphicx}
\usepackage{subfig}
\usepackage{caption}
\usepackage[dvipsnames]{xcolor}
\usepackage{booktabs}
\usepackage[ruled]{algorithm2e}

\usepackage{xparse}%
\usepackage{blindtext}
\usepackage{morewrites}
\usepackage{wrapfig}
\usepackage{xkeyval}%

\newwrite\authorbibfile%

\AtBeginDocument{%
  \immediate\openout\authorbibfile=\jobname.aub%
}%
\AtEndDocument{%
\immediate\closeout\authorbibfile
\InputIfFileExists{\jobname.aub}{}{}
}%

\makeatletter

\define@key{authorbib}{scale}[1]{%
\def\AuthorbibKVMacroScale{#1}%
}

\define@key{authorbib}{wraplines}[10]{%
\def\AuthorbibKVMacroWraplines{#1}%
}

\define@key{authorbib}{imagewidth}[4cm]{%
\def\AuthorbibKVMacroImagewidth{#1}%
}
\define@key{authorbib}{overhang}[10pt]{%
\def\AuthorbibKVMacroOverhang{#1}%
}

\define@key{authorbib}{imagepos}[l]{%
\def\AuthorbibKVMacroImagepos{#1}%
}

\makeatother

\presetkeys{authorbib}{imagepos=l,imagewidth=4cm,wraplines=15,overhang=20pt}{}

\newlength{\AuthorbibTopSkip}
\newlength{\AuthorbibBottomSkip}
\NewDocumentCommand{\authorbibliography}{+o+m+m+m}{%
  \IfNoValueTF{#1}{%
  }{%
    \setkeys{authorbib}{#1}%
    \immediate\write\authorbibfile{%
      \string\begin{wrapfigure}[\AuthorbibKVMacroWraplines]{\AuthorbibKVMacroImagepos}[\AuthorbibKVMacroOverhang]{\AuthorbibKVMacroImagewidth}^^J
        \string\includegraphics[scale=\AuthorbibKVMacroScale]{#2}^^J
        \string\end{wrapfigure}^^J
    }%
  }%
  \IfNoValueTF{#3}{%
    \typeout{Warning: No author name}%
  }{%
      \immediate\write\authorbibfile{%
      \unexpanded{\vspace{\AuthorbibTopSkip}}^^J
      \string\noindent\relax
      \unexpanded{\textbf{#3}\par}^^J
      \string\noindent\relax
      \unexpanded{#4}^^J%
      \unexpanded{\vspace{\AuthorbibBottomSkip}}^^J
      }%
  }%
}%

\journal{XXXX}

\biboptions{authoryear}

\begin{document}

\begin{frontmatter}

\title{InTEn-LOAM: Intensity and Temporal Enhanced LiDAR Odometry and Mapping}

\author[label1]{Shuaixin Li}

\author[label2]{Bin Tian\corref{cofir}}
\cortext[cofir]{Co-first Authors}

\author[label1]{Zhu Xiaozhou}

\author[label1]{Gui Jianjun}

\author[label1]{Yao Wen}

\author[label3]{Guangyun Li\corref{cor}}
\cortext[cor]{Corresponding author:}
\ead{guangyun\_li@163.com}

\address[label1]{the National Innovation Institute of Defence Technology, PLA Academy of Military Sciences, Beijing 100079, China}
\address[label2]{the State Key Laboratory of Management and Control for Complex Systems, Institute of Automation, Chinese Academy of Science, Beijing 100081, China}
\address[label3]{the Department of Geospatial Information, PLA Information Engineering University, Zhengzhou 450001, China}

\begin{abstract}
Traditional LiDAR odometry (LO) systems mainly leverage geometric information obtained from the traversed surroundings to register laser scans and estimate LiDAR ego-motion, while it may be unreliable in dynamic or unstructured environments. This paper proposes InTEn-LOAM, a low-drift and robust LiDAR odometry and mapping method that fully exploits implicit information of laser sweeps (i.e., geometric, intensity, and temporal characteristics). Scanned points are projected to cylindrical images, which facilitate the efficient and adaptive extraction of various types of features, i.e., ground, beam, facade, and reflector. We propose a novel intensity-based points registration algorithm and incorporate it into the LiDAR odometry, enabling the LO system to jointly estimate the LiDAR ego-motion using both geometric and intensity feature points. To eliminate the interference of dynamic objects, we propose a temporal-based dynamic object removal approach to filter them out before map update. Moreover, the local map is organized and downsampled using a temporal-related voxel grid filter to maintain the similarity between the current scan and the static local map. Extensive experiments are conducted on both simulated and real-world datasets. The results show that the proposed method achieves similar or better accuracy w.r.t the state-of-the-arts in normal driving scenarios and outperforms geometric-based LO in unstructured environments.
\end{abstract}


\end{frontmatter}

\section{INTRODUCTION} \label{Sect:1}
Autonomous robots and self-driving vehicles must have the ability to localize themselves and intelligently perceive the external surroundings. Simultaneous localization and mapping (SLAM) focuses on the issue of vehicle localization and navigation in unknown environments, which plays a major role in many autonomous driving and robotics-related applications, such as mobile mapping\cite{li2020slam}, space exploration\cite{ebadi2020lamp}, robot localization\cite{filipenko2018comparison}, and high-definition map production\cite{yang2018robust}. In accordance with the on-board perceptional sensors, it can be roughly classified into two categories, i.e., camera-based and LiDAR-based SLAM. Compared with images, LiDAR (Light detection and ranging) point clouds are invariant to the changing illumination and sufficiently dense for 3D reconstruction tasks. Accordingly, LiDAR SLAM solutions have become a preferred choice for self-driving car manufacturers than vision-based solutions \cite{milz2018visual, campos2020orb, qin2018vins}. Note that methods with loop closure are often called 'SLAM solutions', while those without the module are called 'odometry solutions'. However, both of them owns abilities of self-localization in unknown scenes and mapping the traversed environments.For instance, though LOAM \cite{zhang2017low} and LeGO-LOAM \cite{shan2018lego} achieve low-drift and real-time pose estimation and mapping, only LeGO-LOAM can be referred as complete SLAM solution since it is a loop closure-enabled system. 
\begin{figure}[ht]
  \centering
  \includegraphics[width=0.7\textwidth] {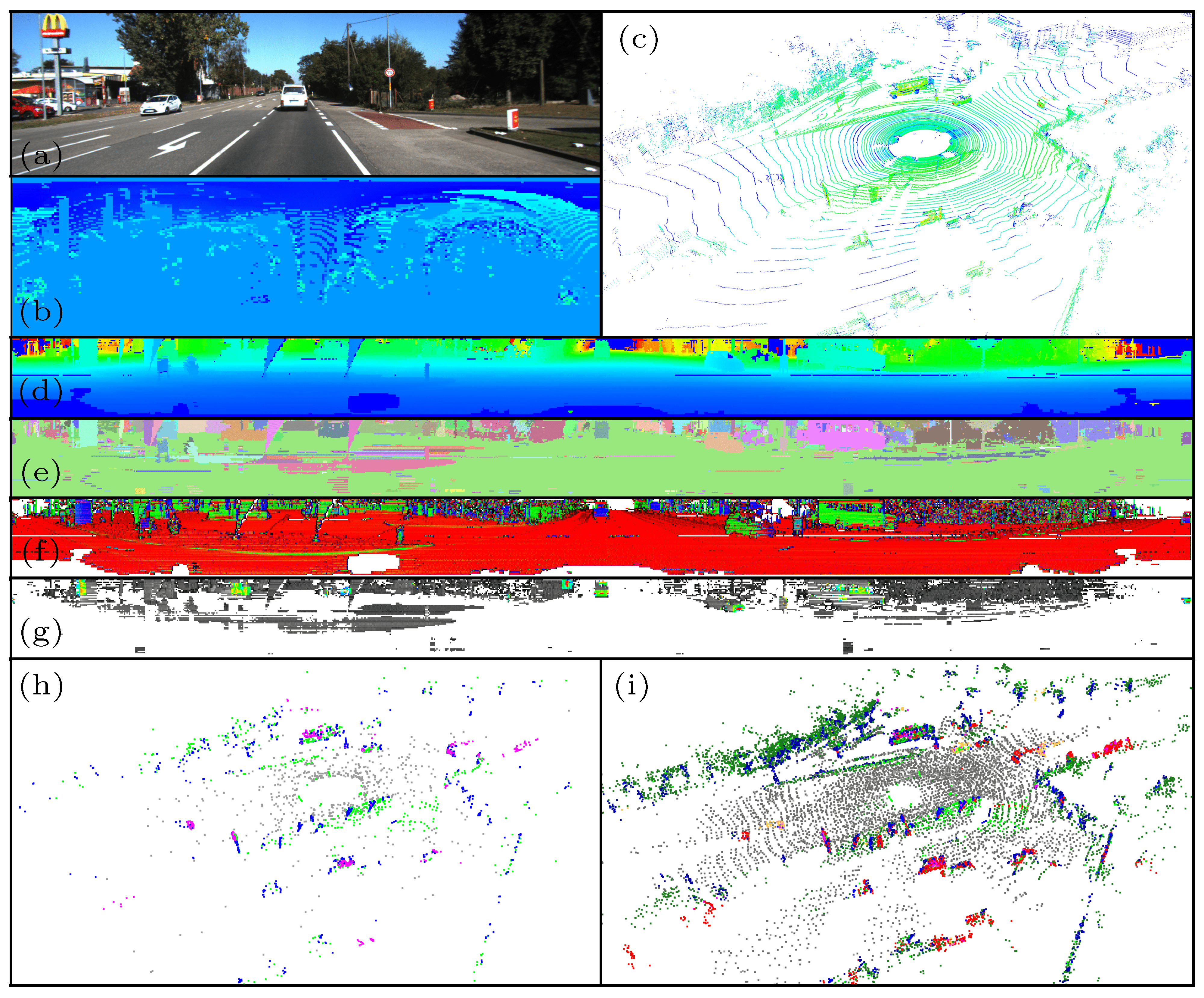}
  \caption[fontsize=6pt]{Overview of the proposed InTEn-LOAM system (a) The color image from the on-borad camera. (b) The projected scan-context segment image. (c) The raw point cloud from the Velodyne HDL-64 LiDAR colorized by intensity. (d) The projected cylindrical range image colorized by depth. 
  (e) The sgemented label image. (f) The estimated normal image (\textcolor{blue}{x}, \textcolor{green}{y}, \textcolor{red}{z}). (g) The intensity image. Only refletor features are colorized, while the ground is removed. (h) Various types of feature (\textcolor{Gray}{ground}, \textcolor{green}{facade}, \textcolor{blue}{edge}, \textcolor{Magenta}{reflector}) extracted from the laser scan. (i) The current point features aligned with the so-far local feature map(\textcolor{YellowOrange}{dynamic object}).
  } \label{FIG:1}
\end{figure}

It has witnessed remarkable progress in LiDAR-based SLAM for the past decade \cite{zhang2017low,behley2018efficient,shan2018lego,jiao2020robust,zhou2021s4,koide2019portable}. The state-of-the-art solutions have shown remarkable performances, especially in structured urban and indoor scenes. Recent years have seen solutions for more intractable problems, e.g., fusion with multiple sensors \cite{zhao2019robust,palieri2020locus,shan2020lio,qin2020lins,lin2021r2live}, adapting to cutting-edge solid-state LiDAR \cite{li2021towards}, global localization \cite{dube2020segmap}, improving the efficiency of optimization back-end \cite{droeschel2017continuous,ding2020lidar}, etc., yet many issues remain unsolved. Specifically, most conventional LO solutions currently ignore intensity information from the reflectance channel, though it reveals reflectivities of different objects in the real world. An efficient incorporation approach of making use of point intensity information is still an open problem since the intensity value is not as straightforward as the range value. It is a value w.r.t many factors, including the material of target surface, the scanning distance, the laser incidence angle, as well as the transmitted energy. Besides, the laser sweep represents a snapshot of surroundings, and thus moving objects, such as pedestrians, vehicles, etc., may be scanned. These dynamic objects result in 'ghosting points' in the accumulated points map and may increase the probability of incorrect matching, which deteriorates the localization accuracy of LO. Moreover, improving the robustness of point registration in some geometric-degraded environments, e.g., long straight tunnel, is also a topic worthy of in-depth discussion.    
In this paper, we present InTEn-LOAM (as shown in Fig.\ref{FIG:1}) to cope with the aforementioned challenges. 
The main contributions of our work are summarized as four-fold:
\begin{itemize}

  \item We propose an efficient range-image-based feature extraction 
  method that is able to adaptively extract features from the raw laser 
  scan and categorize them into four different types in real-time.
  
  \item We propose a coarse-to-fine, model-free method for online dynamic object 
  removal enabling the LO system to build a purely static map by removing all 
  dynamic outliers in raw scans. Besides, we improved the voxel-based downsize 
  filter, making use of the implicitly temporal information of consecutive laser 
  sweeps to ensure the similarity between the current scan and the local map.

  \item We propose a novel intensity-based points registration algorithm that 
  directly leverages reflectance measurements to align point clouds, and we 
  introduce it into the LO framework to achieve jointly pose estimation utilizing 
  both geometric and intensity information.

  \item Extensive experiments are conducted to evaluate the proposed system. Results show that InTEn-LOAM achieves similar or better accuracy in comparison with state-of-the-art LO systems and outperforms them in unstructured scenes with sparse geometric features.

\end{itemize}

\section{RELATED WORK} \label{Sect:2}
\subsection{Point cloud registration and LiDAR odometry}\label{Sect:2:1}
Point cloud registration is the most critical problem in LiDAR-based autonomous driving, which is centered on finding the best relative transformation of point clouds. Existing registration techniques can be either categorized into feature-based and scan-based methods \cite{furukawa2015fast} in terms of the type of data, or local and global methods \cite{zong2018asurvey} in terms of the registration reference. Though the local registration requires a good transformation initial, it has been widely used in LO solutions since sequentially LiDAR sweeps commonly share large overlap, and a coarse initial can be readily predicted.

For feature-based approaches, different types of encoded features, e.g., FPFH (fast point feature histogram) \cite{rusu2009fast}, CGF (compact geometric feature) \cite{khoury2017learning}, and arbitrary shapes are extracted to establish valid data associations. LOAM \cite{zhang2017low} is one of the pioneering works of feature-based LO, which extracts plane and edge features based on the sorted smoothness of each point. Many follow-up works follow the proposed feature extraction scheme \cite{shan2020lio,li2021towards,qin2020lins,lin2021r2live}. For example, LeGO-LOAM \cite{shan2018lego} additionally segmented ground to bound the drift in the ground norm direction. MULLS (multi-metric linear least square) \cite{pan2021mulls} explicitly classifies features into six types, (facade, ground, roof, beam, pillar, and encoded points) using the principle component analysis (PCA) algorithm and employs the least square algorithm to estimate the ego-motion, which remarkably improves the LO performance, especially in unstructured environments. \cite{yin2020cae} proposes a convolutional auto-encoder (CAE) to encode feature points for conducting a more robust point association.

Scan-based local registration methods iteratively assign correspondences based on the closest-distance criterion. The iterative closest point (ICP) algorithm, introduced by \cite{besl1992method}, is the most popular scan registration method. Many variants of ICP have been derived for the past three decades, such as Generalized ICP (GICP) \cite{segal2009generalized} and improved GICP \cite{yokozuka2021litamin2}. Many LO solutions apply variants of ICP to align scans for its simplicity and low computational complexity. For example, \cite{moosmann2011velodyne} employs standard ICP, while \cite{palieri2020locus} and \cite{behley2018efficient} employ GICP and normal ICP. The normal distributions transform (NDT) method, first introduced by \cite{biber2003normal}, is another popular scan-based approach, in which surface likelihoods of the reference scan are used for scan matching. Because of that, there is no need for computationally expensive nearest-neighbor searching in NDT, making it more suitable for LO with large-scale map points \cite{zhou2021s4,zhao2019robust,koide2019portable}.


\subsection{Fusion with point intensity}\label{Sect:2:2}
Some works have attempted to introduce the intensity channel into scan registration. Inspired by GICP, \cite{servos2017multi} proposes the multi-channel GICP (MCGICP), which integrates clolor and intensity information into the GICP framework by incorporating additional channel measurements into the covariances of points. In \cite{khan2016modeling}, a data-driven intensity calibration approach is presented to acquire a pose-invariant measure of surface reflectivity. Based on that, \cite{wang2021intensity} establishes voxel-based intensity constraints to complement the geometric-only constraints in the mapping thread of LOAM. \cite{pan2021mulls} assigns higher weights for associations with similar intensities to suppress the effect of outliers adaptively. Besides, the end-to-end learning-based registration framework, named Deep VCP (virtual corresponding points)\cite{lu2019deepvcp}, is proposed, achieving comparable accuracy to prior state-of-the-arts. 
The intensity channel is used to find stable and robust feature associations, which are helpful to avoid the interference of negative true matchings.


\subsection{Dynamic object removal}\label{Sect:2:3}
A good amount of learning-based works related to dynamic removal have been 
reported in \cite{guo2020deep}. In general, the trained model is used to predict 
the probability score that a point originated from dynamic objects. The 
model-based approaches enable to filter out of the dynamics independently, but 
they also require laborious training tasks, and the segmentation performance is 
highly dependent on the training dataset.

Traditional model-free approaches rely on differences between the current laser scan 
and previous scans \cite{yoon2019mapless,dewan2016motion, kim2020remove}. Though it's 
convenient and straightforward, only points that have fully moved outside their 
original position can be detected/removed.

\section{METHODOLOGY} \label{Sect:3}
\begin{figure}[ht]
  \centering
  \includegraphics[width=0.7\textwidth] {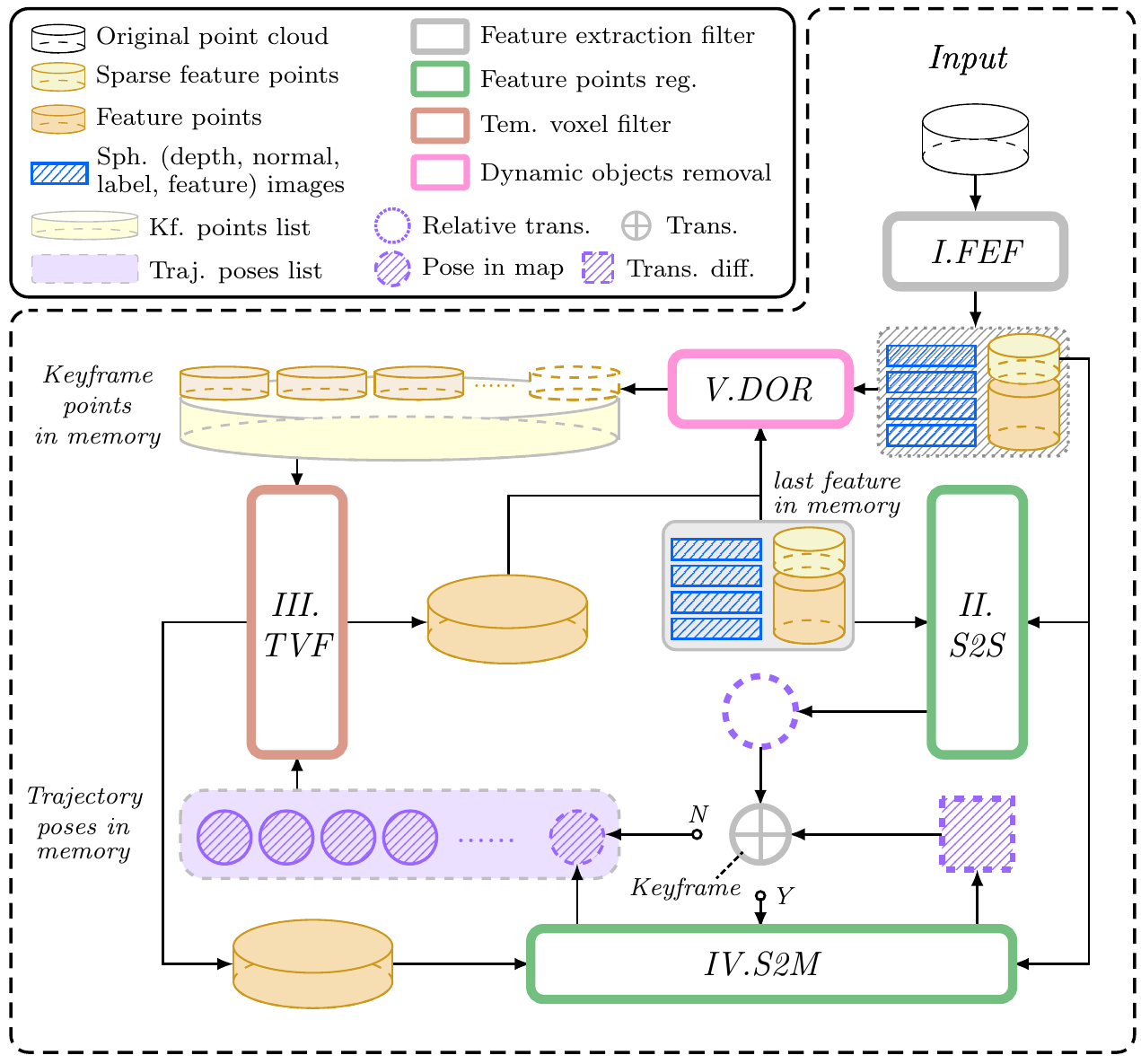}
  \caption[fontsize=6pt]{Overall workflow of InTEn-LOAM.
  } \label{FIG:2}
\end{figure}  
The proposed framework of InTEn-LOAM consist of 5 submodules, i.e., feature extraction 
filter (FEF), scan-to-scan registration (S2S), scan-to-map registration (S2M), 
temporal-based voxel filter (TVF) and dynamic object removal (DOR) (see 
Fig.\ref{FIG:2}). 
Following LOAM, the LiDAR odometry and mapping are executed on two parallel threads 
to improve the running efficiency.
\begin{figure}[htb]
  \centering
  \vspace{-4mm}
  \includegraphics[width=0.9\textwidth] {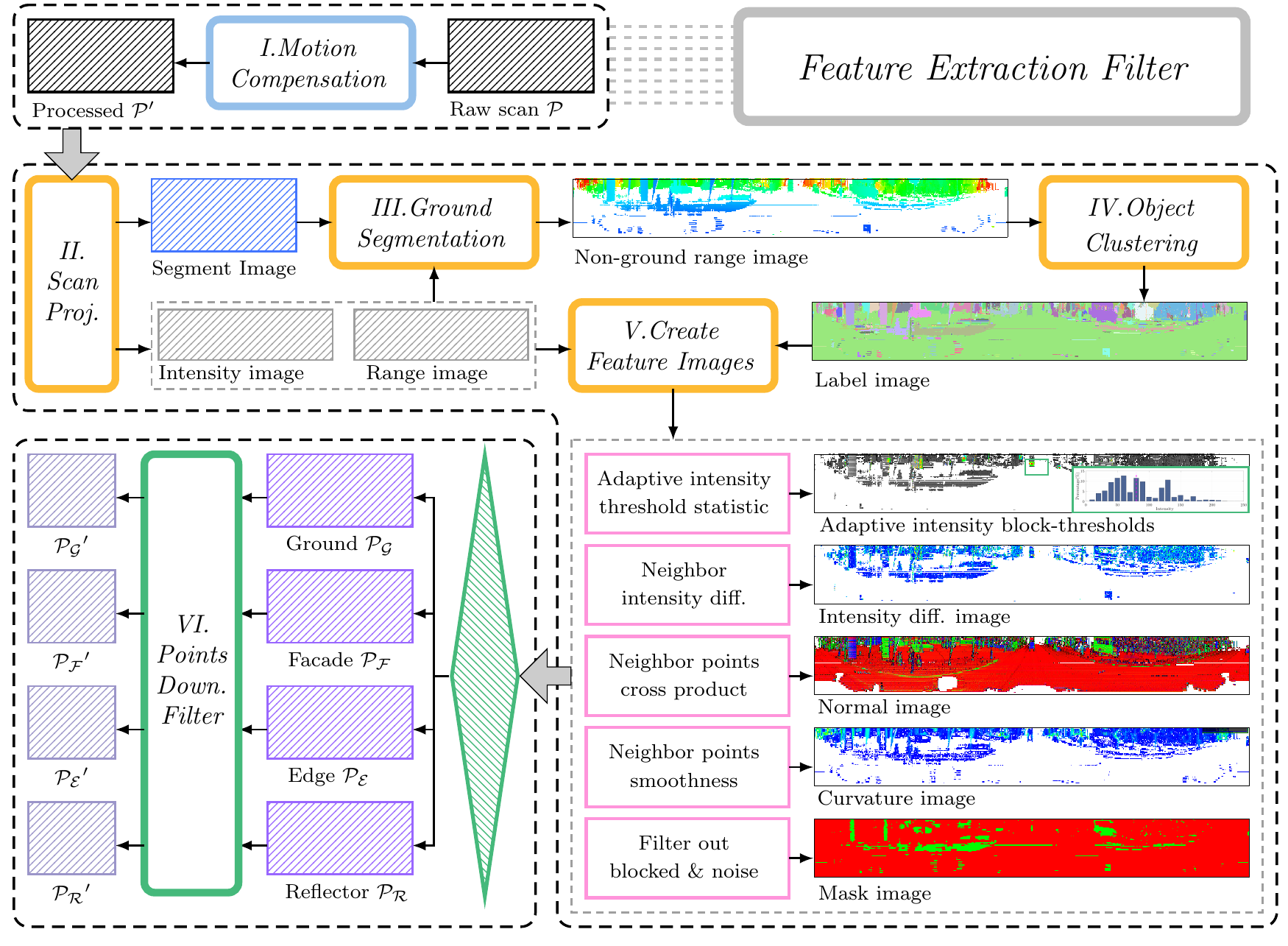}
  \caption[fontsize=6pt]{The workflow of FEF.
  } \label{FIG:3}
\end{figure}  

\subsection{Feature extraction filter} \label{Sect:3:1}
The workflow of FEF is summarized in Fig.\ref{FIG:3}, which corresponds to the gray 
block in Fig.\ref{FIG:2}. 
The FEF receives a raw scan frame and outputs four types of 
features, i.e., ground, facade, edge, and reflector, and two types of 
cylindrical images, i.e., range and label image.
\subsubsection{\textbf{Motion compensation}}
Given the point-wise timestamp of a scan $\mathcal{P}$, the reference pose for 
a point ${\bf{p}}_i \in \mathcal{P}$ at timestamp $\tau_i$ can be interpolated by 
the relative transformation ${\bf{T}}_{e,s} = [{\bf{R}}_{e,s}, {\bf{t}}_{e,s}]$ under 
the assumption of uniform motion:
\begin{equation} \label{EQ:1}
  {\bf{T}}_{s,i} = [\text{slerp}({\bf{R}}_{e,s}, s_i)^{\top}, 
                   -s_i \cdot {\bf{T}}_{e,s}^{-1} \cdot {\bf{t}}_{e,s}],
\end{equation}
where $\text{slerp}(\cdot)$ represents the spherical linear interpolation. The time 
ratio $s_i$ is $s_i = \frac{\tau_i-\tau_s}{\tau_e-\tau_s}$, where $\tau_s$, $\tau_e$ 
stand for the start and end timestamps of the laser sweep, respectively. Then, the 
distorted points can be deskewed by transforming to the start timestamp 
${\bf{T}}_{s,i} \cdot {\bf{p}}_i \in {\mathcal{P}}^{\prime}$.

\subsubsection{\textbf{Scan preprocess}}
The undisordered points ${\mathcal{P}}^{\prime}$ are first preprocessed. The main steps are as below: 

\textit{I. Scan projection.} ${\mathcal{P}}^{\prime}$ is projected into a cylindrical plane to generate range and intensity images, i.e. $\mathcal{D}$ and $\mathcal{I}$ (see Fig.\ref{FIG:1}(d)and (e)). A point with 3D coordinates ${\mathbf{p}}_{i} = [x,y,z]^{\top}$ can be projected as a cylindrical image pixel $[u,v]^{\top}$ by:
\begin{equation} \label{EQ:2}
  \begin{pmatrix}
    u\\ 
    v
  \end{pmatrix} = 
  \begin{pmatrix}
    [1 - \arctan(y,x) \cdot \pi^{-1}] \cdot \frac{w}{2}\\ 
    (\arcsin(\frac{z}{\sqrt{x^2+y^2+z^2}})+\theta_{d}) \cdot \frac{h}{\theta}
  \end{pmatrix},
\end{equation}
where $\theta = \theta_{d} + \theta_{t}$ is the vertical field-of-view of the LiDAR, 
and $w, h$ are the width and height of the resulting image. In $\mathcal{D}$ and 
$\mathcal{I}$, each pixel contains the smallest range and the largest reflectance 
of scanning points falling into the pixel, respectively. In addition, 
${\mathcal{P}}^{\prime}$ is also preprocessed as segment image $\mathcal{S}$ (see 
Fig.\ref{FIG:1}(b)) according to azimuthal and radial directions of 3D points, 
and each pixel contains the lowest $z$. The former converter is the same as $u$ in 
E.q.(\ref{EQ:2}), while the latter is equally spaced with the distance interval 
$\Delta\rho$:
\begin{equation} \label{EQ:3}
  \rho = \lfloor \sqrt{(x^2+y^2+z^2)}/{\Delta\rho} \rfloor,
\end{equation}
where $\lfloor \cdot \rfloor$ indicates rounding down operator. Note that the size 
of $\mathcal{S}$ is not the same as $\mathcal{D}$.

\textit{II. Ground segmentation}. The method from \cite{himmelsbach2010fast}
is applied in this paper with the input of segment image $S$. Each column 
of $\mathcal{S}$ is fitted as a ground line ${\mathbf{l}}_i = a_i \cdot \rho + b_i$. 
Then, residuals can be calculated, which represents the differences between the 
predicted and the observed $z$:
\begin{equation} \label{EQ:4}
  r(u, v) = {\mathbf{l}}_{i}({\mathcal{D}}(u, v)) 
          - {\mathcal{D}}(u, v) \cdot \sin(\theta_{v}),
\end{equation}
where $\theta_{v}$ indicates the vertical angle of the $v$th row in $\mathcal{D}$. 
Pixels with residuals smaller than the threshold ${\mathrm{Th}}_g$ will be marked 
as ground pixels with label identity $1$.

\textit{III. Object clustering}. After the ground segmentation, the angle-based object 
clustering approach from \cite{bogoslavskyi2017efficient} is conducted to group 
pixels into different clusters with identified labels and generate a label image 
$\mathcal{L}$ (see the label image in Fig.\ref{FIG:2}).  

\textit{IV. Create feature images}. We partition the intensity image ${\mathcal{I}}$ into $M \times N$ blocks and establish intensity histograms for each block. The extraction threshold $\mathrm{Th}_{I,n}$ of each intensity block is adaptively determined by taking the median of the histogram. Besides, intensity difference image ${\mathcal{I}}_{\Delta}$, normal image ${\mathcal{N}}$, curvature image ${\mathcal{C}}$, are created by:
\begin{equation} \label{EQ:5}
  \begin{aligned}
    &{\mathcal{I}}_{\Delta}(u,v) = {\mathcal{I}}(u,v) - {\mathcal{I}}(u,v+1), \\ 
    &\begin{aligned}
      {\mathcal{N}}(u,v) &= ({\Pi}[{\mathcal{D}}(u+1,v)] - 
                                 {\Pi}[{\mathcal{D}}(u,v)]) \\
       &\times ({\Pi}[{\mathcal{D}}(u,v+1)] - {\Pi}[{\mathcal{D}}(u,v)]), 
    \end{aligned} \\
    &\begin{aligned}
      & {\mathcal{C}}(u,v) = \\
      & \frac{1}{N \cdot {\mathcal{D}}(u,v)} \cdot 
      \sum_{i,j \in N}{({\mathcal{D}}(u,v) - {\mathcal{D}}(u+i,v+j))}   
    \end{aligned} \\
  \end{aligned} 
\end{equation}
where ${\Pi}[\cdot]:{{\mathcal{D}} \mapsto {\mathcal{P}}}$ denotes the mapping 
function from a range image pixel to a 3D point. $N$ is the neighboring pixels count. 
Furthermore, pixels in the cluster with fewer than 15 points are marked as noises 
and blocked. All the valid-or-not flag is stored in a binary mask image ${\mathcal{B}}$.

\subsubsection{\textbf{Feature extraction}}
According to the above feature images, pixels of four categories of features can be 
extracted. Then 3D feature points, i.e., ground $\mathcal{P_{G}}$, facade 
$\mathcal{P_{F}}$, edge $\mathcal{P_{E}}$, and reflector $\mathcal{P_{R}}$, can be 
obtained per the pixel-to-point mapping relationship. Specifically, 

\begin{itemize}

  \item Points correspond to pixels that meet $\mathcal{L}(u, v)=1 $ and 
  $\mathcal{B}(u, v)\neq0$ are categorized as $\mathcal{P_{G}}$.
  
  \item Points correspond to pixels that meet $\mathcal{C}(u, v)>Th_E $ and 
  $\mathcal{B}(u, v)\neq0$ are categorized as $\mathcal{P_{E}}$.

  \item Points correspond to pixels that meet $\mathcal{C}(u, v)<Th_F $ and 
  $\mathcal{B}(u, v)\neq0$ are categorized as $\mathcal{P_{F}}$.

  \item Points correspond to pixels that meet $\mathcal{I_\Delta}(u, v)>Th_{\Delta I}$ 
  and $\mathcal{B}(u, v)\neq0$ are categorized as $\mathcal{P_{R}}$. 
  Besides, to keep the gradient of local intensities points in pixels that meet 
  $\mathcal{I}(u, v)>Th_{I,n}$, as well as their neighbors are all included in 
  $\mathcal{P_{R}}$.

\end{itemize}

To improve the efficiency of scan registration, the random downsample filter (RDF) 
is applied on $\mathcal{P_{G}}$ and $\mathcal{P_{R}}$ to obtain ${N}_{\mathcal{G}}$ 
downsampled edge features $\mathcal{P_{G}}^{\prime}$ and ${N}_{\mathcal{R}}$ facade 
features $\mathcal{P_{R}}^{\prime}$. To obtain ${N}_{\mathcal{E}}$ refined edge 
features $\mathcal{P_{E}}^{\prime}$ and ${N}_{\mathcal{F}}$ refined facade features 
$\mathcal{P_{F}}^{\prime}$, the non-maximum suppression (NMS) filter based on point 
curvatures is applied on $\mathcal{P_{E}}$ and $\mathcal{P_{F}}$. 

\subsection{Intensity-based scan registration} \label{Sect:3:2}
Similar to the geometric-based scan registration, given the initial guess of the 
transformation $\bar{\mathbf{T}}_{t,s}$ from source points ${\mathcal{P}}_{s}$ to 
target points ${\mathcal{P}}_{t}$, we try to estimate the LiDAR motion 
$\mathbf{T}_{t,s}$ by matching the local intensities of the source and target. In the 
case of geometric features registration, the motion estimation is solved through 
nonlinear iterations by minimizing the sum of Euclidean distances from each source 
feature to their correspondence in the target scan. In the case of reflecting 
features registration, however, we minimize the sum of intensity differences instead. 
The fundamental idea of the intensity-based point cloud alignment method proposed in 
this paper is to make use of the similarity of intensity gradients within the local 
region of laser scans to achieve scan matching. 
\begin{figure}[ht]
  \centering
  \vspace{-2mm}
  \includegraphics[width=0.4\textwidth] {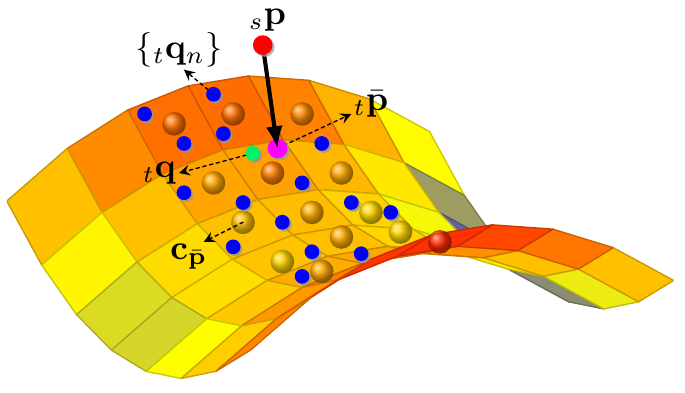}
  \caption[fontsize=6pt]{An simple example of B-spline intensity surface model. The 
  grid surface depicts the modeled continuous intensity surface with colors 
  representing intensities, and spheres in the center of surface grids representing 
  contol points of the B-spline surface model. \textcolor{red}{$_{s}{\mathbf{p}}$} 
  denotes the selected point, and \textcolor{blue}{$\left\{ _{t}{\mathbf{q}_n} \right\}$} denotes querry 
  points. \textcolor{red}{$_{s}{\mathbf{p}}$} is transformed to the reference frame 
  of \textcolor{blue}{$\left\{ _{t}{\mathbf{q}_n} \right\}$} and denoted as 
  \textcolor{magenta}{$_{t}{\bar{\mathbf{p}}}$}. \textcolor{green}{$_{t}{\mathbf{q}}$} 
  denotes the nearest neighboring querry point of \textcolor{magenta}{$_{t}{\bar{\mathbf{p}}}$}.
  } \label{FIG:5}
\end{figure}  
Because of the discreteness of the laser scan, sparse 3D points in a local area are 
not continuous, causing the intensity values of the laser sweep non-differentiable.
To solve this issue, we introduce a continuous intensity surface model using the 
local support characteristic of the B-spline basis function. A simple intensity 
surface example is shown in Fig.\ref{FIG:5}.

\subsubsection{\textbf{B-spline intensity surface model}}
The intensity surface model presented in this paper uses the uniformly distributed 
knots of the B-spline; thus, the B-spline is defined fully by its degree 
\cite{sommer2020efficient}. Specifically, the intensity surface is a space spanned 
by three $d$-degree B-spline functions on the orthogonal axes, and each B-spline is 
controlled by $d+1$ knots on the axis. Mathematically, the B-spline intensity surface 
in local space is a scalar-valued function ${\mathrm{\mu}}({\mathbf{p}}):{\mathbb{R}}^3 \to {\mathbb{R}}$, 
which builds the mapping relationship between a 3D point ${\mathbf{p}} = 
[x, y, z]^{\top}$ and its intensity value. The mapping function is defined by the 
tensor product of three B-spline functions and control points ${\mathrm{c}}_{i,j,k} \in C$ in the local space: 
\begin{equation} \label{EQ:6}
  \begin{aligned}
    {\mathrm{\mu}}({\mathbf{p}}) &= \sum_{i=0}^{d+1}\sum_{j=0}^{d+1}\sum_{k=0}^{d+1} 
                    {{\mathrm{c}}_{i,j,k} b_{i}^{d}(x) b_{j}^{d}(y) b_{k}^{d}(z)} \\ 
    &= {\mathrm{vec}({\mathbf{b}}_{x}^d \otimes {\mathbf{b}}_{x}^d \otimes {\mathbf{b}}_{z}^d)^{\top}
     \cdot \mathrm{vec}(C)} \\
    &= \bm{\phi} ({\mathbf{p}})^{\top} \cdot {\mathbf{c}}
  \end{aligned}
\end{equation}
where ${\mathbf{b}}^d$ is the $d$ degree B-spline function. We use the vectorization 
operator ${\mathrm{vec}}(\cdot)$ and Kronecker product operator $\otimes$ to transform 
the above equation in the form of matrix multiplication. In this paper, the cubic 
$(d = 3)$ B-spline function is employed.

\subsubsection{\textbf{Observation constraint}}
The intensity observation constraint is defined as the residual between the intensities 
of source points and their predicted intensities in the local intensity surface model. 
Fig.\ref{FIG:5} demonstrates how to predict the intensity on the surface patch for 
a reflector feature point. The selected point ${_{s}\mathbf{p}} \in {\mathcal{P}}_{s}$ 
with intensity measurement $\eta$ is transformed to the model frame by 
$_{t}\bar{\mathbf{q}} = \bar{\mathbf{T}}_{t,s} \cdot _{s}{\mathbf{p}}$. 
Then the nearest point $_{t}{\mathbf{q}} \in {\mathcal{P}}_{t}$ and its R-neighbor 
points $_{t}{\mathbf{q}}_{n} \in {\mathcal{P}}_{t}, n = 1 \cdots N$ can be searched. 
Given the uniform space of the B-spline function $\kappa$, the neighborhood points 
$_{t}{\mathbf{q}}_{n}$ can be voxelized with the center $_{t}{\mathbf{q}}$ and the 
resolution $\kappa \times \kappa \times \kappa$ to generate control knots 
$\mathbf{c}_{\bar{\mathbf{p}}}$ for the local intensity surface. The control knot takes 
the value of the average intensities of all points in a voxel. To sum up, the 
residual is defined as: 
\begin{equation} \label{EQ:7}
  r_{\mathcal{I}}(\tilde{\mathbf{T}}_{t,s}) = [\bm{\phi} (\bar{\mathbf{T}}_{t,s} \cdot 
  _s{\mathbf{p}})^{\top} \cdot {{\mathbf{c}}_{\bar{\mathbf{q}}}} - \eta].
\end{equation}
Stacking normalized residuals to obtain residual vector 
${\mathbf{r}}_{\mathcal{I}}(\tilde{\mathbf{T}}_{t,s})$, and computing the Jacobian 
matrix of ${\mathbf{r}}_{\mathcal{I}}$ w.r.t ${\mathbf{T}}_{t,s}$, denoted as 
${\mathbf{J}}_{\mathcal{I}} = \partial {\mathbf{r}}_{\mathcal{I}} / 
\partial {\mathbf{T}}_{t,s}$. 
The constructed nonlinear optimization problem can be solved by minimizing ${\mathbf{r}}_
{\mathcal{I}}$ toward zero using L-M algorithm. Note that \textit{Lie group} 
and \textit{Lie algebra} are implemented for the 6-DoF transformation in this paper.

\subsection{Dynamic object removal}
\begin{figure}[ht]
  \centering
  \includegraphics[width=0.8\textwidth] {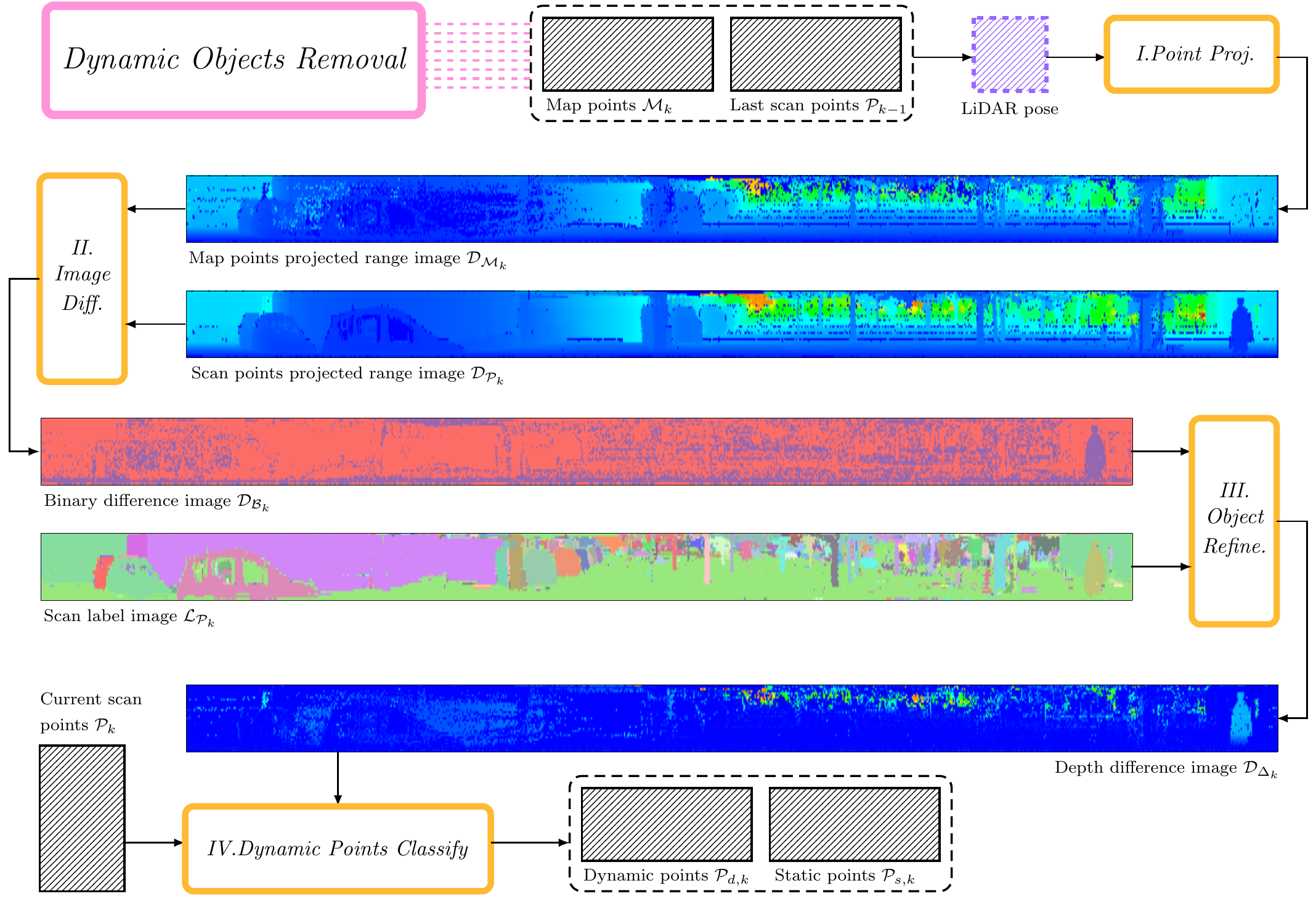}
  \caption[fontsize=6pt]{The workflow of DOR.
  } \label{FIG:6}
\end{figure} 

The workflow of the proposed DOR is shown in Fig.\ref{FIG:6}, which corresponds to 
the pink block in Fig.\ref{FIG:2}. Inputs of the DOR filter include the current 
laser points ${\mathcal{P}}_{k}$, the previous static laser points 
${\mathcal{P}}_{s,k-1}$, the local map points ${\mathcal{M}}_{k}$, the current 
range image ${\mathcal{D}}_{\mathcal{P}_k}$, the current label image 
${\mathcal{L}}_{\mathcal{P}_k}$, and the estimated LiDAR pose in the world frame 
$\tilde{{\mathbf{T}}}_{w,k}$. The filter divides ${\mathcal{P}}_{k}$ into two 
categories, i.e., the dynamic ${\mathcal{P}}_{d,k}$ and static ${\mathcal{P}}_{s,k}$. 
Only static points will be appended into the local map for map update. The DOR filter 
introduced in this paper exploits the similarity of point clouds in the adjacent 
time domain for dynamic points filtering and verifies dynamic objects based on the 
segmented label image. 

\subsubsection{\bf{Rendering range image for the local map}} Both downsampling with 
coarse resolution and uneven distribution of map points may result in pixel holes in 
the rendered range image. Considering the great similarity of successive laser sweeps 
in the time domain, we use both the local map points ${\mathcal{M}}_{k}$ and the 
previous static laser points ${\mathcal{P}}_{s,k-1}$ to generate the 
to-be-rendered map points $\mathcal{E}_{k}$:
\begin{equation} \label{EQ:8}
  \mathcal{E}_{k} = \mathbf{T}_{w,k}^{-1} \cdot \mathcal{M}_{k} \cup 
  \mathbf{T}_{k,k-1} \cdot {\mathcal{P}}_{s,k-1}.
\end{equation} 
The rendered iamge ${\mathcal{D}}_{\mathcal{M}_k}$ and the current scan image ${\mathcal{D}}_{\mathcal{P}_k}$ are shown in the second and third rows of Fig.\ref{FIG:6}. A pedestrian can be clearly distinguished in ${\mathcal{D}}_{\mathcal{P}_k}$ but not in ${\mathcal{D}}_{\mathcal{M}_k}$. 
\subsubsection{\bf{Temporal-based dynamic points searching}} Dynamic pixels in 
${\mathcal{D}}_{\mathcal{P}_k}$ can be coarsely screened out in accordance with the 
depth differences between ${\mathcal{D}}_{\mathcal{P}_k}$ and ${\mathcal{D}}_{\mathcal{M}_k}$. 
In particular if the depth difference at $[u, v]^{\top}$ is larger than the 
threshold $Th_{\Delta d}$, the pixel will be marked as dynamic. Consecutively, 
we can also generate a binary image ${\mathcal{D}}_{\mathcal{B}_k}$ indicating whether 
the pixel is dynamic or not:
\begin{equation} \label{EQ:9}
  \begin{aligned}
    {\mathcal{D}}_{\Delta_k}(u,v) = |{\mathcal{D}}_{\mathcal{M}_k}(u, v) &- {\mathcal{D}}_{\mathcal{P}_k}(u, v)| > 
    Th_{\Delta d}? \\
    {\mathcal{D}}_{\mathcal{B}_k}(u, v) &= 1 : 0,
  \end{aligned}
\end{equation} 
where ${\mathcal{D}}_{\mathcal{M}_k}(u, v) \neq 0$ and 
${\mathcal{D}}_{\mathcal{P}_k}(u, v) \neq 0$. An example of 
${\mathcal{D}}_{\mathcal{B}_k}$ is shown in the fourth row of Fig.\ref{FIG:6}, in 
which red pixels represent the static and purple pixels represent the dynamic. To 
improve the robustness of the DOR filter to different point depths, we use the adaptive 
threshold $Th_{\Delta d} = s_d \cdot {\mathcal{D}}_{\mathcal{P}_k}(u, v)$, where $s_d$ 
is a constant coefficient. 

\subsubsection{\bf{Dynamic object validation}} It can be seen from 
${\mathcal{D}}_{\mathcal{B}_k}$ that it generates a large number of false positive 
(FP) dynamic pixels using the pixel-by-pixel depth comparison. To handle the above 
issue, we utilize the label image to validate dynamic according to the fact that 
points originating from the same object should have the same status label. We denote 
the pixel number of a segmented object and the dynamic pixel number as $N_i$ 
and $N_{d,i}$, which can be counted from ${\mathcal{L}}_{\mathcal{P}_k}$ and 
${\mathcal{D}}_{\mathcal{B}_k}$, respectively. Two basic assumptions generally hold 
in terms of dynamic points, i.e., a. ground points cannot be dynamic; b. the 
percentage of FP dynamic pixels in a given object will not be significant. According 
to the above assumptions, we can validate dynamic pixels at the object level:
\begin{equation} \label{EQ:10}
  \begin{aligned}
    \frac{N_{d,i}}{N_i} \geq Th_{N} \ &\& \ {\mathcal{L}}_{\mathcal{P}_k}(u, v) \neq 1? \\  
    {\mathcal{D}}_{\Delta_k}(u, v) &= {\mathcal{D}}_{\Delta_k}(u, v) : 0.
  \end{aligned}
\end{equation} 
The E.q.(\ref{EQ:10}) indicates that only objects that is marked as the non-ground object or own the dynamic pixel ratio larger than the threshold will be recognized as dynamic. In ${\mathcal{D}}_{\Delta_k}$, pixels belonging to dynamic objects will retain the depth differences, while the others will be reset as 0. As the depth difference image shown in the sixth row of Fig.\ref{FIG:6}, though many FP dynamic pixels are filtered out after the validation, the true positive (TP) dynamic pixels from the moving pedestrian on the right side are still remarkable. Then, the binary image ${\mathcal{D}}_{\mathcal{B}_k}$ is updated by substituting the refined ${\mathcal{D}}_{\Delta_k}$ into E.q.(\ref{EQ:9}).
\subsubsection{\bf{Points classification}}
According to ${\mathcal{D}}_{\mathcal{B}_k}$, dynamic 3D points in extracted features can be marked using the mapping function ${\Pi}[\cdot]:{{\mathcal{D}} \mapsto {\mathcal{P}}}$. Since the static feature set is the complement of the dynamic feature set w.r.t. the full set of extracted features, the static features can be filtered by ${\mathcal{P}}_{s,k} = {\mathcal{P}}_{k} - {\mathcal{P}}_{d,k}$. 

\subsection{LiDAR odometry}
Given the initial guess $\bar{\mathbf{T}}_{k,k-1}$, extracted features, i.e., downsampled ground and reflector features $\mathcal{P_G}^{\prime}$ and $\mathcal{P_R}^{\prime}$, as well as refined edge and facade features $\mathcal{P_E}^{\prime}$ and $\mathcal{P_F}^{\prime}$, are utilized to estimate the optimal estimation of $\mathbf{T}_{k,k-1}$, and then the LiDAR pose $\mathbf{T}_{w,k}$ in the global frame is reckoned. The odometry thread corresponds to the green S2S block in Fig.\ref{FIG:2}, and the pseudo code is shown in Algorithm \ref{ALG:1}. To improve the performance of geometric-only scan registration, the proposed LO incorporates reflector features and estimate relative motion by jointly solving the multi-metric nonlinear optimization(NLO).
\begin{figure}[ht] 
  \centering
  \subfloat[]
  {
    \begin{minipage}[b]{0.17\textwidth}
      \centering
      \includegraphics[width=\textwidth]{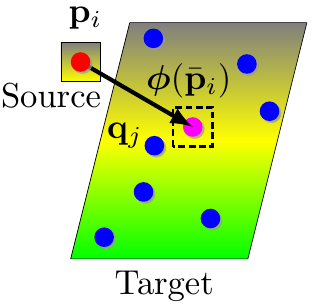}\vspace{4.pt}	
	  \end{minipage}
  }    
  \subfloat[]{
    \begin{minipage}[b]{0.113\textwidth}
	    \centering
	    \includegraphics[width=\textwidth]{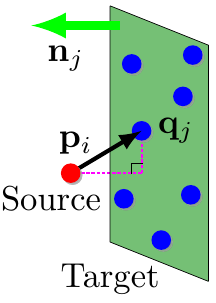}\vspace{4.pt}	
	  \end{minipage}
  }
  \subfloat[]{
    \begin{minipage}[b]{0.111\textwidth}
	    \centering
	    \includegraphics[width=\textwidth]{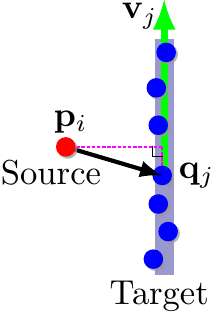}\vspace{4.pt}	
	  \end{minipage}
  }
  \subfloat[]{
    \begin{minipage}[b]{0.208\textwidth}
	    \centering
	    \includegraphics[width=\textwidth]{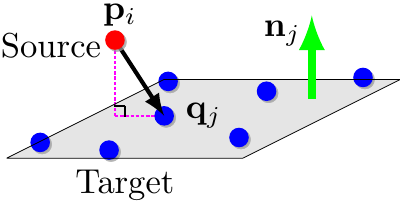}\vspace{4.pt}	
	  \end{minipage}
  } 
  \caption{Overview of four different types of feature associations. (a) Reflector ; 
  (b) Facade; (c) Edge; (d) Ground feature association.}\label{FIG:4}	
\end{figure}

\begin{algorithm}[!t]\small
  \caption{LiDAR Odometry}
  \label{ALG:1}
  \LinesNumbered 
  \SetKwFunction{GroundAssoc}{GroundAssoc}
  \SetKwFunction{FacadeAssoc}{FacadeAssoc}
  \SetKwFunction{EdgeAssoc}{EdgeAssoc}
  \SetKwFunction{ReflectAssoc}{ReflectAssoc}
  \SetKwFunction{MultiMetricNLO}{MultiMetricNLO}
  \SetKwFunction{ConvergCond}{ConvergCond}
  \KwIn{Extracted feature points $\mathcal{P}_{\mathcal{G},k}^{\prime}$, $\mathcal{P}_{\mathcal{F},k}^{\prime}$, $\mathcal{P}_{\mathcal{E},k}^{\prime}$, $\mathcal{P}_{\mathcal{R},k}^{\prime}$, initial transform $\bar{\mathbf{T}}_{k,k-1}$}
  \KwOut{estimated transform $\mathbf{T}_{k,k-1}$, reckoned pose $\mathbf{T}_{w^{\prime},k}$}
  \emph{reference feature points $\mathcal{P}_{\mathcal{G},k-1}^{\prime}$, $\mathcal{P}_{\mathcal{F},k-1}^{\prime}$, $\mathcal{P}_{\mathcal{E},k-1}^{\prime}$, $\mathcal{P}_{\mathcal{R},k-1}^{\prime}$, and the last LO reckoned pose $\mathbf{T}_{w^{\prime},k-1}$ can be loaded from the buffer\;} 
  
  \tcp{\footnotesize{Main}}
  \If{the system is not initialized}{
    $\mathbf{T}_{k,k-1} \gets \mathbf{I}_{4 \times 4}$\;
    $\mathbf{T}_{w^{\prime},k} \gets \mathbf{I}_{4 \times 4}$\;
  }
  \Else{
    \For{a number of iterations}{
      \tcp{\footnotesize{Find feature associations by parallel threads}}
      $\mathbf{r}_{\mathcal{G}}, \mathbf{J}_{\mathcal{G}} \gets$ \GroundAssoc{$\bar{\mathbf{T}}_{k,k-1}$, $\mathcal{P}_{\mathcal{G},k-1}^{\prime}$, $\mathcal{P}_{\mathcal{G},k}^{\prime}$}\; \label{L:2}

      $\mathbf{r}_{\mathcal{F}}, \mathbf{J}_{\mathcal{F}} \gets$ \FacadeAssoc{$\bar{\mathbf{T}}_{k,k-1}$, $\mathcal{P}_{\mathcal{F},k-1}^{\prime}$, $\mathcal{P}_{\mathcal{F},k}^{\prime}$}\; \label{L:3}

      $\mathbf{r}_{\mathcal{E}}, \mathbf{J}_{\mathcal{E}} \gets$ \EdgeAssoc{$\bar{\mathbf{T}}_{k,k-1}$, $\mathcal{P}_{\mathcal{E},k-1}^{\prime}$, $\mathcal{P}_{\mathcal{E},k}^{\prime}$}\; \label{L:1}

      $\mathbf{r}_{\mathcal{R}}, \mathbf{J}_{\mathcal{R}} \gets$ \ReflectAssoc{$\bar{\mathbf{T}}_{k,k-1}$, $\mathcal{P}_{\mathcal{R},k-1}^{\prime}$, $\mathcal{P}_{\mathcal{R},k}^{\prime}$}\; \label{L:4}

      \tcp{\footnotesize{Update relative transform by the nonlinear optimization}}
      $\tilde{\mathbf{T}}_{k-1, k} \gets$ \MultiMetricNLO{$\mathbf{J}, \mathbf{r}$}\; \label{L:5}

      \tcp{\footnotesize{Convergency}}
      $convergency \gets$ \ConvergCond{$\tilde{\mathbf{T}}_{k-1, k} \cdot \bar{\mathbf{T}}_{k,k-1}^{-1}$}\;

      \tcp{\footnotesize{Update parameters}}
      ${\mathbf{T}}_{k,k-1} \gets \tilde{\mathbf{T}}_{k-1, k}$\;
      $\mathbf{T}_{w^{\prime},k} \gets \mathbf{T}_{w^{\prime},k-1} \cdot {\mathbf{T}}_{k,k-1}^{-1}$\;
      \If{$convergency$}{
        \bf{break}\;
      }
    }
  }
\end{algorithm}
\subsubsection{\bf{Constraint model}} As shown in Fig.\ref{FIG:4}, constraints are modeled as the point-to-model intensity difference (for reflector feature) and the point-to-line (for edge feature)/point-to-plane (for ground and facade feature) distance, respectively. 

\textit{I.Point-to-line constraint.} Let $\mathbf{p}_i \in \mathcal{P}_{\mathcal{E},k}^{\prime}, i = 1 \cdots N_E$ be a edge feature point. The association of $\mathbf{p}_i$ is the line connected by $\mathbf{q}_j, \mathbf{q}_m \in \mathcal{P}_{\mathcal{E}, k-1}^{\prime}$, which represent the closest point of $\bar{\mathbf{T}}_{k-1, k} \cdot \mathbf{p}_i$ in $\mathcal{P}_{\mathcal{E},k-1}^{\prime}$ and the closest neighbor in the preceding and following scan lines to the $\mathbf{q}_j$ respectively. The constraint equation is formulated as the point-to-line distance:
\begin{equation} \label{EQ:12}
  \begin{aligned}
    r_{\mathcal{E}, i} = \Vert \mathbf{v}_{j} &\times 
    \left( \mathbf{T}_{k-1, k} \cdot \mathbf{p}_i \right) \Vert, \\
    \mathbf{v}_{j} &= \frac{\mathbf{q}_j - \mathbf{q}_m}
    {\Vert \mathbf{q}_j - \mathbf{q}_m \Vert}.
  \end{aligned}
\end{equation}
The $N_E \times 1$ edge feature error vector $\mathbf{r}_{\mathcal{E}}$ is constructed by stacking all the normalized edge residuals(Line \ref{L:1}).

\textit{II.Point-to-plane constraint.} Let $\mathbf{p}_i \in \mathcal{P}_{\mathcal{F},k}^{\prime}, (\mathcal{P}_{\mathcal{G},k}^{\prime}), i = 1 \cdots N_F (N_G)$ be a facade or ground feature point. The association of $\mathbf{p}_i$ is the plane constructed by $\mathbf{q}_j, \mathbf{q}_m, \mathbf{q}_n$ in the last ground and facade feature points, which represent the closest point of $\bar{\mathbf{T}}_{k-1, k} \cdot \mathbf{p}_i$, the closest neighbor in the preceding and following scan lines to $\mathbf{q}_j$ and the closest neighbor in the same scan line to $\mathbf{q}_j$ respectively. The constraint equation is formulated as the point-to-plane distance:
\begin{equation} \label{EQ:13}
  \begin{aligned}
    r_{\mathcal{G}, i} &= r_{\mathcal{F}, i} = \mathbf{n}_{j} \cdot 
    \left( \mathbf{T}_{k-1, k} \cdot \mathbf{p}_i \right), \\
    \mathbf{n}_{j} &= \frac{(\mathbf{q}_j - \mathbf{q}_m) \times (\mathbf{q}_j - \mathbf{q}_n)}
    {\Vert (\mathbf{q}_j - \mathbf{q}_m)  \times (\mathbf{q}_j - \mathbf{q}_n)\Vert}.
  \end{aligned}
\end{equation}
The $N_F \times 1$ facade feature error vector $\mathbf{r}_{\mathcal{F}}$ and the $N_G \times 1$ ground feautre error vector $\mathbf{r}_{\mathcal{G}}$ are constructed by stacking all normalized facade and ground residuals(Line \ref{L:2}-\ref{L:3}).

\textit{III.Point-to-model intensity difference constraint.} The constraint equation is formulated as E.q.(\ref{EQ:7}). The $N_R \times 1$ intensity feature error vector $\mathbf{r}_{\mathcal{R}}$ is constructed by stacking all reflector features (Line \ref{L:4}).

\subsubsection{\bf{Transformation estimation}}
According to constraint models introduced above, the nonlinear least square (LS) function can be established for the transformation estimation (Line \ref{L:5}):
\begin{equation} \label{EQ:14}
  \tilde{\mathbf{T}}_{k-1, k} = \mathop{\text{argmin}} \limits_{\mathbf{T}_{k-1, k}} 
  \left( \mathbf{r}_{\mathcal{G}}^{\top}\mathbf{r}_{\mathcal{G}} + 
         \mathbf{r}_{\mathcal{F}}^{\top}\mathbf{r}_{\mathcal{F}} + 
         \mathbf{r}_{\mathcal{E}}^{\top}\mathbf{r}_{\mathcal{E}} + 
         \mathbf{r}_{\mathcal{R}}^{\top}\mathbf{r}_{\mathcal{R}} \right).
\end{equation}
The \textit{special euclidean group} $\exp{(\bm{\xi}_{k-1,k}^{\wedge})} = \mathbf{T}_{k-1, k}$ is implemented during the nonlinear optimization iteration. Then $\mathbf{T}_{k-1, k}$ can be incrementally updated by:
\begin{equation} \label{EQ:15}
  \bm{\xi}_{k-1,k} \leftarrow \bm{\xi}_{k-1,k} + \delta\bm{\xi}_{k-1,k}.
\end{equation}
where 
\begin{equation}\label{EQ:16}
\begin{gathered} 
  \delta\bm{\xi}_{k-1,k} = \left(\mathbf{J}^{\top}\mathbf{J}\right)^{-1}\mathbf{J}^{\top}\mathbf{r}, \\
  \mathbf{J} = \left[ \mathbf{J}_{\mathcal{G},i}^{\top}\ \cdots\ \mathbf{J}_{\mathcal{F},i}^{\top}\ \cdots\ \mathbf{J}_{\mathcal{E},i}^{\top}\ \cdots\ \mathbf{J}_{\mathcal{R},i}^{\top} \right]^{\top}, \\
  \mathbf{r} = \left[ \mathbf{r}_{\mathcal{G},i}^{\top}\ \cdots\ \mathbf{r}_{\mathcal{F},i}^{\top}\ \cdots\ \mathbf{r}_{\mathcal{E},i}^{\top}\ \cdots\ \mathbf{r}_{\mathcal{R},i}^{\top} \right]^{\top}.
\end{gathered}
\end{equation}
The Jacobian matrix of constriant equation w.r.t. $\bm{\xi}_{k-1,k}$ is denoted as $\mathbf{J}$. Matrix components are listed as follow.
\begin{equation}\label{EQ:17}
  \begin{aligned} 
    \mathbf{J}_{\mathcal{G},i} &= \frac{\partial \mathbf{r}_{\mathcal{G},i}}
                                       {\partial \delta\bm{\xi}_{k-1,k}}
                               = \mathbf{n}_{j,m,n}^{\top} \cdot 
                                 \frac{\partial(\mathbf{T}_{k-1, k}\mathbf{p}_i)}
                                      {\partial \delta\bm{\xi}_{k-1,k}}, \\
    \mathbf{J}_{\mathcal{F},i} &= \frac{\partial \mathbf{r}_{\mathcal{F},i}}
                                       {\partial \delta\bm{\xi}_{k-1,k}}
                               = \mathbf{n}_{j,m,n}^{\top} \cdot 
                                 \frac{\partial(\mathbf{T}_{k-1, k}\mathbf{p}_i)}
                                      {\partial \delta\bm{\xi}_{k-1,k}}, \\ 
    \mathbf{J}_{\mathcal{E},i} &= \frac{\partial \mathbf{r}_{\mathcal{E},i}}
                                       {\partial \delta\bm{\xi}_{k-1,k}} \\ 
                               &= \frac{(\mathbf{v}_{j,m}^{\wedge}(\mathbf{T}_{k-1, k}
                                         \mathbf{p}_i))^{\top}}
                                       {\Vert \mathbf{v}_{j,m}^{\wedge}
                                        (\mathbf{T}_{k-1, k}\mathbf{p}_i) \Vert} \cdot
                                  \mathbf{v}_{j,m}^{\wedge} \cdot 
                                  \frac{\partial(\mathbf{T}_{k-1, k}\mathbf{p}_i)}
                                       {\partial \delta\bm{\xi}_{k-1,k}}, \\
    \mathbf{J}_{\mathcal{R},i} &= \frac{\partial \mathbf{r}_{\mathcal{R},i}}
                                       {\partial \delta\bm{\xi}_{k-1,k}} \\ 
                               &= \frac{\partial \bm{\phi} (\mathbf{T}_{k-1,k} 
                                                 \mathbf{p}_i)^{\top}}
                                       {\partial (\mathbf{T}_{k-1,k} \mathbf{p}_i)} 
                                  \cdot \frac{\partial(\mathbf{T}_{k-1, k}\mathbf{p}_i)}
                                             {\partial \delta\bm{\xi}_{k-1,k}} 
                                  \cdot {{\mathbf{c}}_{\bar{\mathbf{q}}_i}}.                           
  \end{aligned}
\end{equation}
\subsection{LiDAR mapping}
There is always an inevitable error accumulation in the LiDAR odometry, resulting in a discrepancy $\Delta \mathbf{T}_k$ between the estimated and actual pose. In other words, the estimated transform from the LiDAR odometry thread is not the exact transform from the LiDAR frame $\left\{ L \right\}$ to the world frame $\left\{ W \right\}$ but from $\left\{ L \right\}$ to the drifted world frame $\left\{ W^{\prime} \right\}$: 
\begin{equation} \label{EQ:18}
  \mathbf{T}_{w,k} = \Delta \mathbf{T}_{k} \mathbf{T}_{w^{\prime}, k}.
\end{equation}

\begin{algorithm}[!t]\small
  \caption{LiDAR Mapping}
  \label{ALG:2}
  \LinesNumbered 
  \SetKwFunction{searchSurroundKF}{searchSurroundKF}
  \SetKwFunction{temporalVoxelFilter}{temporalVoxelFilter}
  \SetKwFunction{GroundAssoc}{GroundAssoc}
  \SetKwFunction{FacadeAssoc}{FacadeAssoc}
  \SetKwFunction{EdgeAssoc}{EdgeAssoc}
  \SetKwFunction{ReflectAssoc}{ReflectAssoc}
  \SetKwFunction{MultiMetricNLO}{MultiMetricNLO}
  \SetKwFunction{ConvergCond}{ConvergCond}
  \SetKwFunction{DownsizeFilter}{DownsizeFilter}
  \SetKwFunction{DORFilter}{DORFilter}
  \SetKwFunction{InsertAsKF}{InsertAsKF}
  \KwIn{Extracted feature points for registration $\mathcal{P}_{\mathcal{G},k}^{\prime}$, $\mathcal{P}_{\mathcal{F},k}^{\prime}$, $\mathcal{P}_{\mathcal{E},k}^{\prime}$, $\mathcal{P}_{\mathcal{R},k}^{\prime}$, feature points for mapping, $\mathcal{P}_{\mathcal{G},k}$, $\mathcal{P}_{\mathcal{F},k}$, $\mathcal{P}_{\mathcal{E},k}$, $\mathcal{P}_{\mathcal{R},k}$, estimated transform from LiDAR odometry $\mathbf{T}_{w^{\prime},k}$, scan depth image ${\mathcal{D}}_{k}$, and labaled image ${\mathcal{L}}_{k}$}
  \KwOut{refined pose $\mathbf{T}_{w,k}$, static scan points $\mathcal{P}_{s,k}$}
  \emph{the transform drift $\Delta \mathbf{T}_{k}$ and scan keyframes can be loaded from the buffer\;} 
  
  \tcp{\footnotesize{Roughly transform the reckoned pose to the world frame}}
  $\bar{\mathbf{T}}_{w,k} \gets \Delta \mathbf{T}_{k} \cdot \mathbf{T}_{w^{\prime},k}$\; \label{L:6} 

  \tcp{\footnotesize{Main}}
  \If{skip a number of frames to keep system efficiency}{
    \tcp{\footnotesize{onstruct local points map}}
    $\mathcal{M}_{k} \gets $ \searchSurroundKF{$\bar{\mathbf{T}}_{w,k}$}\; \label{L:7} 
    \temporalVoxelFilter{$\mathcal{M}_{k}$}\; \label{L:8} 

    \For{a number of iterations}{
      \tcp{\footnotesize{Find feature sssociations by parallel threads}}
      $\mathbf{r}_{\mathcal{G}}, \mathbf{J}_{\mathcal{G}} \gets$ \GroundAssoc{$\bar{\mathbf{T}}_{w,k}$, $\mathcal{P}_{\mathcal{G},k}^{\prime}$, $\mathcal{M}_{G,k}$}\; \label{L:9} 

      $\mathbf{r}_{\mathcal{F}}, \mathbf{J}_{\mathcal{F}} \gets$ \FacadeAssoc{$\bar{\mathbf{T}}_{w,k}$, $\mathcal{P}_{\mathcal{F},k}^{\prime}$, $\mathcal{M}_{F,k}$}\; \label{L:10} 

      $\mathbf{r}_{\mathcal{E}}, \mathbf{J}_{\mathcal{E}} \gets$ \EdgeAssoc{$\bar{\mathbf{T}}_{w,k}$, $\mathcal{P}_{\mathcal{E},k}^{\prime}$, $\mathcal{M}_{E,k}$}\; \label{L:11} 

      $\mathbf{r}_{\mathcal{R}}, \mathbf{J}_{\mathcal{R}} \gets$ \ReflectAssoc{$\bar{\mathbf{T}}_{w,k}$, $\mathcal{P}_{\mathcal{R},k}^{\prime}$, $\mathcal{M}_{R,k}$}\; \label{L:12} 

      \tcp{\footnotesize{Update the estimated pose by the nonlinear optimization}}
      $\tilde{\mathbf{T}}_{w, k} \gets$ \MultiMetricNLO{$\mathbf{J}, \mathbf{r}$}\; \label{L:13} 

      \tcp{\footnotesize{Convergency}}
      $convergency \gets$ \ConvergCond{$\tilde{\mathbf{T}}_{w, k} \cdot \bar{\mathbf{T}}_{w,k}^{-1}$}\;

      \If{$convergency$}{
        \bf{break}\;
      }
    }

    \tcp{\footnotesize{Update parameters}}
    ${\mathbf{T}}_{w,k} \gets \tilde{\mathbf{T}}_{w, k}$\;
    $\mathbf{T}_{w^{\prime},w}^{-1} = {\mathbf{T}}_{w,k} \cdot \bar{\mathbf{T}}_{w,k}$\; \label{L:16}

    \tcp{\footnotesize{Update the local feature map}}
    \DownsizeFilter{$\mathcal{P}_{\mathcal{G},k}$, $\mathcal{P}_{\mathcal{F},k}$, $\mathcal{P}_{\mathcal{E},k}$, $\mathcal{P}_{\mathcal{R},k}$}\;
    $\mathcal{P}_{s,k} \gets$ \DORFilter{$\mathcal{P}_{k}$, $\mathcal{M}_{k}$, ${\mathcal{D}}_{k}$, ${\mathcal{L}}_{k}$, ${\mathbf{T}}_{w,k}$}\; \label{L:14} 
    \InsertAsKF{$\mathcal{P}_{s,k}, {\mathbf{T}}_{w,k}$}\; \label{L:15} 
  }
\end{algorithm}
One of the main tasks of LiDAR mapping thread is optimizing the estimated pose from the LO thread by the scan-to-map registration (green S2M block in Fig.\ref{FIG:2}). The other is managing the local static map (brown TVF and pink DOR blocks in Fig.\ref{FIG:2}). The pseudo code is shown in Algorithm \ref{ALG:2}.

\subsubsection{\bf{Local feature map construction}}
In this paper, the pose-based local feature map construction scheme is applied. In particular, the pose prediction $\mathbf{\bar{T}}_{w,k}$ is calculated by E.q.(\ref{EQ:18}) under the assumption that the drift between $\Delta \mathbf{T}_k$ and $\Delta \mathbf{T}_{k-1}$ is tiny (Line \ref{L:6}). Feature points scanned in the vicinity of $\mathbf{\bar{T}}_{w,k}$ are merged (Line \ref{L:7}) and filtered (Line \ref{L:8}) to construct the local map $\mathcal{M}_k$. Let $\Gamma(\cdot)$ denotes the filter, and $n \in N$ denotes timestamps of surrounding scans. The local map is built by: 
\begin{equation} \label{EQ:19}
  \mathcal{M}_k = \Gamma \left( \sum_{n \in N} \mathbf{T}_{w,n} \cdot \mathcal{P}_{s,n} \right).
\end{equation}

The conventional voxel-based downsample filter voxelizes the point cloud and retains one point for each voxel. The coordinate of retained point is averaged by all points in the same voxel. However, for the point intensity, averaging may cause the loss of similarity between consecutive scans. To maintain the local characteristic of the point intensity, we utilize the temporal information to improve the voxel-based downsample filter. In the TVF, a temporal window is set for the intensity average. Specifically, the coordinate of the downsampled point is still the mean of all points in the voxel, but the intensity is the mean of points in the temporal window, i.e., $\vert t_k - t_n \vert < Th_t$, where $t_k$ and $t_n$ represent timestamps of the current scan and selected point, respectively. 

\subsubsection{\bf{Mapping update}}
The categorized features are jointly registered with feature maps in the same way as in the LiDAR odometry module. The low-drift pose transform $\mathbf{T}_{w, k}$ can be estimated by scan-to-map alignment (Line \ref{L:9}-\ref{L:13}). Since the distribution of feature points in the local map is disordered, point neighbors cannot be directly indexed through the scan line number. Accordingly, the K-D tree is utilized for nearest points searching, and the PCA algorithm calculates norms and primary directions of neighbouring points.  

Finally, the obtained $\mathbf{T}_{w, k}$ is fed to the DOR filter to filter out dynamic points in the current scan. Only static points $\mathcal{P}_{s,k}$ are retained in the local feature map list (Line \ref{L:14}-\ref{L:15}). Moreover, the odometry reference drift is also updated by E.q.(\ref{EQ:18}), i.e. $\Delta \mathbf{T}_{k} = \mathbf{T}_{w, k}\mathbf{T}_{w^{\prime}, k}^{-1}$ (Line \ref{L:16}).

\section{Experiments} \label{Sect:4}
In this section, the proposed InTEn-LOAM is evaluated qualitatively and quantitatively on both simulated and real-world datasets, covering various outdoor scenes. We first test the feasibility of each functional module, including the feature extraction module, intensity-based scan registration, and dynamic points removal. Then we conduct a comprehensive evaluation for InTEn-LOAM in terms of positioning accuracy and constructed map quality. During experiments, the system processing LiDAR scans runs on a laptop computer with 1.8GHz quad cores and 4Gib memory, on top of the robot operating system (ROS) in Linux. 

\begin{figure}[ht] 
  \centering
  \subfloat[]{
    \centering  
    \begin{minipage}[b]{0.24\linewidth}
      \includegraphics[height=5.4cm]{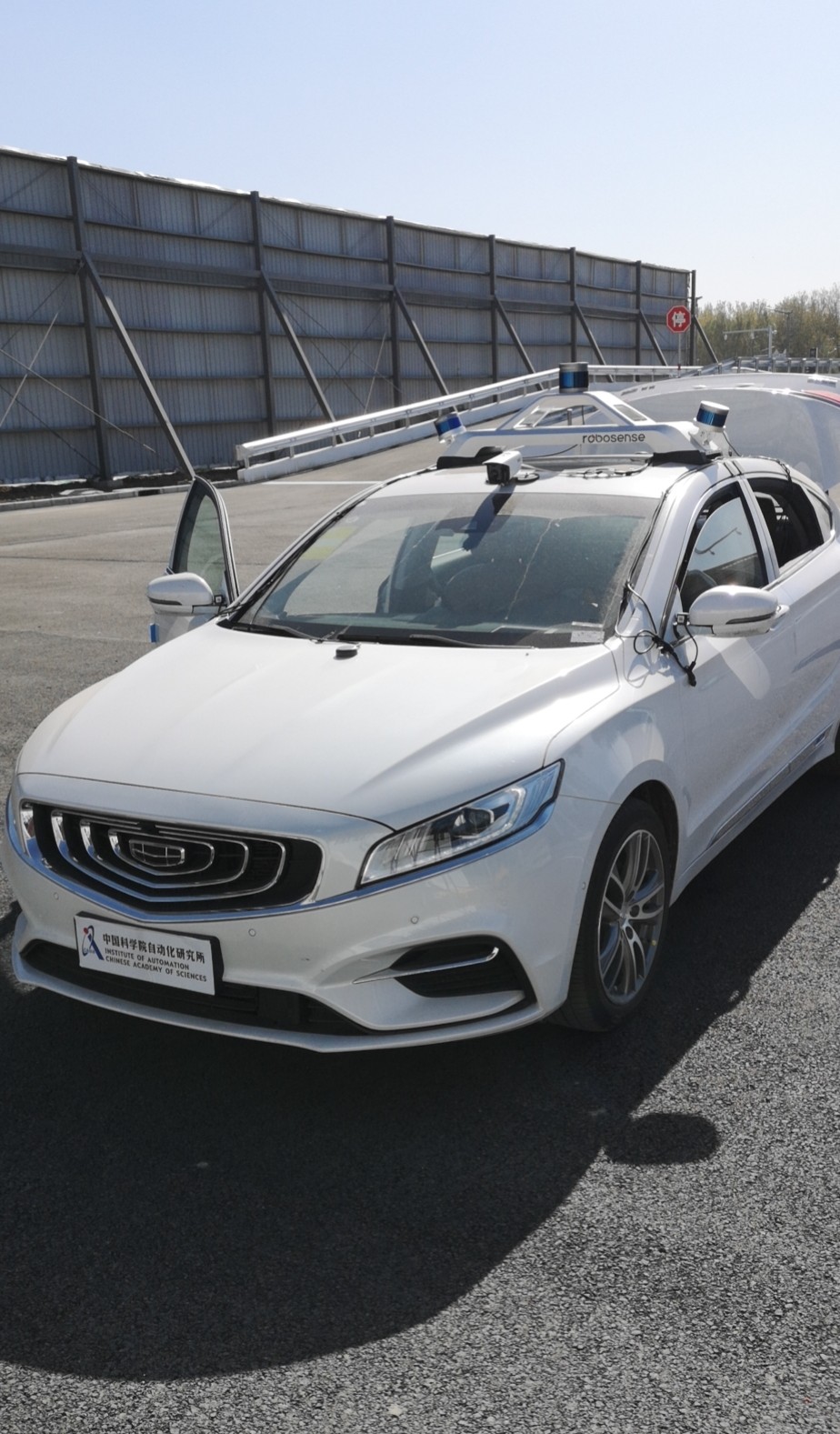}
    \end{minipage}}
  \subfloat[]{
    \centering  
    \begin{minipage}[b]{0.52\linewidth}
      \includegraphics[height=2.54cm]{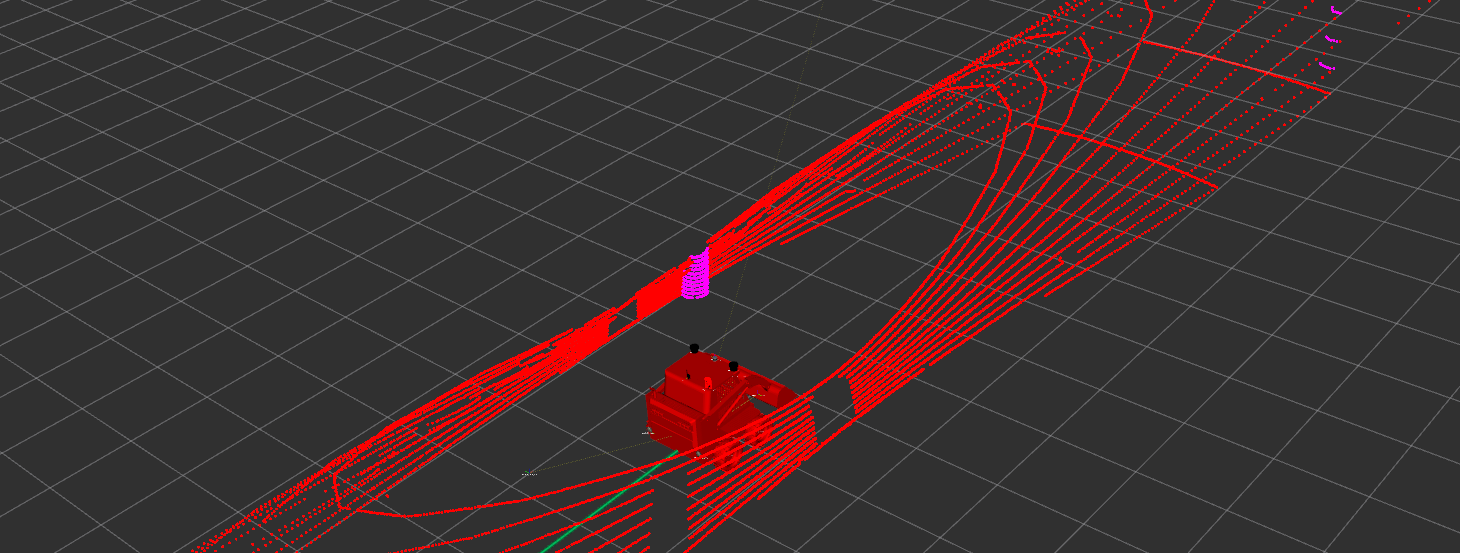}\vspace{0.1cm}

      \includegraphics[height=2.7cm]{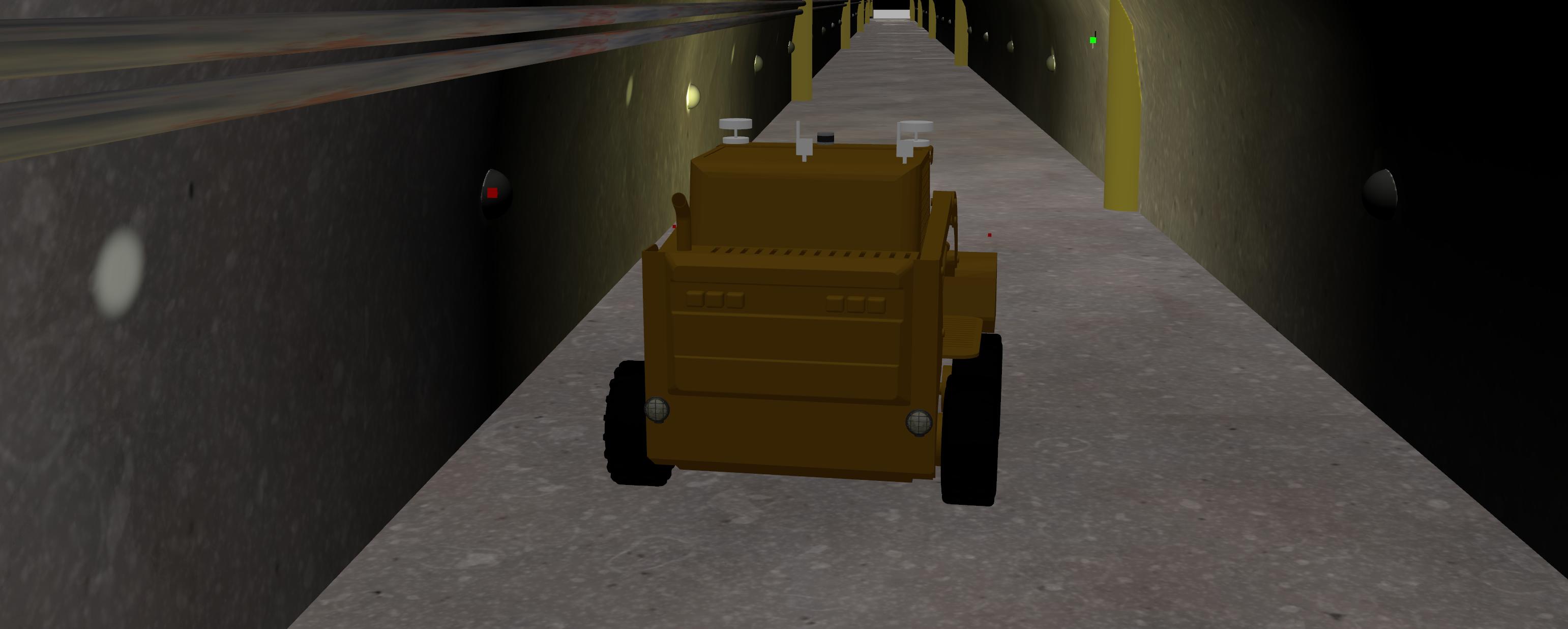}
    \end{minipage}}\vspace{-0.1cm}	
  \caption{Dataset sampling platform. (a) Autonomous driving car; (b) Simulated mine 
  car and scan example. \textcolor{magenta}{Magenta} laser points are reflected from 
  \textcolor{brown}{brown} signs in the simulated environemnts }\label{FIG:8}	
\end{figure}

The simulated test environment was built based on the challenging scene provided by the DARPA Subterranean (SubT) Challenge\footnote{https://github.com/osrf/subt}. We simulated a $1000m$ long straight mine tunnel (see Fig.\ref{FIG:8}(b)) with smooth walls and reflective signs that are alternatively posted on both sides of the tunnel at $30m$ intervals. Physical parameters of the simulated car, such as ground friction, sensor temperature and humidity are consistent with reality to the greatest extent. A 16-scanline LiDAR is on the top of the car. Transform groundtruths were exported at $100Hz$. The real-world dataset was collected by an autonomous driving car with a 32-scanline LiDAR (see Fig.\ref{FIG:8}(a)) in the autonomous driving test field, where a $150m$ long straight tunnel is exsited. Moreover, the KITTI odometry benchmark\footnote{http://www.cvlibs.net/datasets/kitti/eval\_odometry.php} was also utilized to compare with other state-of-the-art LO solutions.

\subsection{Functional module test}
\subsubsection{\bf{Feature extraction module}}
We validated the feature extraction module on the real-world dataset. In the test, we set the edge feature extraction threshold as $Th_E = 0.3$, the facade feature extraction threshold as $Th_E = 0.1$, and the intensity difference threshold as $Th_{\Delta I} = 80$, and partitioned the intensity image into $16 \times 4$ blocks. 
\begin{figure}[ht] 
  \centering
  \subfloat[]
  {
    \begin{minipage}[b]{0.96\textwidth}
      \centering
      \includegraphics[width=\textwidth]{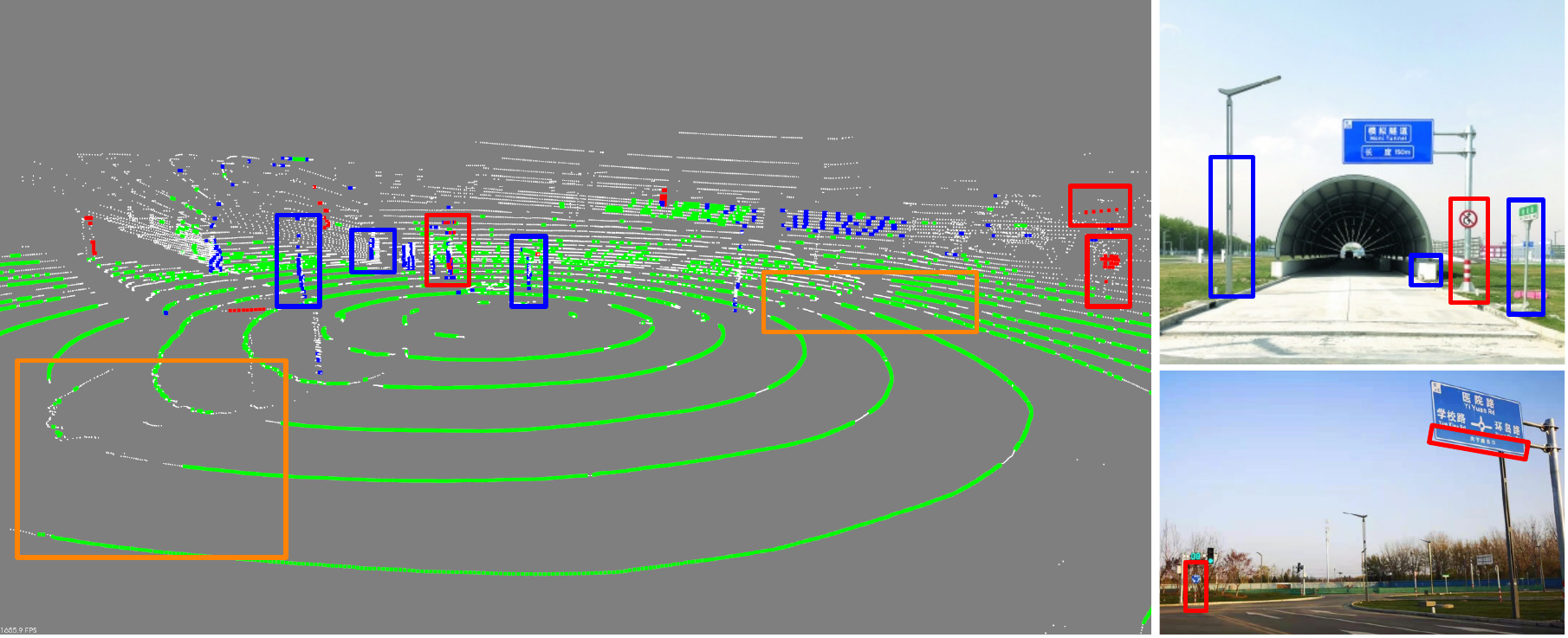}	
	  \end{minipage}
  }\vspace{-7.pt}	    

  \subfloat[]{
    \begin{minipage}[b]{0.96\textwidth}
	    \centering
	    \includegraphics[width=\textwidth]{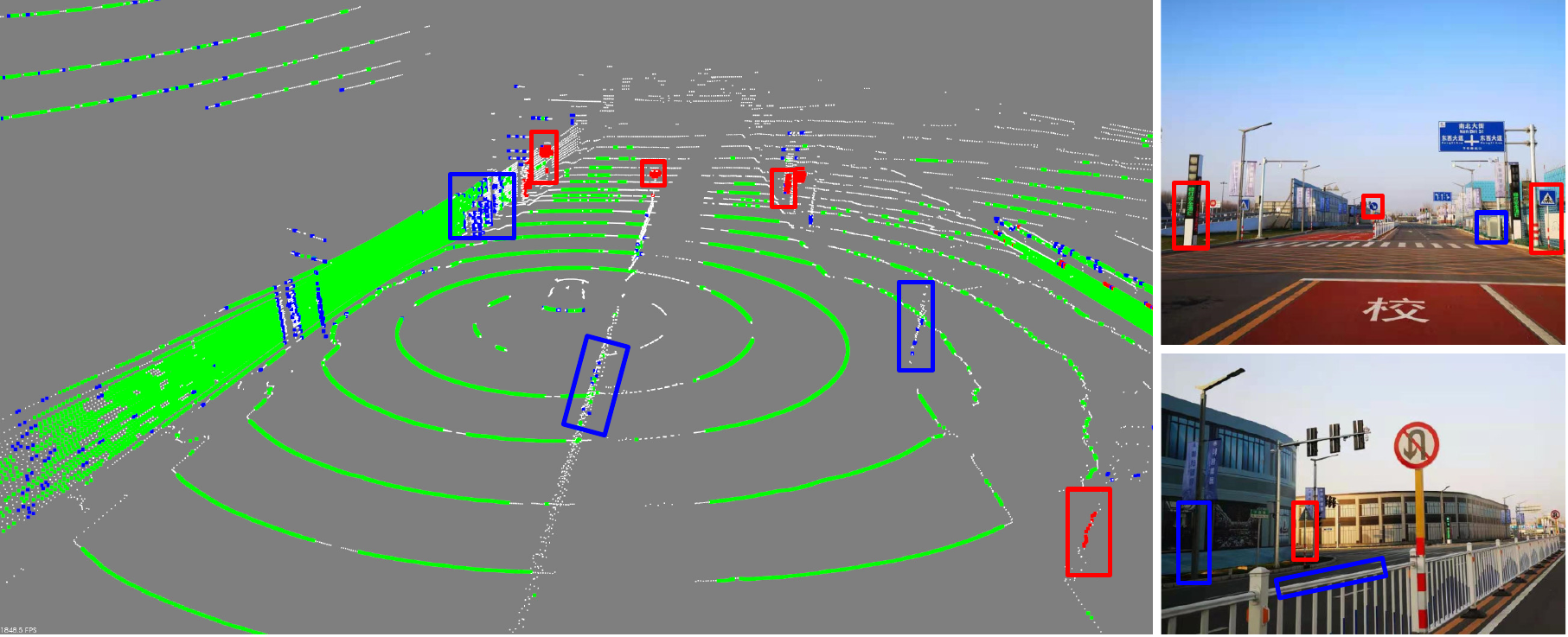}
	  \end{minipage}
  }\vspace{-7.pt}	

  \subfloat[]{
    \begin{minipage}[b]{0.96\textwidth}
	    \centering
	    \includegraphics[width=\textwidth]{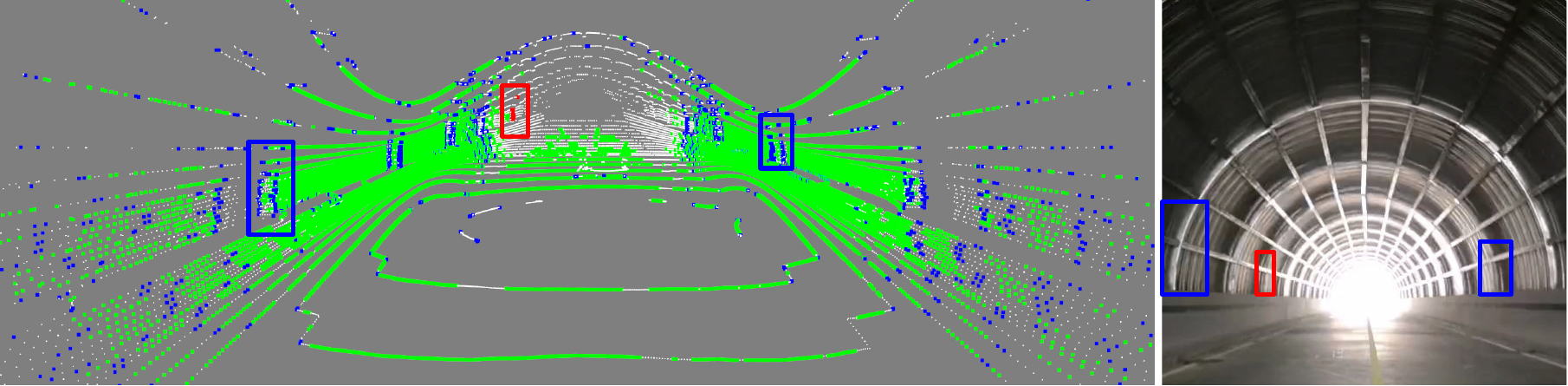}
	  \end{minipage}
  }\vspace{-7.pt}	

  \subfloat[]{
    \begin{minipage}[b]{0.96\textwidth}
	    \centering
	    \includegraphics[width=\textwidth]{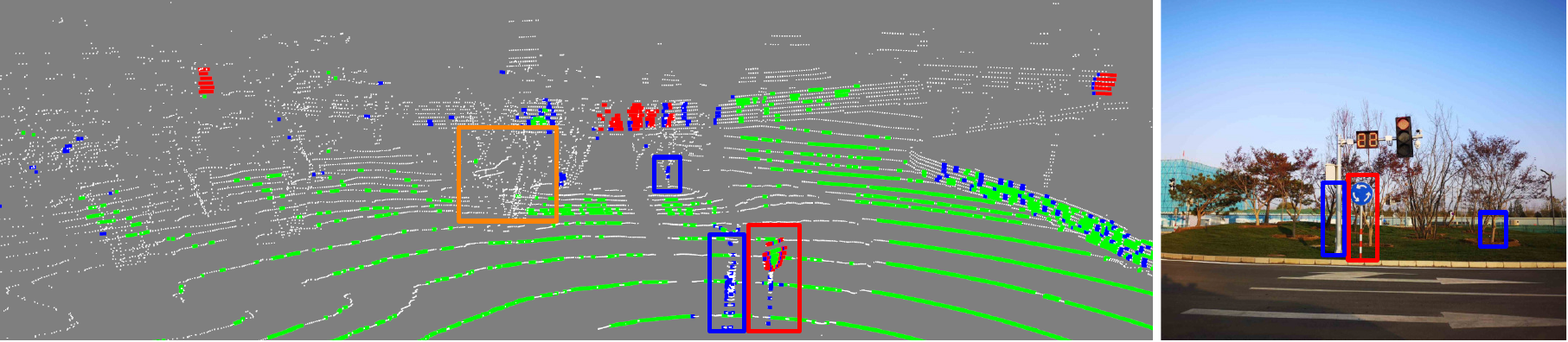}
	  \end{minipage}
  } 
  \caption{Feature extraction results in different scenes. (a) Open road; (b) City avenue; (c) Long straight tunnel; (d) Roadside green belt. (\textcolor{Green}{plane}, \textcolor{red}{reflector}, \textcolor{blue}{edge} and \textcolor{gray}{raw scan} points). Objects in the real-world scenes and their counterparts in laser scans are circled by boxes (\textcolor{red}{reflector features}, \textcolor{blue}{edge features}, \textcolor{orange}{some special areas}).}\label{FIG:7}
\end{figure}

Fig.\ref{FIG:7} shows feature extraction results. It can be seen that edges, planes, and reflectors can be correctly extracted in various road conditions. With the effect of ground segmentation, breakpoints on the ground (see orange box region in Fig.\ref{FIG:7}(a)) are correctly marked as plane, avoiding the issue that breakpoints are wrongly marked as edge features due to their large roughness values. In the urban city scene, conspicuous intensity features can be easily found, such as landmarks and traffic lights (see Fig.\ref{FIG:7}(b)). Though there are many plane features in the tunnel, few valid edge features can be extracted (see Fig.\ref{FIG:7}(c)). In addition, sparse and scattered plant points with large roughness values (see orange box region in Fig.\ref{FIG:7}(d)) are filtered as outliers with the help of the object clustering.

According to the above results, some conclusions can be drawn: (1) The number of plane features is always much greater than that of edge features, especially in open areas, which may cause the issue of constraint-unbalance during the multi-metric nonlinear optimization. (2) Static reflector features widely exist in real-world environments, which are useful for the feature-based scan alignment, and should not be ignored. (3) The adapative intensity feature extraction approach makes it possible to manually add reflective targets in feature-degraded environments.

\subsubsection{\bf{Intensity-based scan registration}}
\begin{figure}[ht] 
  \centering
  \subfloat[]{
    \begin{minipage}[b]{0.7\textwidth}
	    \centering
	    \includegraphics[width=\textwidth]{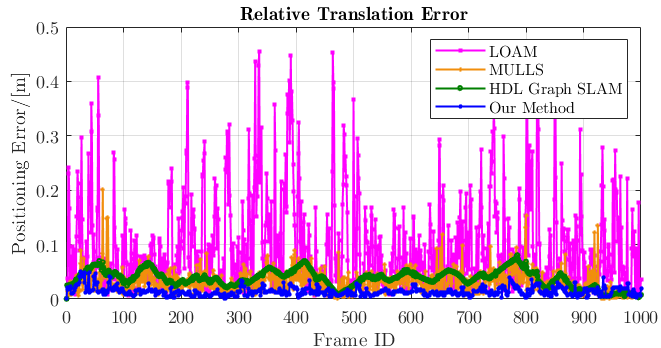}
	  \end{minipage}
  }\vspace{-7.pt}	

  \subfloat[]{
    \begin{minipage}[b]{0.7\textwidth}
	    \centering
	    \includegraphics[width=\textwidth]{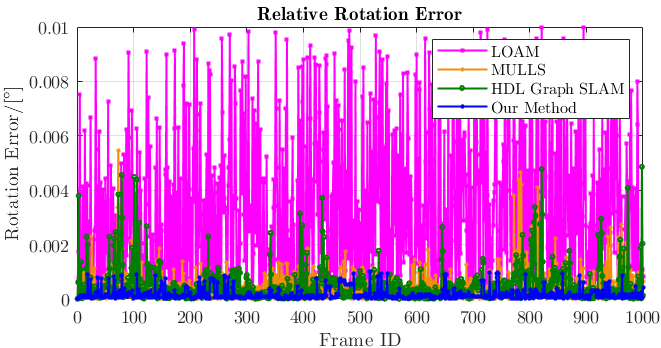}
	  \end{minipage}
  } 
  \caption{Relative error plots. (a) Relative translation error curves; (b) Relative  rotation error curves. (\textcolor{green}{NDT} of HDL-Graph-SLAM, \textcolor{magenta}{feature-based registration} approach of LOAM, the proposed \textcolor{blue}{intensity-based registration} approach).}\label{FIG:9}
\end{figure}

We validated the intensity-based scan registration method on the simulated dataset. To highlight the role of intensity-based scan registration, we quantitatively evaluated the relative accuracy of the proposed method and compared the result with prevalent geometric-based scan registration methods, i.e., edge and surface feature registration of LOAM\cite{zhang2017low}, multi-metric regisration of MULLS\cite{pan2021mulls} and NDT of HDL-Graph-SLAM\cite{koide2019portable}. The evaluation used the simulated tunnel dataset, which is a typical geometric-degraded environment. The measure used to evaluate the accuracy of scan registration is the relative transformation error. In particular, differences between the groundtruth $\mathbf{T}_{k+1,k}^{GT}$ and the estimated relative transformation $\mathbf{T}_{k+1,k}$ are calculated and represented as an error vector, i.e., $\mathbf{r}_{k} = \text{vec}( \mathbf{T}_{k+1,k}^{GT} \cdot \mathbf{T}_{k+1,k}^{-1})$. The norms of translational and rotational parts of $\mathbf{r}_{k}$ are illustrated in Fig.\ref{FIG:9}. Note that the result of intensity-based registration only utilizes measurements from the intensity channel of laser scan, instead of all information including range, bearing and intensity of laser scan. 
The figures show that all four rotation errors of different approaches are less than $0.01^{\circ}$, while errors of InTEn-LOAM and MULLS are less than $0.001^{\circ}$. It demonstrates that laser points from the tunnel wall and ground enable to provide sufficient geometric constraints for the accuracy of relative attitude estimation. However, there are significant differences in relative translation errors (RTE). The intensity-based scan registration achieves the best RTE (less than $0.02m$), which is much better than the feature-based of LOAM and NDT of HDL-Graph-SLAM ($0.4m$ and $0.1m$), and better than the intensity-based weighting of MULLS($0.05$). The result proves the correctness and feasibility of the proposed intensity-based approach under the premise of sufficient intensity features. It also reflects the necessity of fusing reflectance information of points in poorly structured environments.

\subsubsection{\bf{Dynamic object removal}}
We validated the DOR module on Seq.07 and 10 of the KITTI odometry dataset. The test result was evaluated by qualitative evaluation method, i.e., marking dynamic points for each scan frame and qualitatively judging the accuracy of the dynamic object segmentation according to the actual targets in the real world the dynamic points correspond to. 
\begin{figure}[hp] 
  \centering
  \subfloat[]{
    \begin{minipage}[b]{0.98\textwidth}
	    \centering
	    \includegraphics[width=0.48\textwidth]{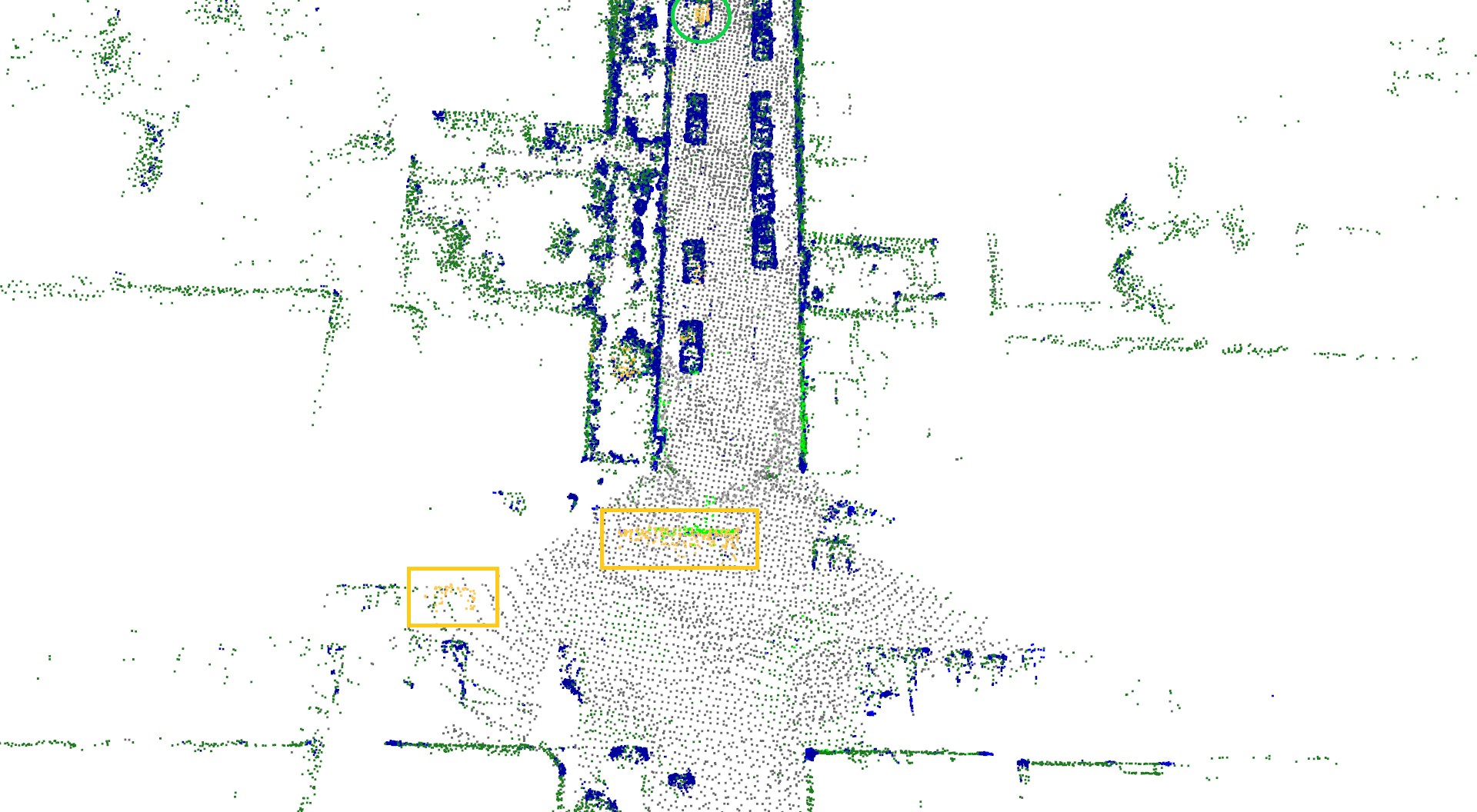}
      \includegraphics[width=0.48\textwidth]{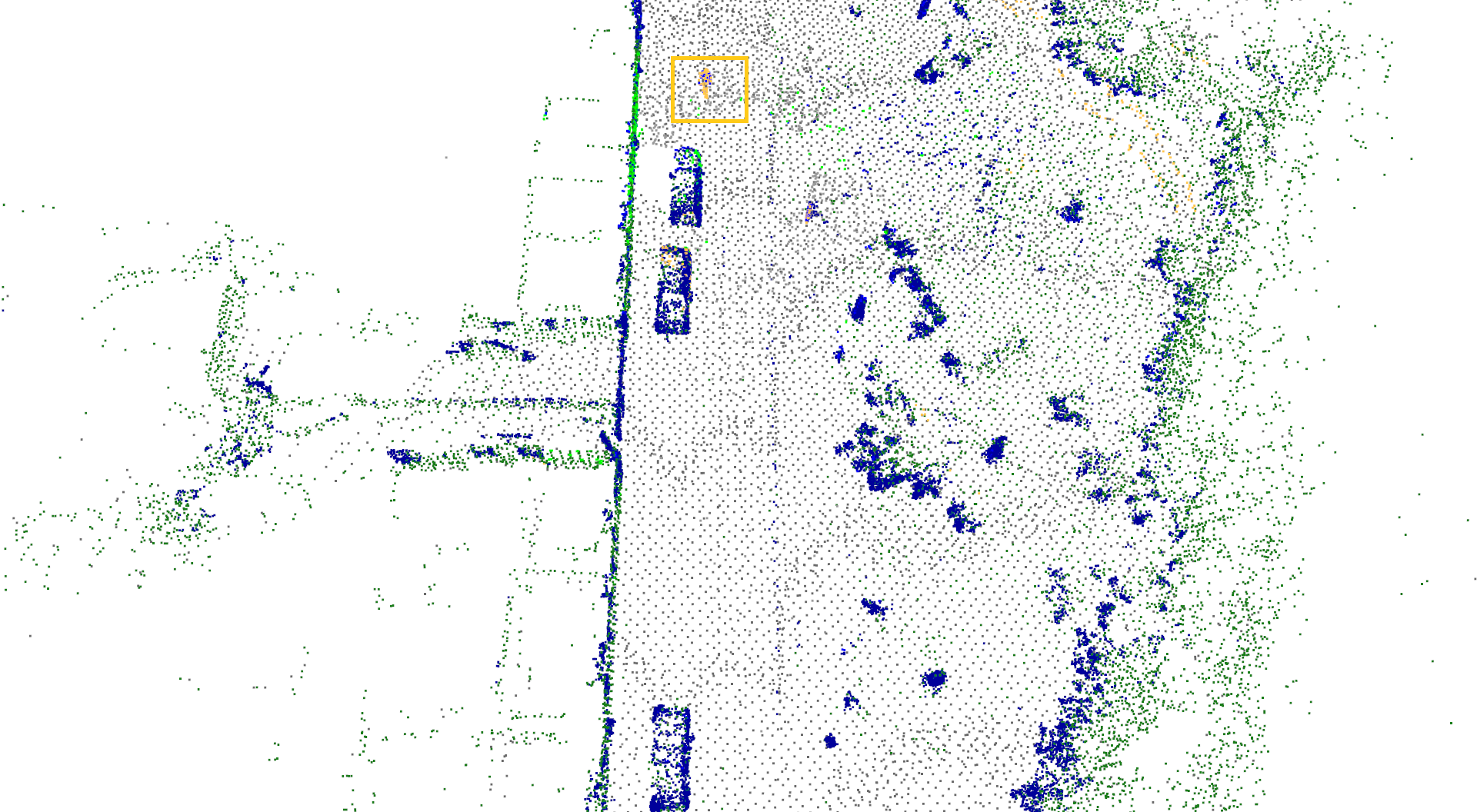}
	  \end{minipage}
  }\vspace{-5.pt}	

  \subfloat[]{
    \begin{minipage}[b]{0.98\textwidth}
	    \centering
	    \includegraphics[width=0.48\textwidth]{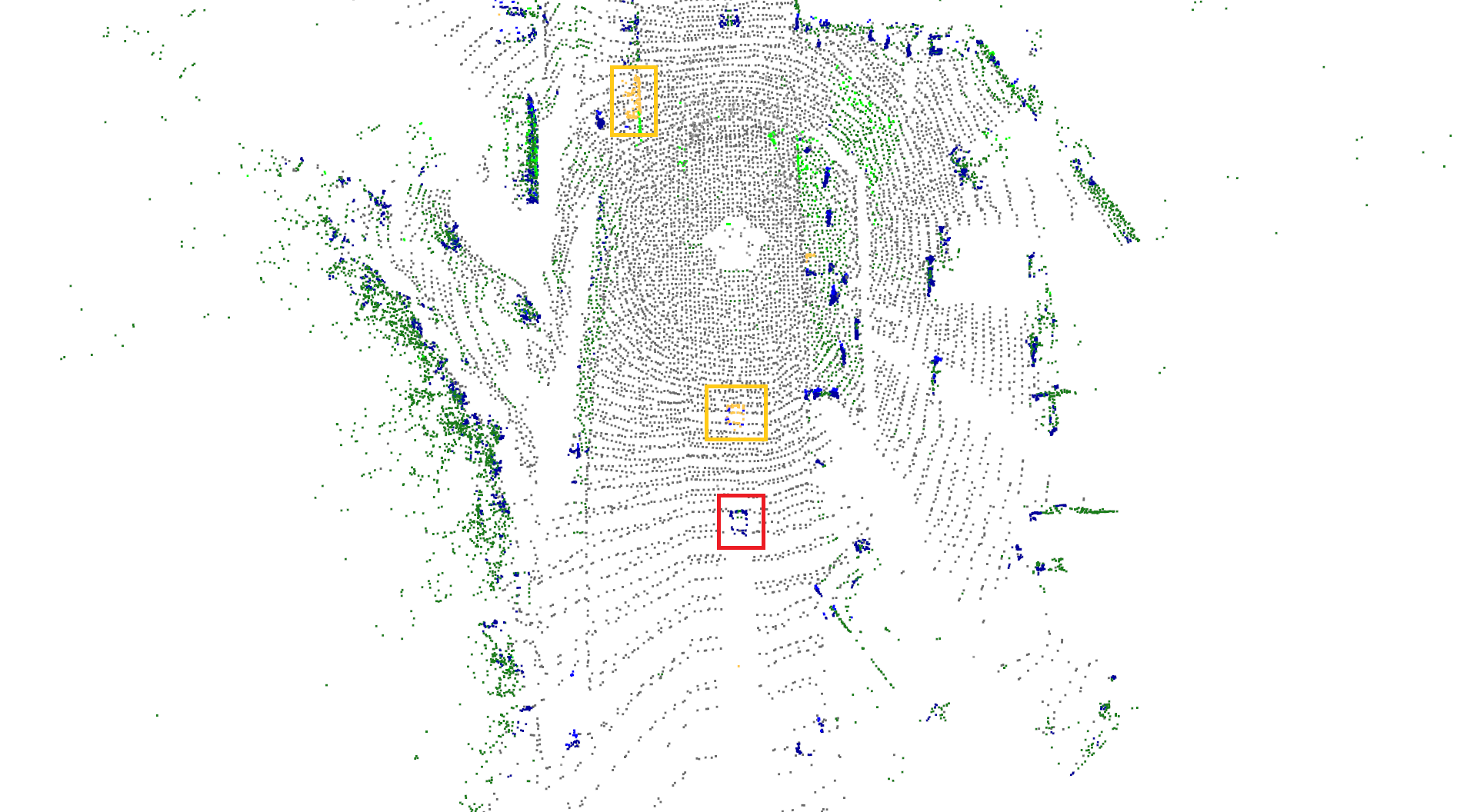}
      \includegraphics[width=0.48\textwidth]{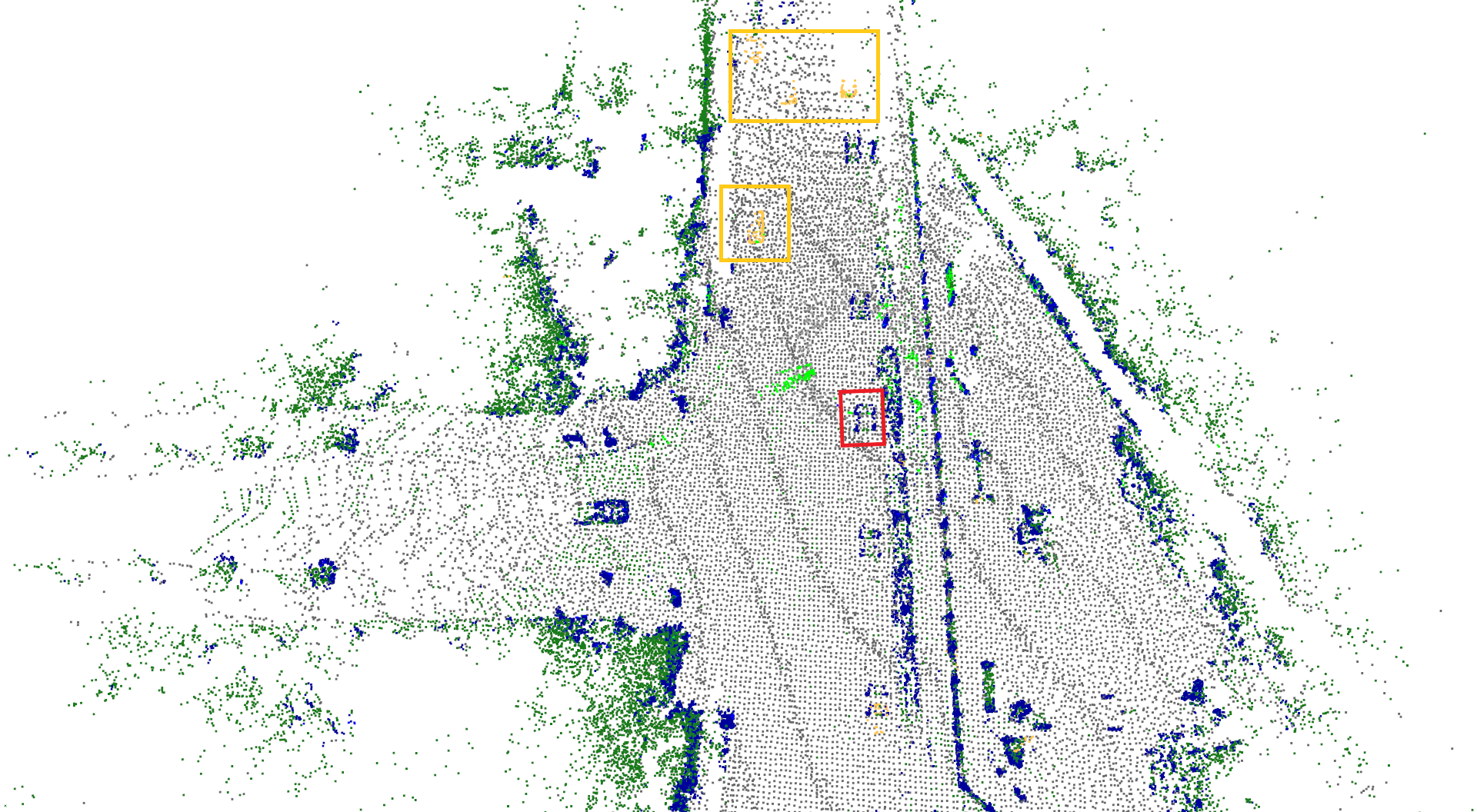}
	  \end{minipage}
  } 
  \caption{DOR examples for a single frame of laser scan. (a) Seq.07. Vehicles crossing the intersection when the data collection vehicle stops and waits for the traffic light (top); The cyclist traveling in the opposite direction when the data collection vehicle driving along the road (bottom). (b) Seq.10. Followers behind the data collection vehicle as it travels down the highway at high speed (top); Vehicles driving in the opposite direction and in front of the data collection vehicle when it slows down (bottom). (\textcolor{green}{facade}, \textcolor{gray}{ground}, \textcolor{blue}{edge} and \textcolor{orange}{dynamics} for points \textcolor{orange}{true positive}, \textcolor{red}{false positive} and \textcolor{green}{true negative} for dynamic segmentation boxes.)
  }\label{FIG:10}
\end{figure}
\begin{figure}[hp] 
  \centering
  \subfloat[]{
    \begin{minipage}[b]{0.96\textwidth}
	    \centering
	    \includegraphics[width=0.48\textwidth]{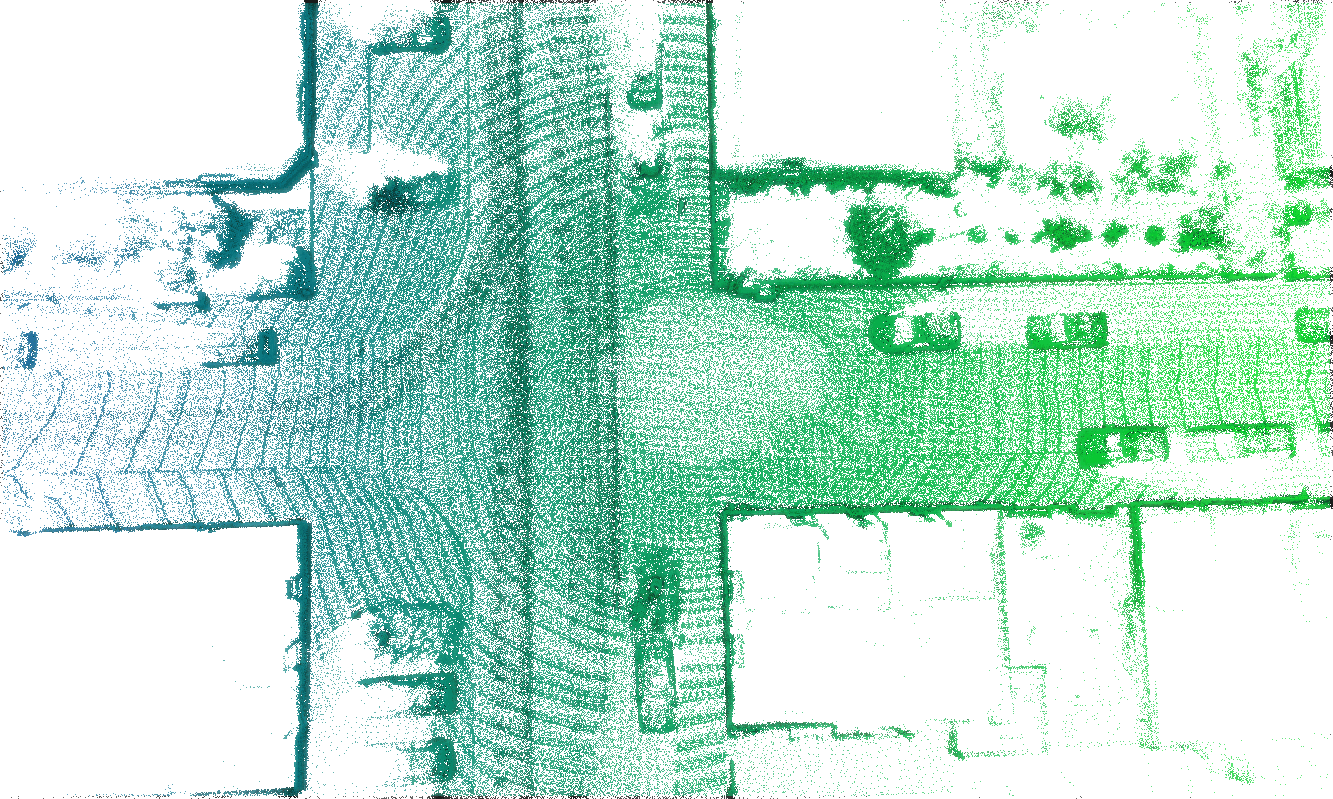}\vspace{2.pt}
      \includegraphics[width=0.48\textwidth]{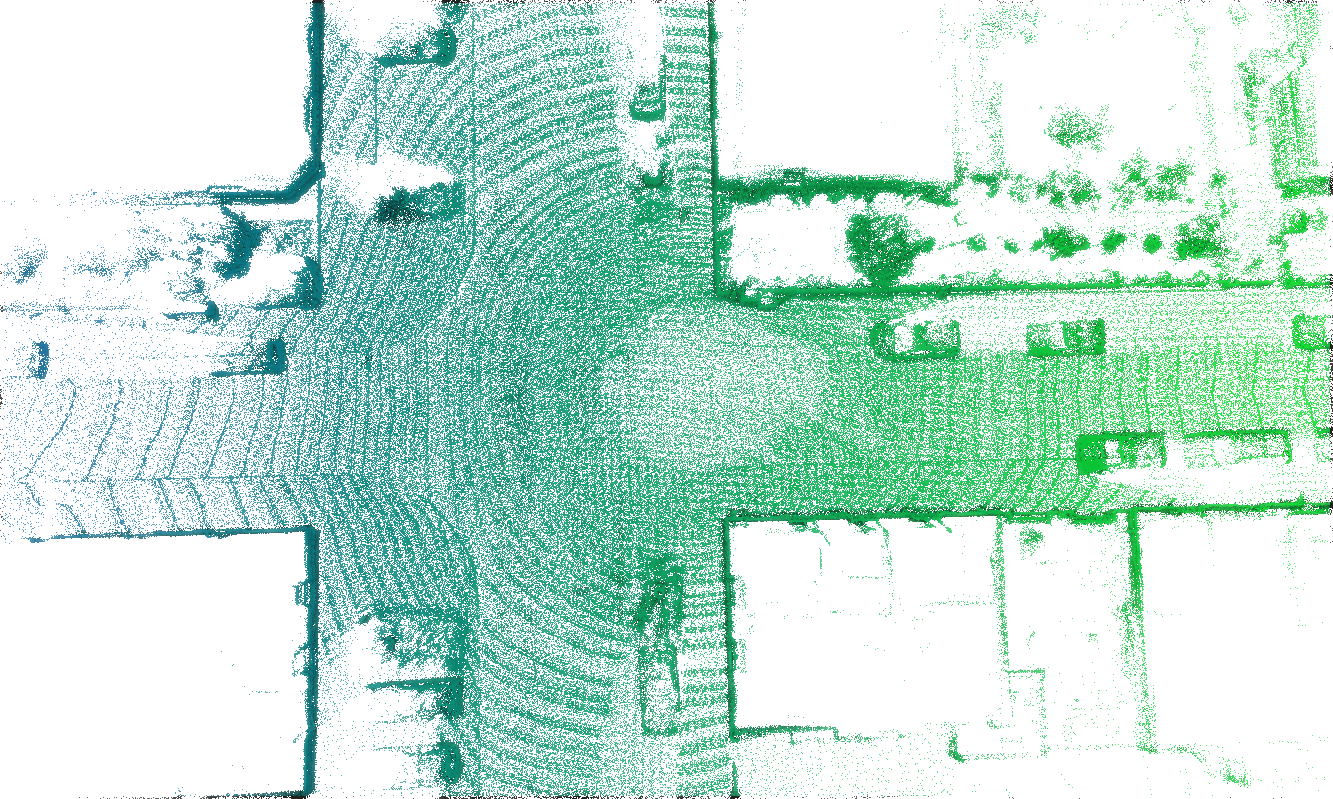}
	  \end{minipage}
  }\vspace{-5.pt}	

  \subfloat[]{
    \begin{minipage}[b]{0.96\textwidth}
	    \centering
	    \includegraphics[width=0.48\textwidth]{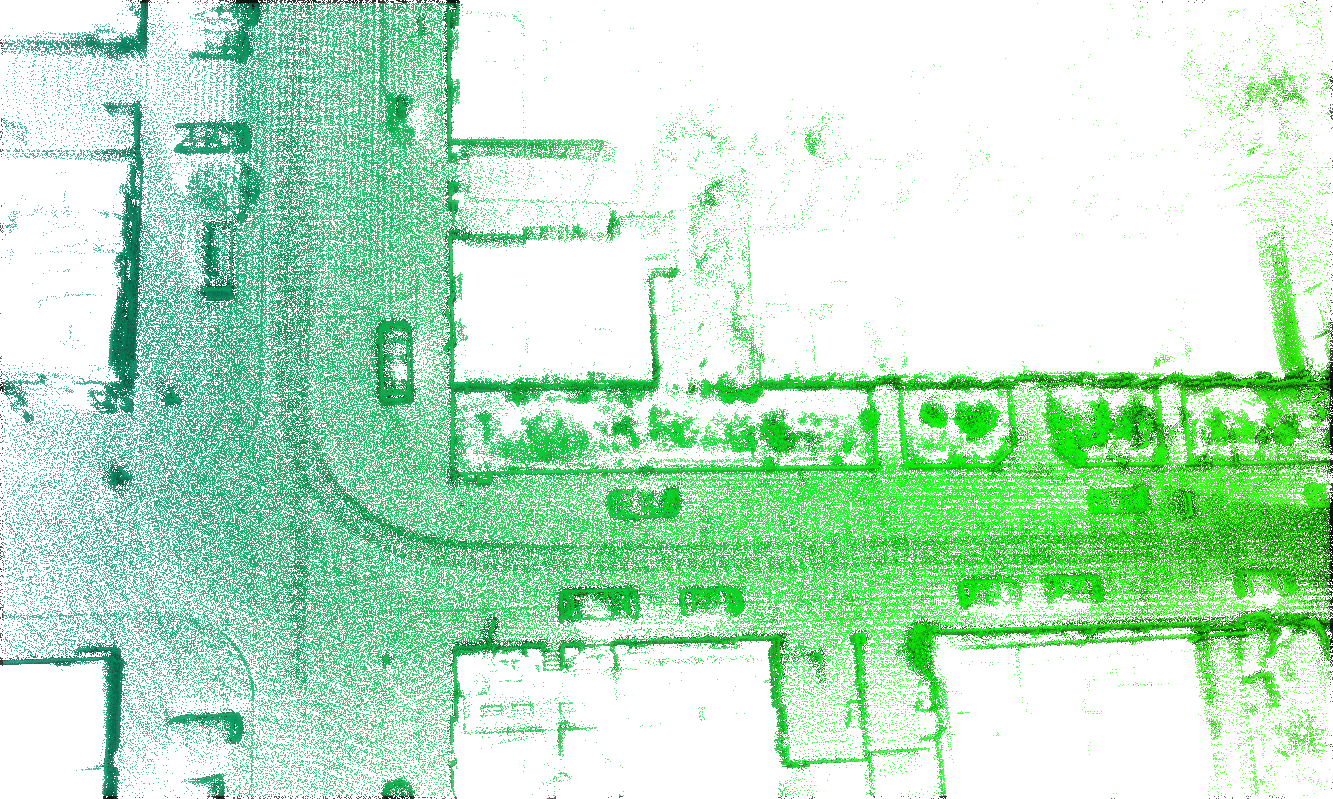}\vspace{2.pt}
      \includegraphics[width=0.48\textwidth]{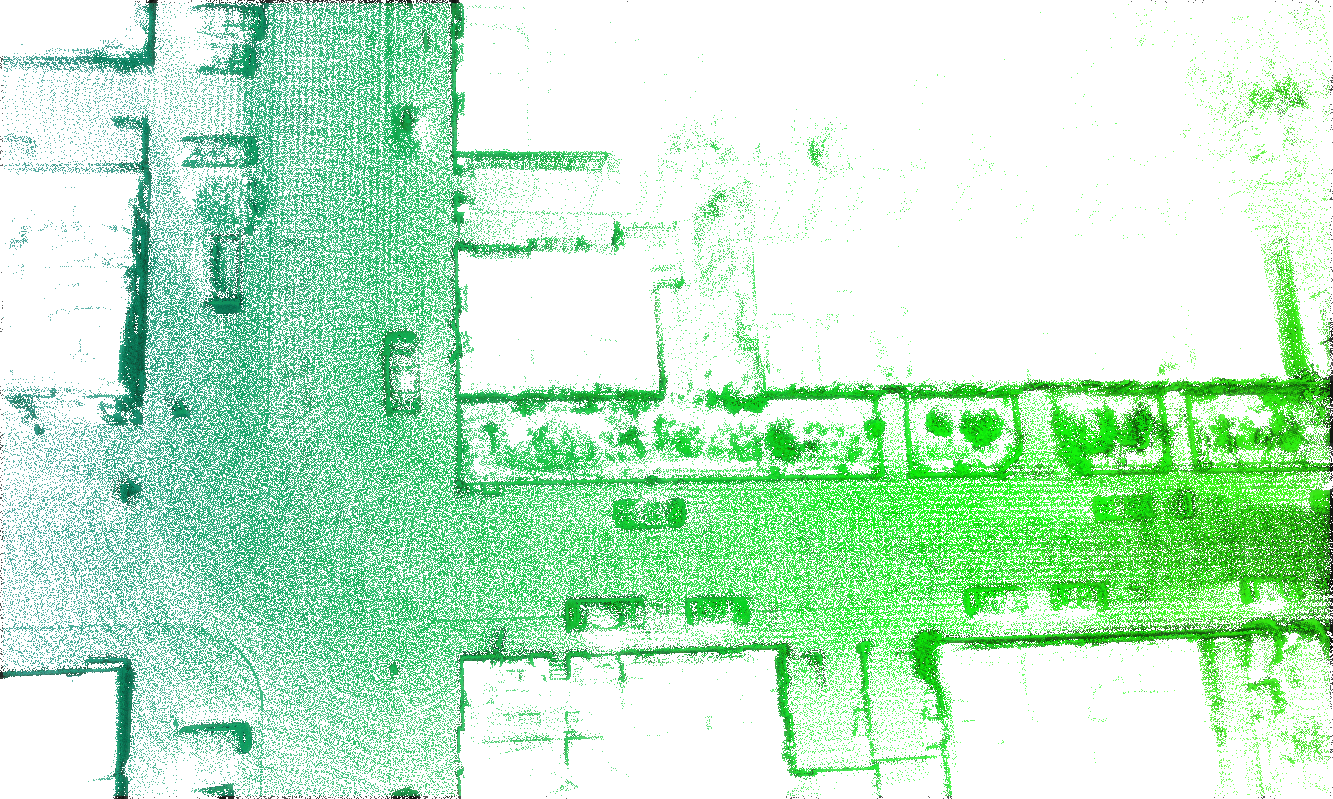}
	  \end{minipage}
  } 
  \caption{The comparison between local maps of LOAM and InTEn-LOAM. (a) Map at the intersection (b) Map at the busy road. Iu each subfigure, the top represents the map of LOAM w/o DOR, while the bottom represents the map of InTEn-LOAM w/ DOR.}\label{FIG:11}
\end{figure}

Fig.\ref{FIG:10} exhibits DOR examples for a single frame of laser scan at typical urban driving scenes. It can be seen that dynamic objects, such as vehicles crossing the intersection, vehicles and pedestrians traveling in front of/behind the data collection car, can be correctly segmented by the proposed DOR approach no matter the sampling vehicle is stationary or in motion. Fig.\ref{FIG:11} shows constructed maps at two representative areas, i.e., intersection and busy road. Maps were incrementally built by LOAM (without DOR) and InTEn-LOAM (with DOR) method. We can figure out from the figure that the map built by InTEn-LOAM is better since the DOR module effectively filters out dynamic points to help to accumulate a purely static points map. In contrast, the map constructed by LOAM owns a large amount of 'ghost points' increasing the possibility of erroneous point matching.

In general, the above results prove that the DOR method proposed in this paper owns the ability to segment dynamic objects for a scan frame correctly. However, it also has some shortcomings. For instance, (1) The proposed comparison-based DOR filter is sensitive to the quality of laser scan and the density of the local points map, causing the omission or mis-marking of some dynamic points (see the green circle box in the top of Fig.\ref{FIG:10}(a) and the red rectangle box in the bottom of Fig.\ref{FIG:10}(b)); (2) Dynamic points in the first frame of scan cannot be marked using the proposed approach (see the red rectangle box in the top of Fig.\ref{FIG:10}(b)). 
\subsection{Pose transform estimation accuracy} 
\begin{table*}[thp]
  \Large
  \centering
  \caption{Quantitative evaluation and comparison on KITTI dataset.} \vspace{-0.2cm}
  \label{TAB:1}
  \resizebox{\textwidth}{!}{%
  \begin{tabular}{cccccccccccccc}
  \multicolumn{14}{l}{
    \begin{tabular}[c]{@{}l@{}}{\textbf{Note:}} All errors are epresented as average RTE$[\%]$/RRE$[\circ/100m]$. \textcolor{red}{\textbf{Red}} and \textcolor{blue}{\textbf{blue}} fonts denote the first and second place, respectively. \\{\textbf{Denotations:}} U: Urban road; H: Highway; C: Country road;
    \end{tabular}
  } \\ \hline

  \multicolumn{1}{|c|}{Method} &
    \#00U &
    \#01H &
    \#02C &
    \#03C &
    \#04C &
    \#05C &
    \#06U &
    \#07U &
    \#08U &
    \#09C &
    \#10C &
    \multicolumn{1}{c|}{Avg.} &
    \multicolumn{1}{c|}{time$[s]$/frame} \\ \hline
   &
     &
     &
     &
     &
     &
     &
     &
     &
     &
     &
     &
     &
     \\ \hline
  \multicolumn{1}{|c|}{LOAM} &
    0.78/- &
    1.43/- &
    0.92/- &
    0.86/- &
    0.71/- &
    0.57/- &
    0.65/- &
    0.63/- &
    1.12/- &
    0.77/- &
    0.79/- &
    \multicolumn{1}{c|}{0.84/-} &
    \multicolumn{1}{c|}{0.10} \\ \hline
  \multicolumn{1}{|c|}{IMLS-SLAM} &
    \textcolor{red}{\textbf{0.50}}/- &
    0.82/- &
    \textcolor{red}{\textbf{0.53}}/- &
    0.68/- &
    \textcolor{red}{\textbf{0.33}}/- &
    0.32/- &
    0.33/- &
    \textcolor{blue}{\textbf{0.33}}/- &
    \textcolor{blue}{\textbf{0.80}}/- &
    0.55/- &
    0.53/- &
    \multicolumn{1}{c|}{0.57/-} &
    \multicolumn{1}{c|}{1.25} \\ \hline
  \multicolumn{1}{|c|}{MC2SLAM} &
    \textcolor{blue}{\textbf{0.51}}/- &
    0.79/- &
    \textcolor{blue}{\textbf{0.54}}/- &
    0.65/- &
    0.44/- &
    \textcolor{red}{\textbf{0.27}}/- &
    0.31/- &
    0.34/- &
    0.84/- &
    0.46/- &
    \textcolor{blue}{\textbf{0.52}}/- &
    \multicolumn{1}{c|}{0.56/-} &
    \multicolumn{1}{c|}{0.10} \\ \hline
  \multicolumn{1}{|c|}{SuMa} &
    0.70/0.30 &
    1.70/0.50 &
    1.10/0.40 &
    0.70/0.50 &
    0.40/\textcolor{blue}{\textbf{0.30}} &
    0.40/\textcolor{blue}{\textbf{0.20}} &
    0.50/\textcolor{blue}{\textbf{0.30}} &
    0.70/0.60 &
    1.00/0.40 &
    0.50/0.30 &
    0.70/0.30 &
    \multicolumn{1}{c|}{0.70/0.30} &
    \multicolumn{1}{c|}{0.07} \\ \hline
  \multicolumn{1}{|c|}{LO-Net} &
    0.78/0.42 &
    1.42/0.40 &
    1.01/0.45 &
    0.73/0.59 &
    0.56/0.54 &
    0.62/0.35 &
    0.55/0.35 &
    0.56/0.45 &
    1.08/0.43 &
    0.77/0.38 &
    0.92/0.41 &
    \multicolumn{1}{c|}{0.83/0.42} &
    \multicolumn{1}{c|}{0.10} \\ \hline
  \multicolumn{1}{|c|}{MULLS-LO} &
    \textcolor{blue}{\textbf{0.51}}/\textcolor{red}{\textbf{0.18}} &
    \textcolor{red}{\textbf{0.62}}/\textcolor{red}{\textbf{0.09}} &
    0.55/\textcolor{red}{\textbf{0.17}} &
    \textcolor{red}{\textbf{0.61}}/\textcolor{red}{\textbf{0.22}} &
    \textcolor{blue}{\textbf{0.35}}/\textcolor{red}{\textbf{0.08}} &
    \textcolor{blue}{\textbf{0.28}}/\textcolor{red}{\textbf{0.17}} &
    \textcolor{blue}{\textbf{0.24}}/\textcolor{red}{\textbf{0.11}} &
    \textcolor{red}{\textbf{0.29}}/\textcolor{red}{\textbf{0.18}} &
    \textcolor{blue}{\textbf{0.80}}/\textcolor{red}{\textbf{0.25}} &
    \textcolor{blue}{\textbf{0.49}}/\textcolor{red}{\textbf{0.15}} &
    0.61/\textcolor{red}{\textbf{0.19}} &
    \multicolumn{1}{c|}{\textcolor{red}{\textbf{0.49}}/\textcolor{red}{\textbf{0.16}}} &
    \multicolumn{1}{c|}{0.08} \\ \hline
  \multicolumn{1}{|c|}{\textbf{InTEn-LOAM}} &
    0.52/\textcolor{blue}{\textbf{0.21}} &
    \textcolor{blue}{\textbf{0.64}}/\textcolor{blue}{\textbf{0.35}} &
    \textcolor{blue}{\textbf{0.54}}/\textcolor{blue}{\textbf{0.28}} &
    \textcolor{blue}{\textbf{0.63}}/\textcolor{blue}{\textbf{0.33}} &
    0.37/0.31 &
    0.36/0.25 &
    \textcolor{red}{\textbf{0.24}}/\textcolor{red}{\textbf{0.11}} &
    0.34/\textcolor{blue}{\textbf{0.31}} &
    \textcolor{red}{\textbf{0.71}}/\textcolor{blue}{\textbf{0.29}} &
    \textcolor{red}{\textbf{0.48}}/\textcolor{blue}{\textbf{0.19}} &
    \textcolor{red}{\textbf{0.45}}/\textcolor{blue}{\textbf{0.21}} &
    \multicolumn{1}{c|}{\textcolor{blue}{\textbf{0.54}}/\textcolor{blue}{\textbf{0.26}}} &
    \multicolumn{1}{c|}{0.09} \\ \hline
  \end{tabular}%
  }
\end{table*}

\subsubsection{\bf{KITTI dataset}} 
\begin{figure}[hp] 
  \centering
  \subfloat[]{
    \begin{minipage}[b]{0.7\textwidth}
	    \centering
	    \includegraphics[width=\textwidth]{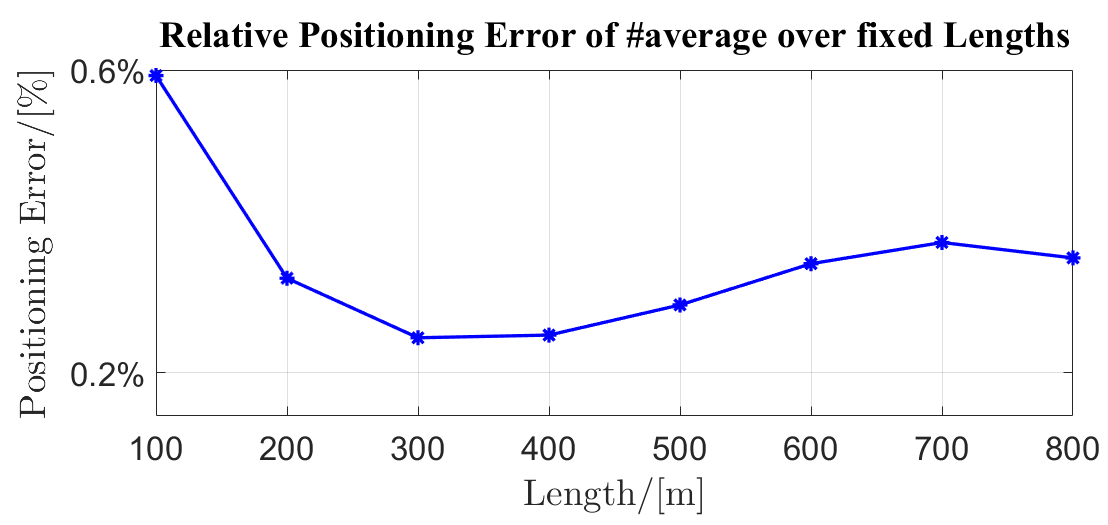}
	  \end{minipage}
  }\vspace{-7.pt}	

  \subfloat[]{
    \begin{minipage}[b]{0.7\textwidth}
	    \centering
	    \includegraphics[width=\textwidth]{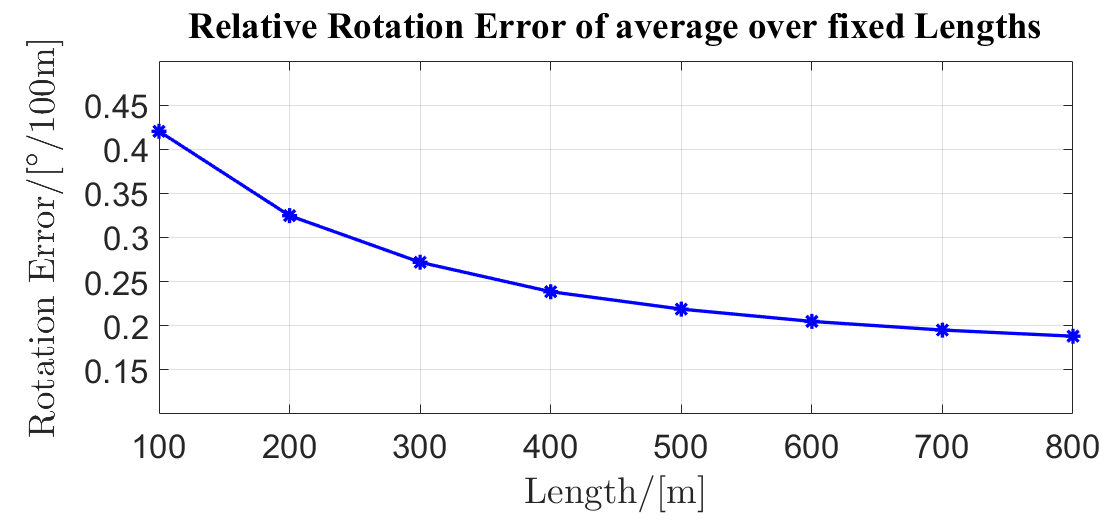}
	  \end{minipage}
  } 
  \caption{The average RTE and RRE of InTEn-LOAM over fixed lengths. (a) RTE; 
  (b) RRE.}\label{FIG:12}
\end{figure}

The quantitative evaluations were conducted on the KITTI odometry dataset, which is composed of 11 sequences of laser scans captured by a Velodyne HDL-64E LiDAR with GPS/INS groundtruth poses. We followed the odometry evaluation criterion from \cite{geiger2012we} and used the average relative translation and rotation errors (RTE and RRE) within a certain distance range for the accuracy evaluation. The performance of the proposed InTEn-LOAM, and other six state-of-the-art LiDAR odometry solutions whose results are taken from their original papers, are reported in Table.\ref{TAB:1}. Plots of average RTE/RRE over fixed lengths are exhibited in Fig.\ref{FIG:12}. Note that all comparison methods did not incorporate the loop closure module for more objective accuracy comparison. Moreover, an intrinsic angle correction of $0.2^{\circ}$ is applied to KITTI raw scan data for better performance \cite{pan2021mulls}.

Fig.\ref{FIG:12} demonstrates that accuracies in different length ranges are stable, and the maximums of average RTE and RRE are less than $0.32\%$ and $0.22^{\circ}/100m$. It also can be seen from the table that the average RTE and RRE of InTEn-LOAM are $0.54\%$ and $0.26^{\circ}/100m$, which outperforms the LOAM accuracy of $0.84\%$. The comprehensive comparison shows that InTEn-LOAM is superior or equal to the current state-of-the-art LO methods. Although the result of MULLS slightly better than that of InTEn-LOAM, the contribution of InTEn-LOAM is significant considering its excenllent performance in long straight tunnel with reflective markers. InTEn-LOAM costs around $90ms$ per frame of scan with about 3k and 30k feature points in the current scan points and local map points, respectively. Accordingly, the proposed LO method is able to operate faster than $10Hz$ on average for all KITTI odometry sequences and achieve real-time performance.
\begin{figure}[hp] 
  \centering
  \subfloat[]{
    \begin{minipage}[b]{0.6\textwidth}
	    \centering
	    \includegraphics[width=\textwidth]{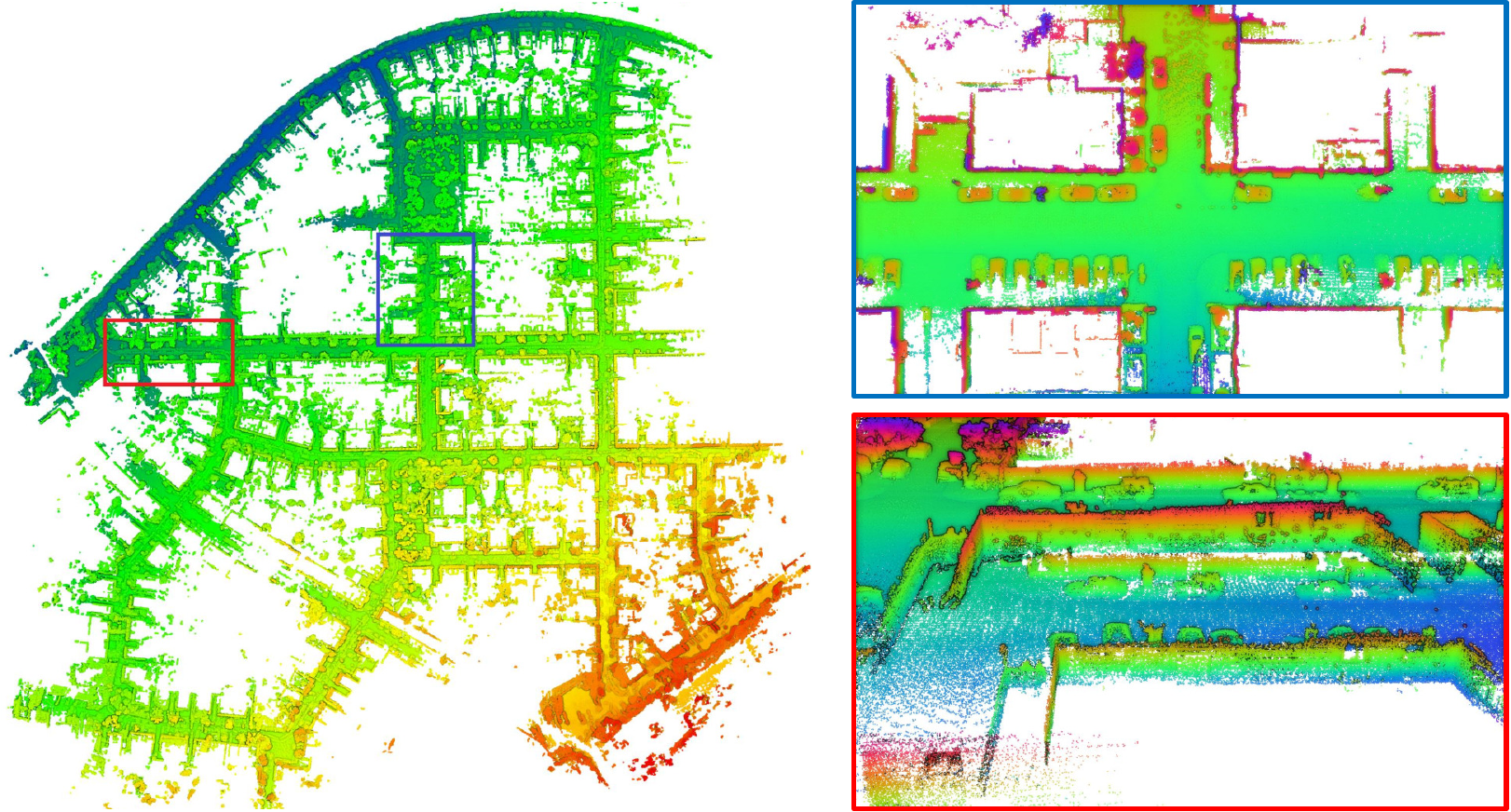}
	  \end{minipage}
  }
  \subfloat[]{
    \begin{minipage}[b]{0.36\textwidth}
	    \centering
	    \includegraphics[width=\textwidth]{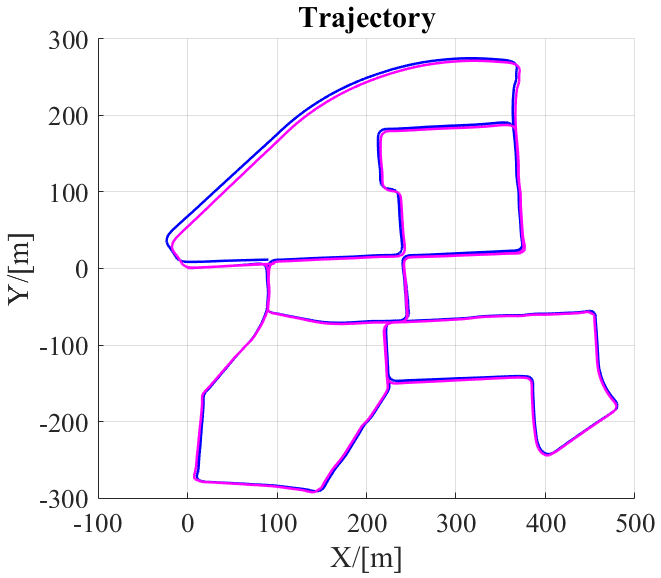}
	  \end{minipage}
  } 

  \subfloat[]{
    \begin{minipage}[b]{0.6\textwidth}
	    \centering
	    \includegraphics[width=\textwidth]{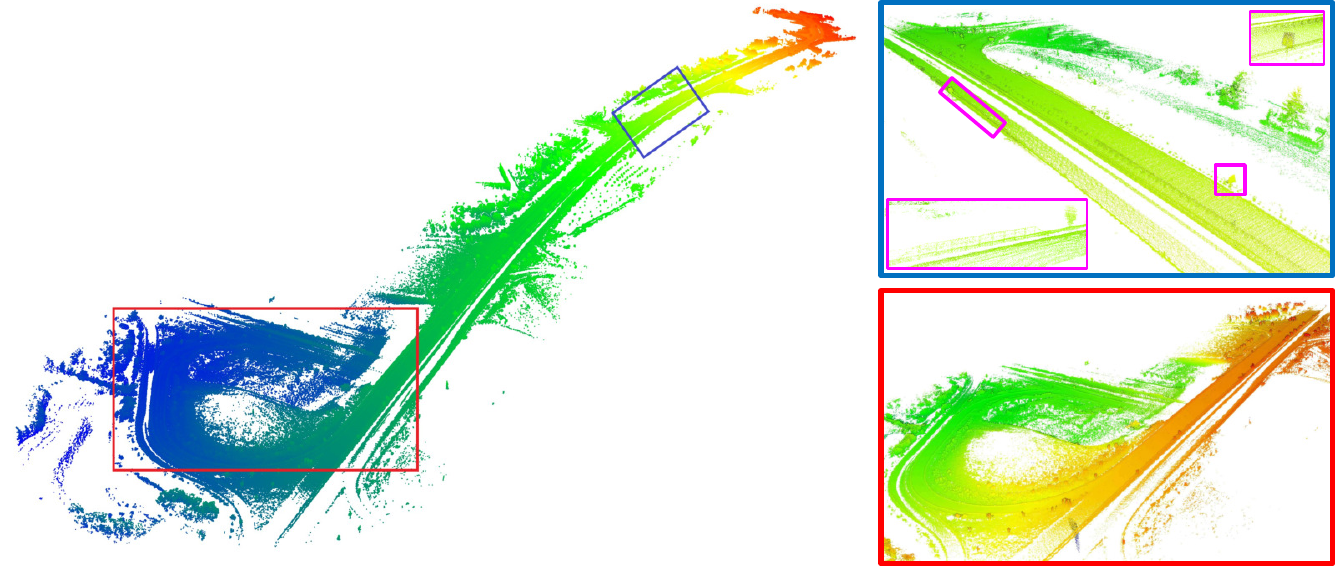}
	  \end{minipage}
  }
  \subfloat[]{
    \begin{minipage}[b]{0.36\textwidth}
	    \centering
	    \includegraphics[width=\textwidth]{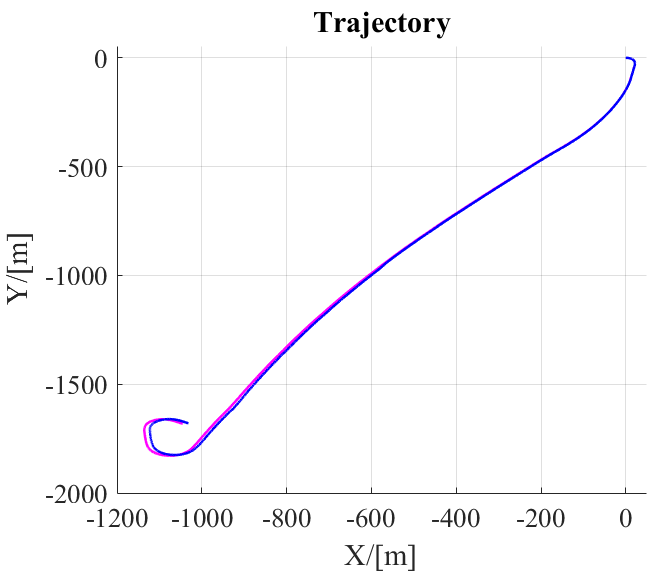}
	  \end{minipage}
  }

  \subfloat[]{
    \begin{minipage}[b]{0.6\textwidth}
	    \centering
	    \includegraphics[width=\textwidth]{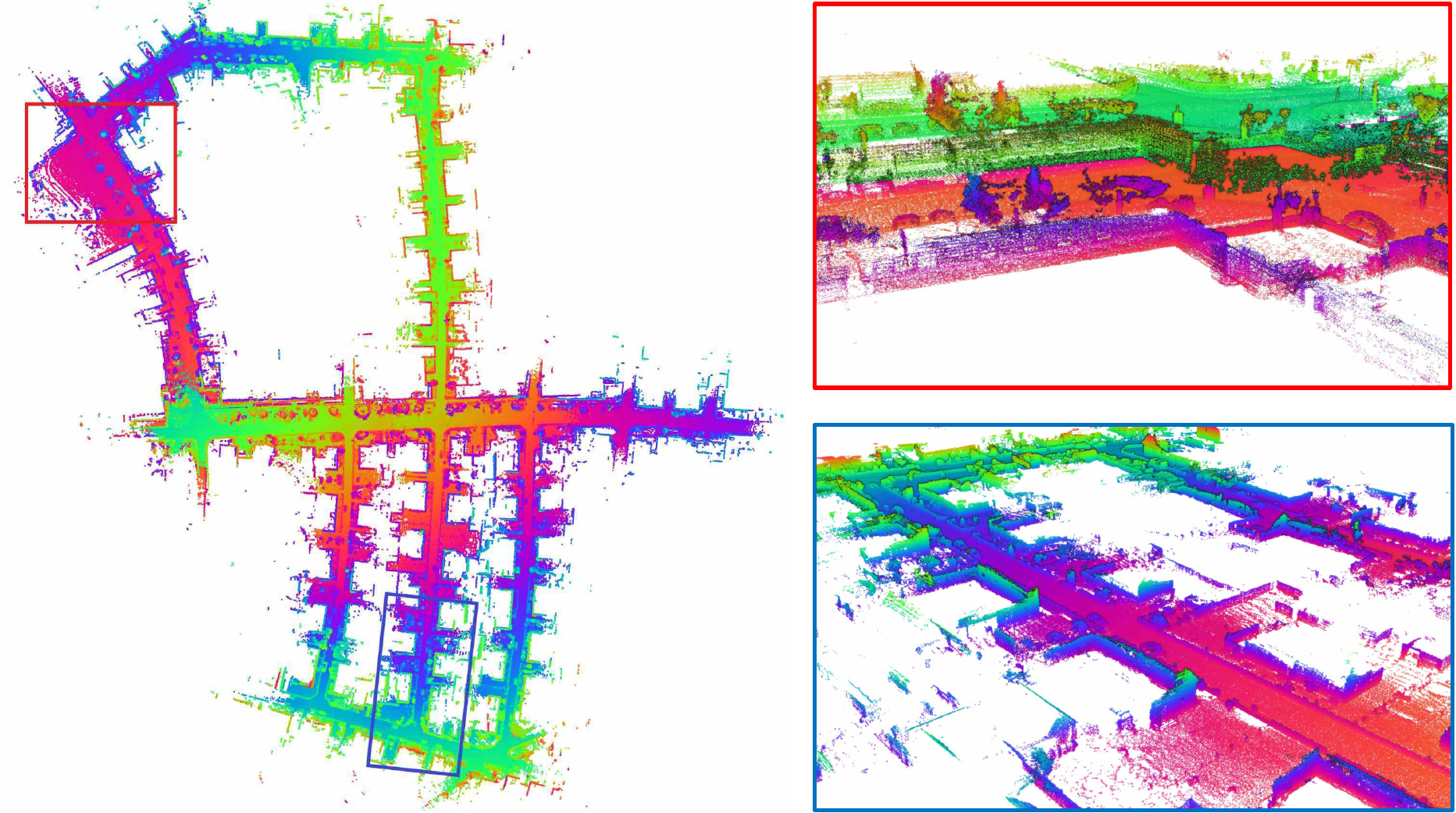}
	  \end{minipage}
  }
  \subfloat[]{
    \begin{minipage}[b]{0.36\textwidth}
	    \centering
	    \includegraphics[width=\textwidth]{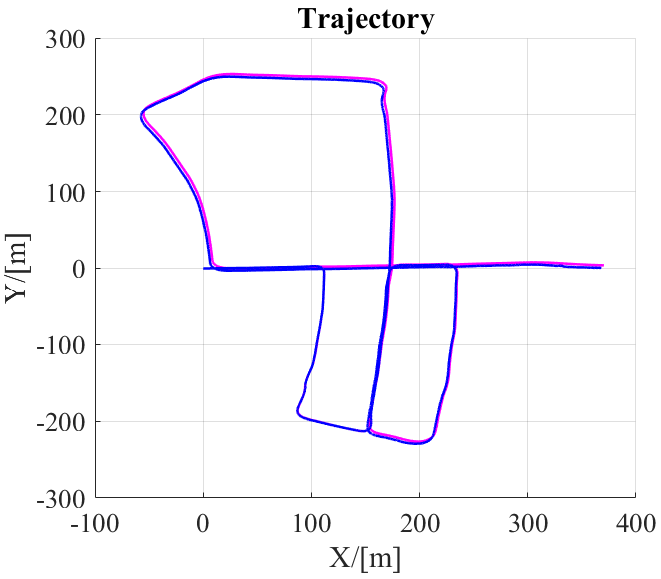}
	  \end{minipage}
  } 
  \caption{Constructed points maps with details and estimated trajectories. (a), (c), (e) maps of Seq.00, 01, and 05; (b), (d), (f) trajectories of Seq.00, 01, and 05 (\textcolor{magenta}{groundtruths} and \textcolor{blue}{InTEn-LOAM})}\label{FIG:13}
\end{figure}

For in-depth analysis, three representative sequences, i.e., Seq.00, 01, and 05, were selected. Seq.00 is a urban road dataset with the longest traveling distance, in which big and small loop closures are included, while geometric features are extremely rich. Consequently, the sequence is suitable for visualizing the trajectory drift of InTEn-LOAM. Seq.01 is a highway dataset with the fastest driving speed. Due to the lack of geometric features in the highway neighborhood, it is the most challenging sequence in the KITTI odometry dataset. Seq.05 is a country road sequence with great variation in elevation and rich structured features.

For Seq.01, it can be seen from Fig.\ref{FIG:13}(c) that areas with landmarks are circled by blue bounding boxes, while magenta boxes highlight road signs on the roadside. The drift of the estimated trajectory of Seq.01 by InTEn-LOAM is quite small (see Fig.\ref{FIG:13}(d)), which reflects that the roadside guideposts can be utilized as reflector features since their high-reflective surfaces, and are conducive to improving the LO performance in such geometric-sparse highway environments. The result also proves that the proposed InTEn-LOAM is capable of adaptively mining and fully exploiting the geometric and intensity features in surrounding environments, which ensures the LO system can accurately and robustly estimate the vehicle pose even in some challenging scenarios. In terms of Seq.00 and 05, both two point cloud maps show excellent consistency in the small loop closure areas (see blue bound regions in Fig.\ref{FIG:13}(a) and (e)), which indicates that InTEn-LOAM owns good local consistency. However, in large-scale loop closure areas, such as the endpoint, the global trajectory drifts incur a stratification issue in point cloud maps (see red bound regions in Fig.\ref{FIG:13}(a) and (e)), which are especially significant in the vertical direction. (see plane trajectory plots in Fig.\ref{FIG:13}(b) and (f)). This phenomenon is because constraints in the z-direction are insufficient in comparison with other directions in the state space since only ground features provide constraints for the z-direction during the point cloud alignment.

\subsubsection{\bf{Autonomous driving dataset}}
The other quantitative evaluation test was conducted on the autonomous driving field dataset, the groundtruth of which is referred to the trajectory output of the onboard positioning and orientation system (POS). There is a $150m$ long tunnel in the data acquisition environment, which is extremely challenging for most LO systems. The root means square errors (RMSE) of horizontal position and yaw angle were used as indicators for the absolute state accuracy. LOAM, MULLS, and HDL-Graph-SLAM were utilized as control groups, whose results are listed in Table.\ref{TAB:2}. 
\begin{figure}[hp] 
  \centering
  \subfloat[]{
    \begin{minipage}[b]{0.7\textwidth}
	    \centering
	    \includegraphics[width=\textwidth]{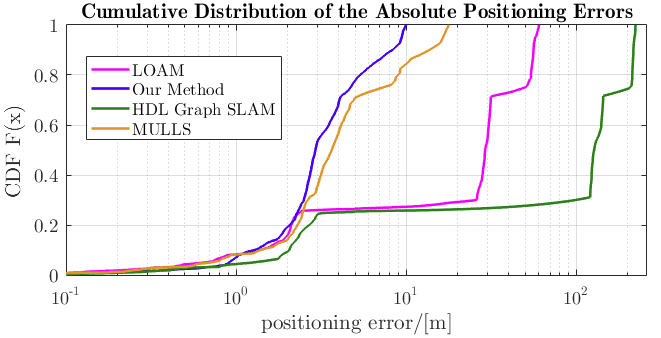}
	  \end{minipage}
  }

  \subfloat[]{
    \begin{minipage}[b]{0.7\textwidth}
	    \centering
	    \includegraphics[width=\textwidth]{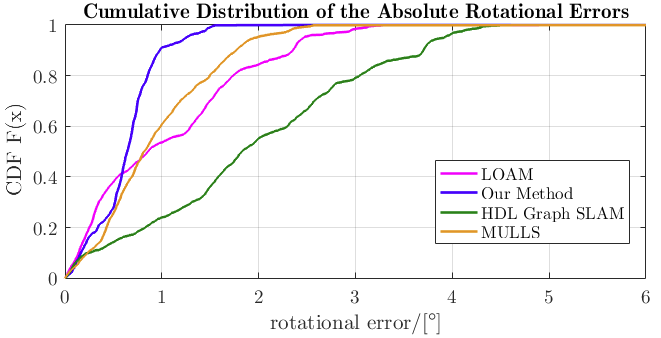}
	  \end{minipage}
  } 

  \subfloat[]{
    \begin{minipage}[b]{0.7\textwidth}
	    \centering
	    \includegraphics[width=\textwidth]{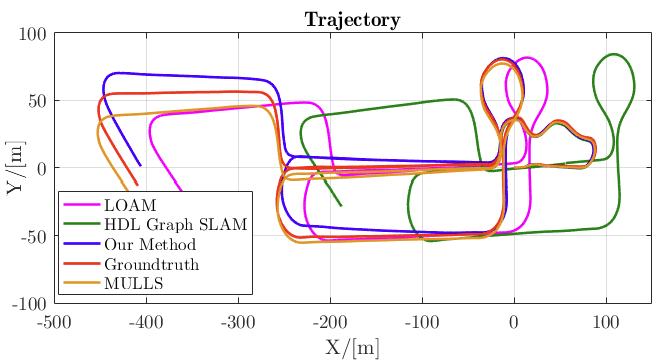}
	  \end{minipage}
  }
  \caption{Cumulative distributions of absolute state errors and estimated trajectories. (a) Cumulative distributions of the absolute positioning errors; (b) Cumulative distributions of the absolute rotational errors; (c) Estimated trajectories. (\textcolor{blue}{InTEn-LOAM}, \textcolor{magenta}{LOAM}, \textcolor{green}{HDL-Graph-SLAM}, \textcolor{Orange}{groundtruth})}\label{FIG:14}
\end{figure}

\begin{table}[h]
  \large
  \centering
  \caption{Quantitative evaluation and comparison on autonomous driving field dataset.}
  \label{TAB:2}
  \resizebox{0.48\textwidth}{!}{%
  \begin{tabular}{|c|ccc|c|}
  \hline
  \multirow{2}{*}{method} & \multicolumn{3}{c|}{Positioning error $[m]$} & Heading error $[^{\circ}]$ \\ \cline{2-5} 
                 & $x$     & $y$    & horizontal    & yaw   \\ \hline
  LOAM           & 29.478  & 18.220 & 34.654  & 1.586 \\ \hline
  HDL-Graph-SLAM & 119.756 & 75.368 & 141.498 & 2.408 \\ \hline
  MULLS-LO       & 4.133   & 5.705  & 7.043   & 1.403 \\ \hline
  \textbf{InTEn-LOAM}     & \textbf{1.851}   & \textbf{1.917}  & \textbf{2.664}   & \textbf{0.476} \\ \hline
  \end{tabular}%
  }
\end{table}

Both LOAM and HDL-Graph-SLAM failed to localize the vehicle with $34.654m$ and $141.498m$ positional errors, respectively. MULLS and the proposed InTEn-LOAM are still able to function properly with $2.664m$ and $7.043m$ of positioning error and $0.476^{\circ}$ and $1.403^{\circ}$ of heading error within the path range of $1.5km$. To further investigate the causes of this result, we plotted the cumulative distribution of absolute errors and horizontal trajectories of three LO systems, as shown in Fig.\ref{FIG:14}. 

\begin{figure}[hp] 
  \centering
  \subfloat[]{
    \begin{minipage}[b]{0.7\textwidth}
	    \centering
	    \includegraphics[width=\textwidth]{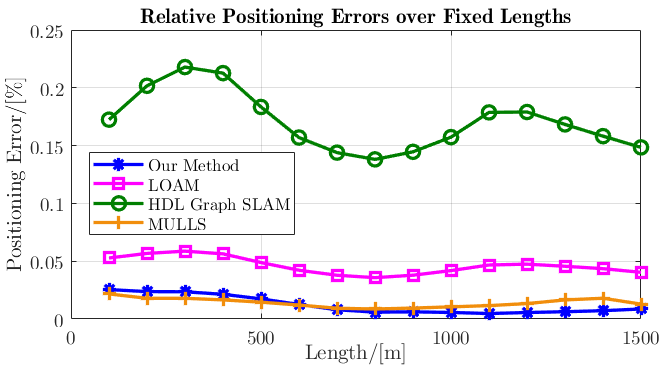}
	  \end{minipage}
  }

  \subfloat[]{
    \begin{minipage}[b]{0.7\textwidth}
	    \centering
	    \includegraphics[width=\textwidth]{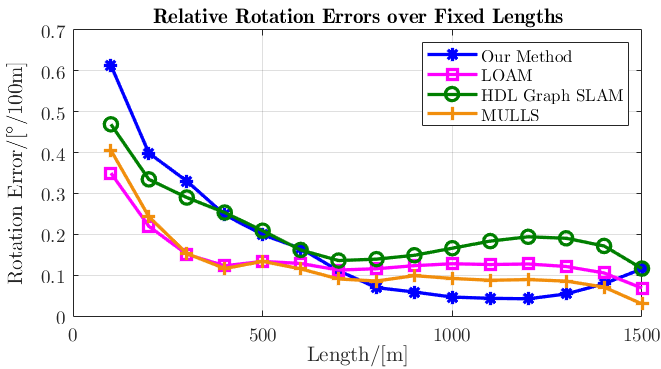}
	  \end{minipage}
  } 
  \caption{The average RTE and RRE of LO systems over fixed lengths. (a) RTE; (b) RRE.}\label{FIG:21}
\end{figure}

From the trajectory plot, we can see that the overall trajectory drift of InTEn-LOAM and MULLS are relatively small, indicating that these two approaches can accurately localize the vehicle in this challenging scene by incorporating intensity features into the point cloud registration, and using intensity information for the feature weighting. The estimated position of LO inevitably suffers from error accumulation which is the culprit causing trajectory drift. It can be seen from the cumulative distribution of absolute errors that the absolute positioning error of InTEn-LOAM is no more than $10m$, and the attitude error is no more than $1.5^{\circ}$. The overall trends of rotational errors of the other three systems are consistent with that of InTEn-LOAM. Results in Table.\ref{TAB:2} also verify that their rotation errors are similar. The cumulative distribution curves of absolute positioning errors of LOAM and HDL-Graph-SLAM do not exhibit smooth growth but a steep increase in some intervals. The phenomenon reflects the existence of anomalous registration in these regions, which is consistent with the fact that the scan registration-based motion estimation in the tunnel is degraded. MULLS, which incorporates intensity meatures by feature constraints weighting, present a smooth curve as similar as InTEn-LOAM. However, the absolute errors of positioning (no more than $19m$) and heading (no more than $2.5^{\circ}$) are both large than those of our proposed LO system. We also plotted the RTE and RRE of all four approaches (see Fig.\ref{FIG:21}). It can bee seen that the differences of the RRE between four systems are small, representing that the heading estimations of all theses LO systems are not deteriorated in the geometric-degraded long straight tunnel. In contrast, the RPEs are quite different. Both LOAM and HDL-Graph-SLAM suffer from serious scan registration drifts, while MULLS and InTEn-LOAM are able to positioning normally, and achieve very close relative accuracy. 

\subsection{Point cloud map quality} 
\subsubsection{\bf{Large-scale urban scenario}}
\begin{figure}[hp] 
  \centering
  \subfloat[]{
    \begin{minipage}[b]{0.21\textwidth}
	    \centering
	    \includegraphics[width=\textwidth]{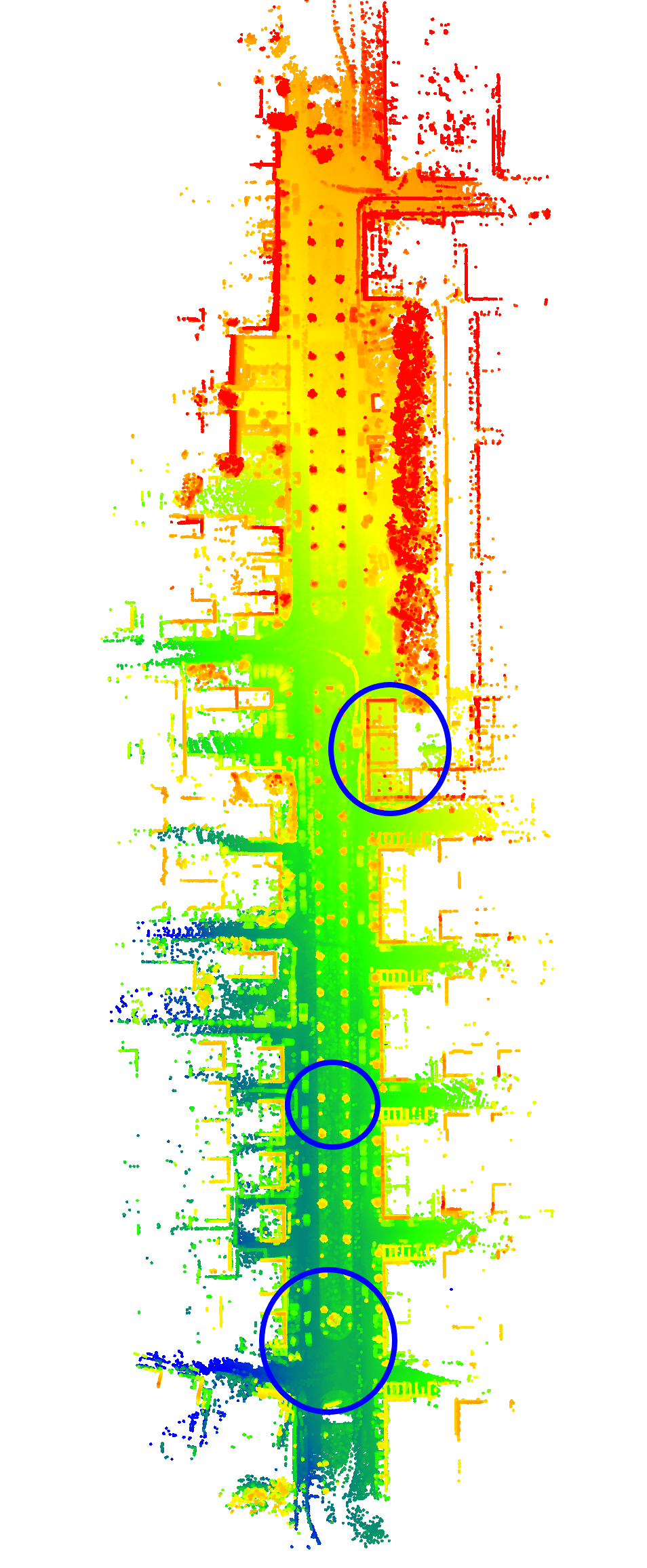}
	  \end{minipage}
  }
  \subfloat[]{
    \begin{minipage}[b]{0.34\textwidth}
	    \centering
	    \includegraphics[width=\textwidth]{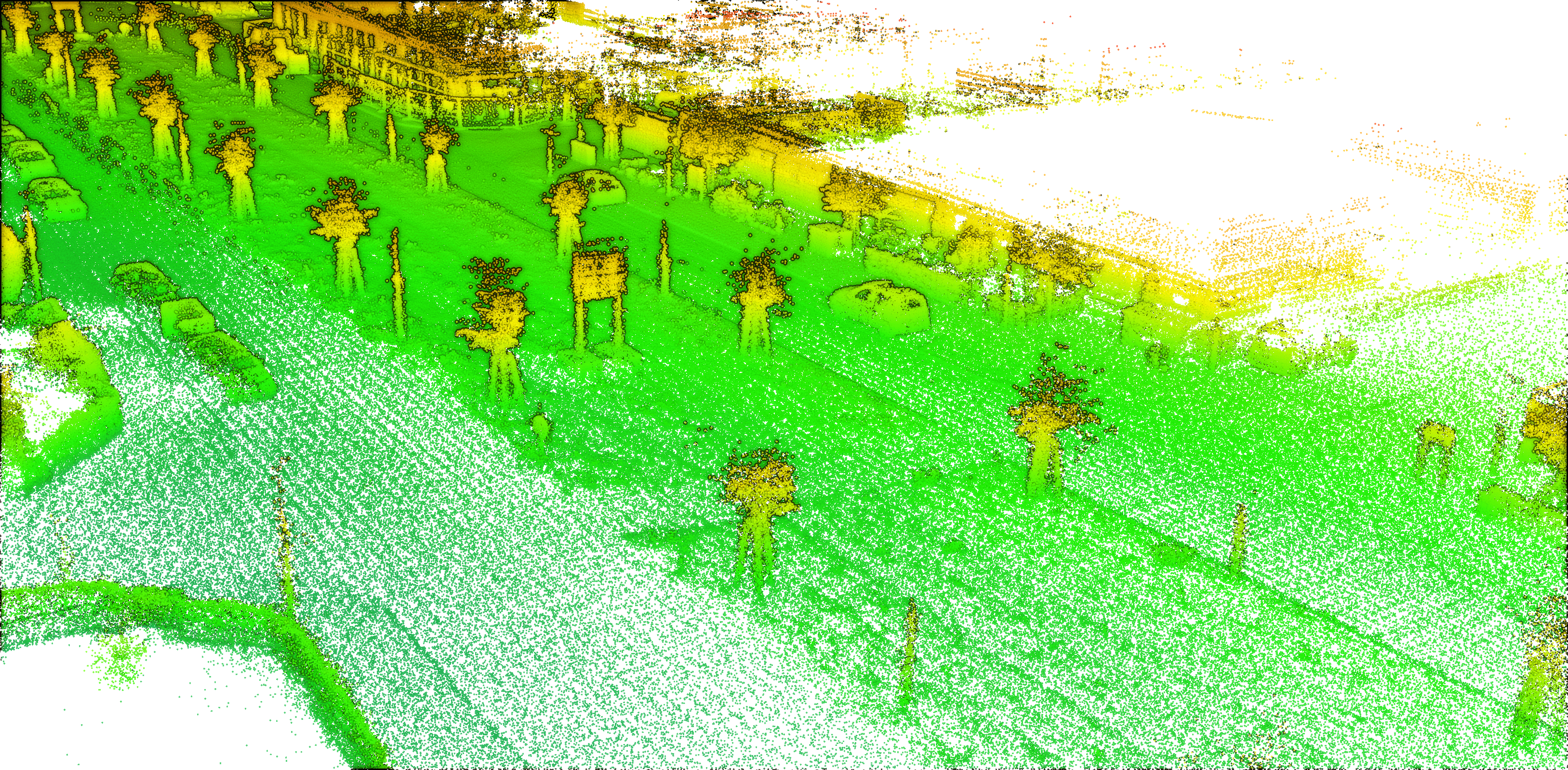}

      \includegraphics[width=\textwidth]{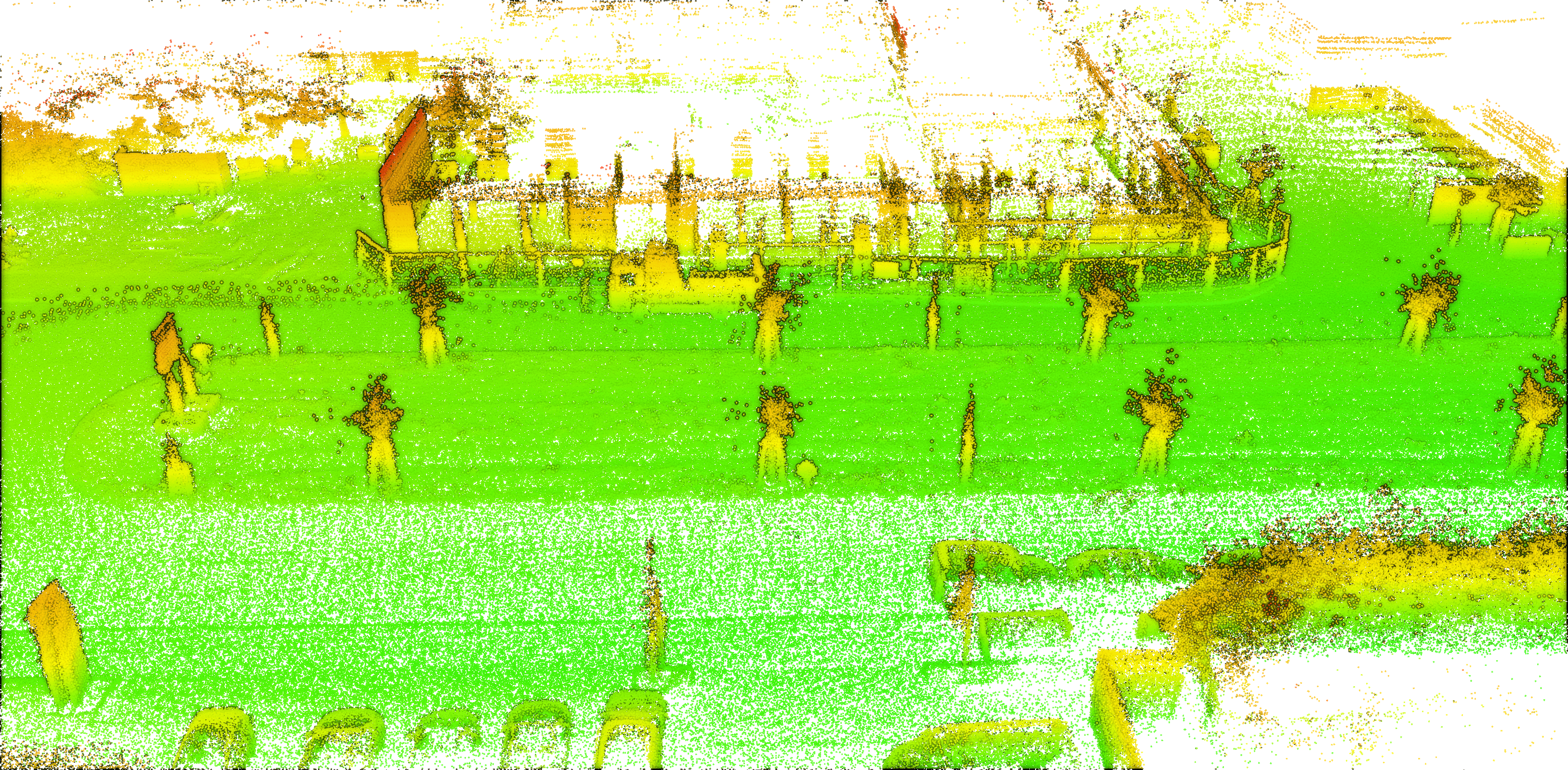}

      \includegraphics[width=\textwidth]{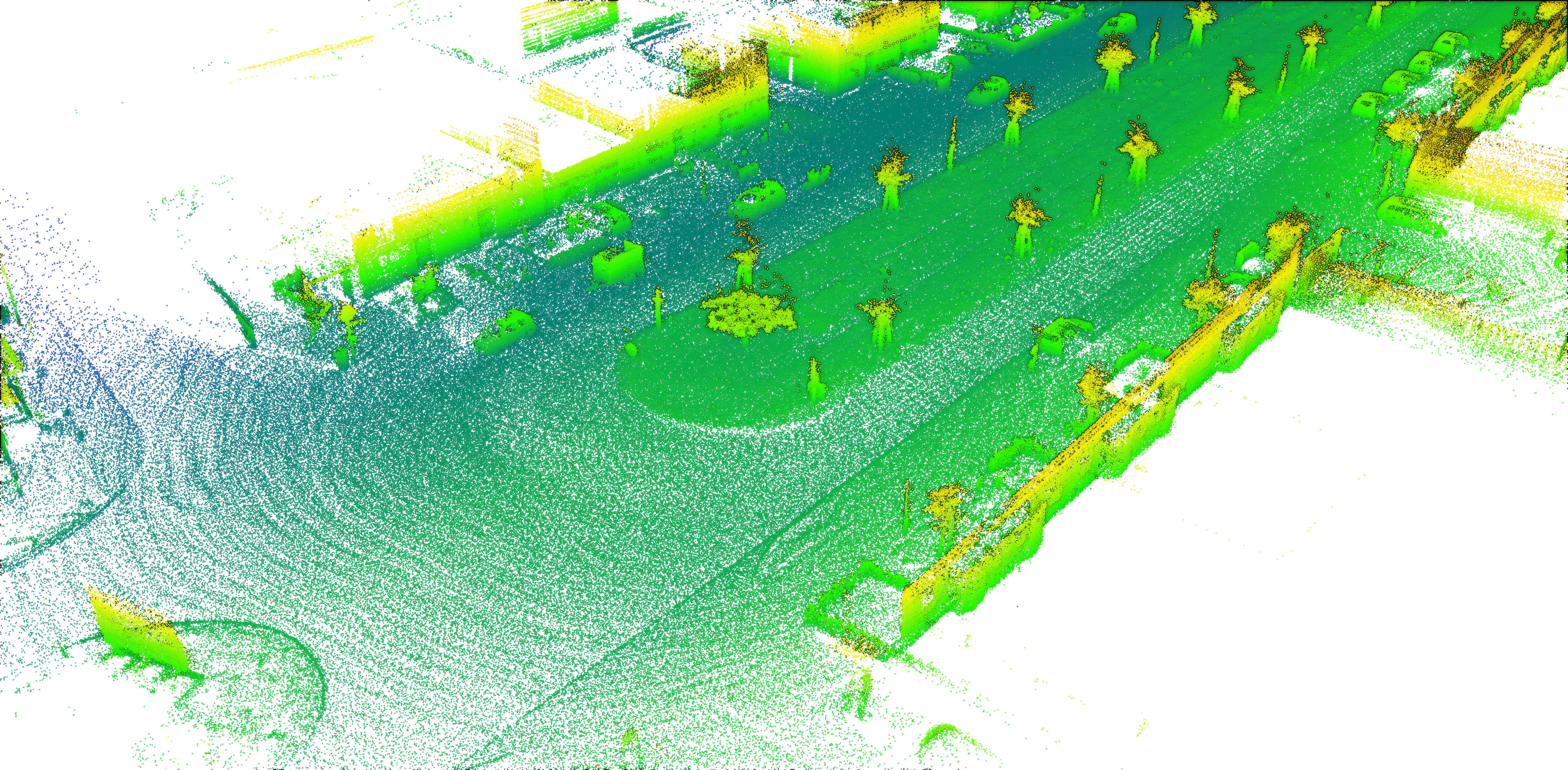}
	  \end{minipage}
  } 
  \subfloat[]{
    \begin{minipage}[b]{0.34\textwidth}
	    \centering
	    \includegraphics[width=\textwidth]{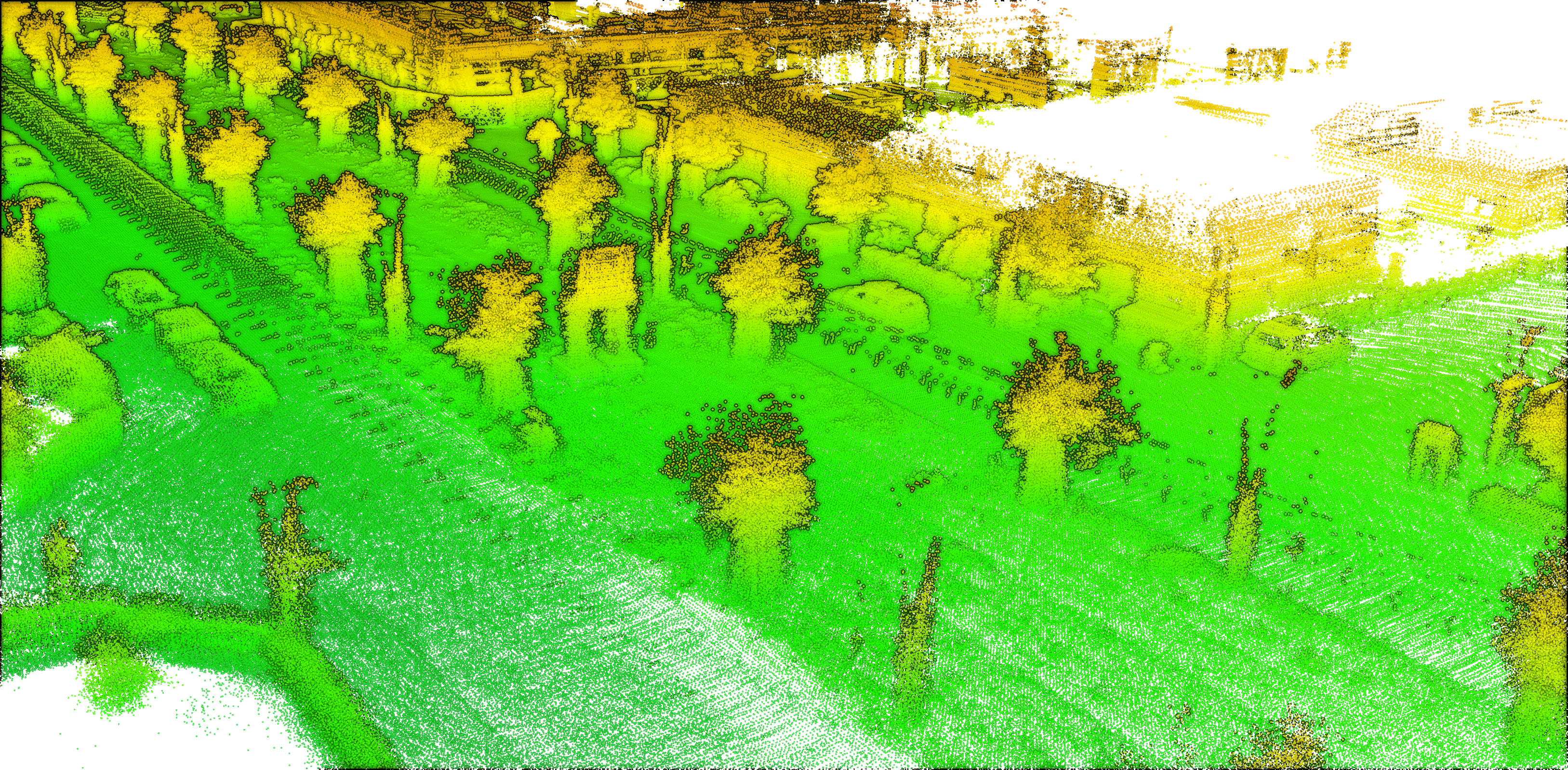}

      \includegraphics[width=\textwidth]{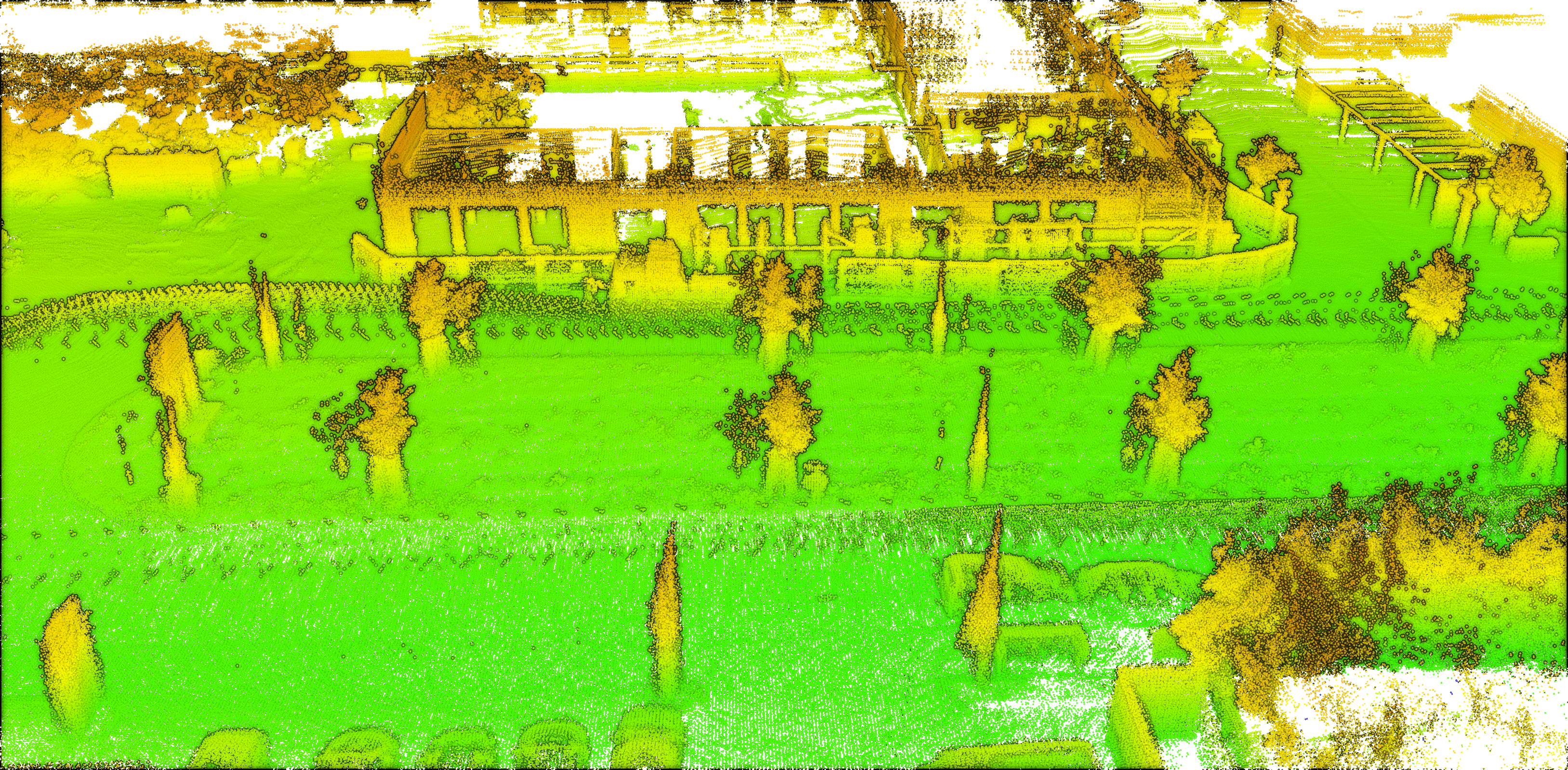}

      \includegraphics[width=\textwidth]{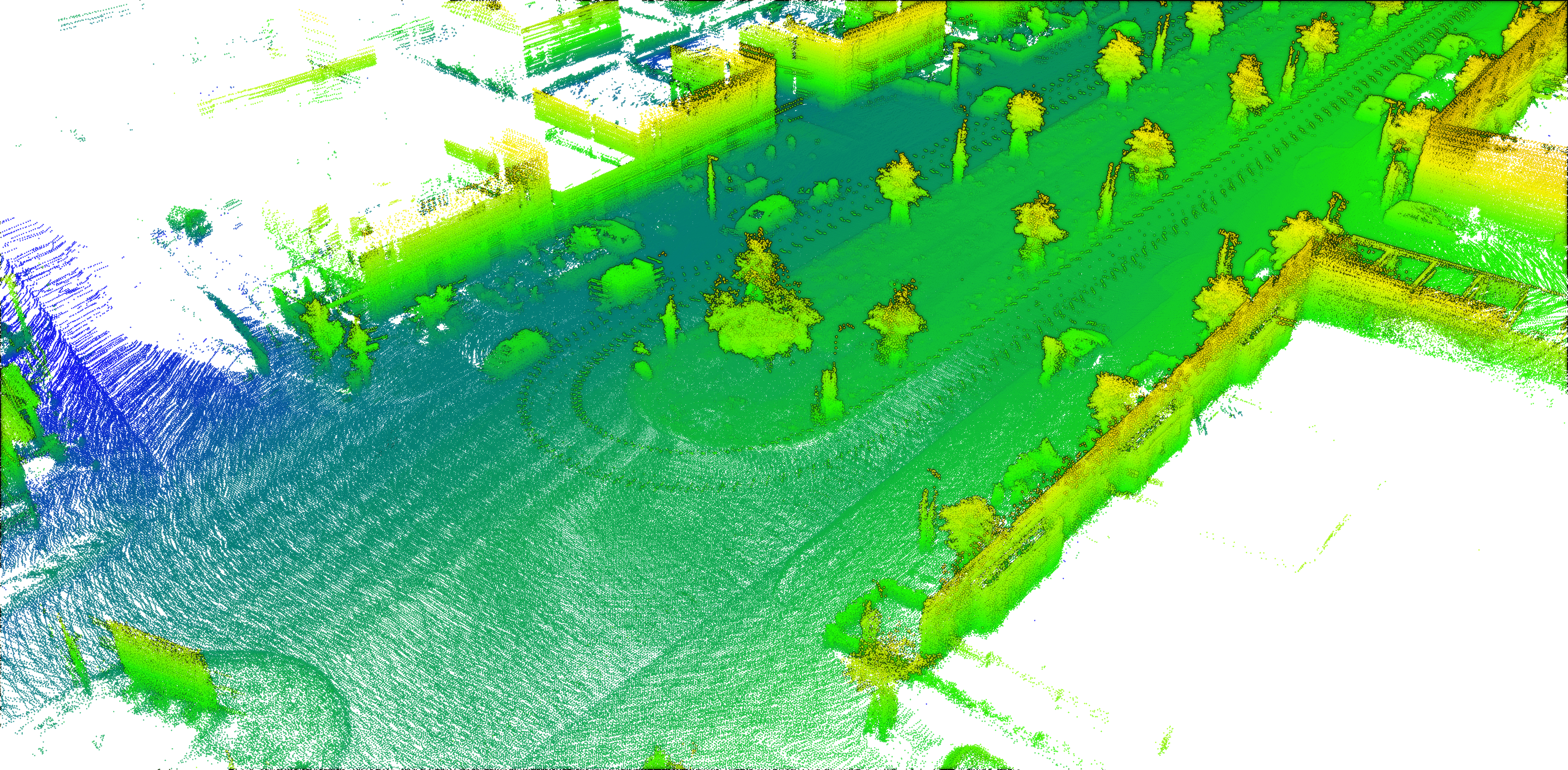}
	  \end{minipage}
  } 
  \caption{InTEn-LOAM's map result on urban scenario (KITTI seq.06): (a) overview, (b) 
  map in detail of circled areas, (c) reference map comparison.}\label{FIG:15}
\end{figure}

\begin{figure}[hp] 
  \centering
  \subfloat[]{
    \begin{minipage}[b]{0.21\textwidth}
	    \centering
	    \includegraphics[width=\textwidth]{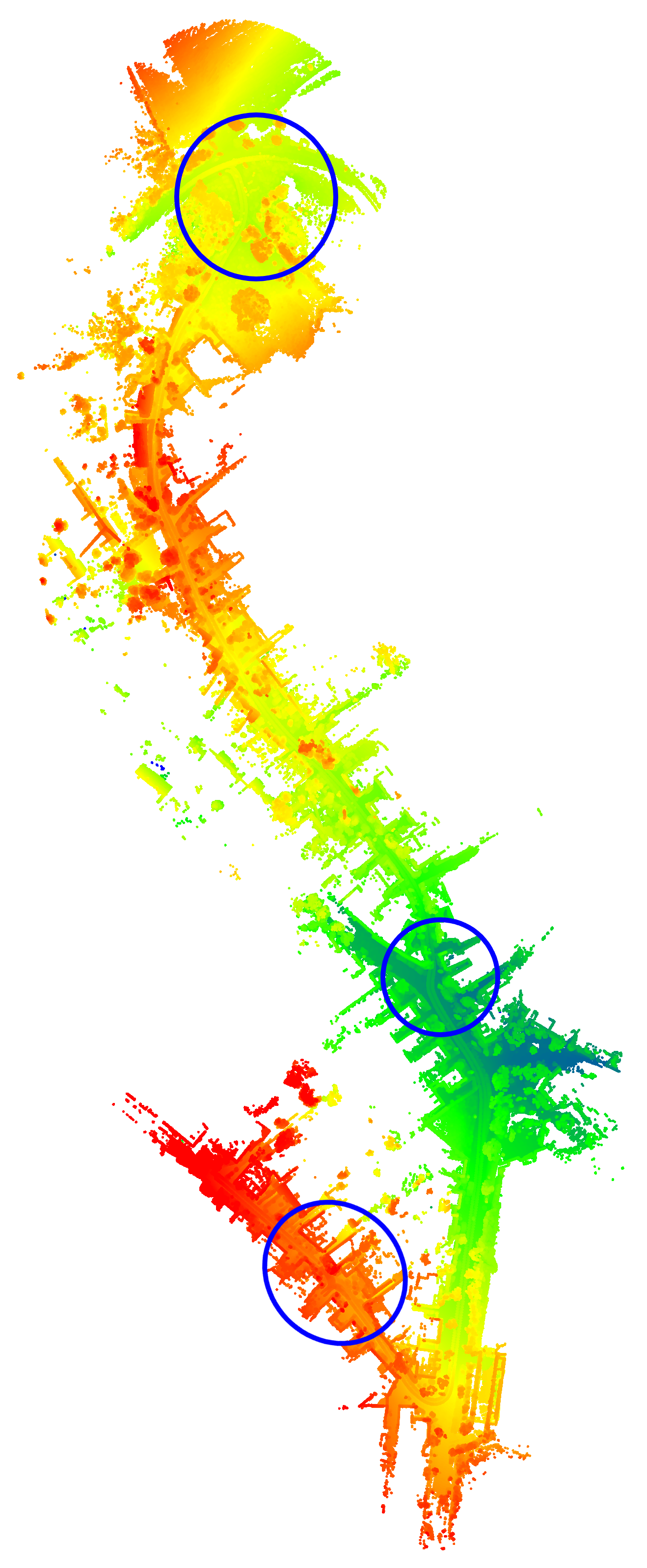}
	  \end{minipage}
  }
  \subfloat[]{
    \begin{minipage}[b]{0.34\textwidth}
	    \centering
	    \includegraphics[width=\textwidth]{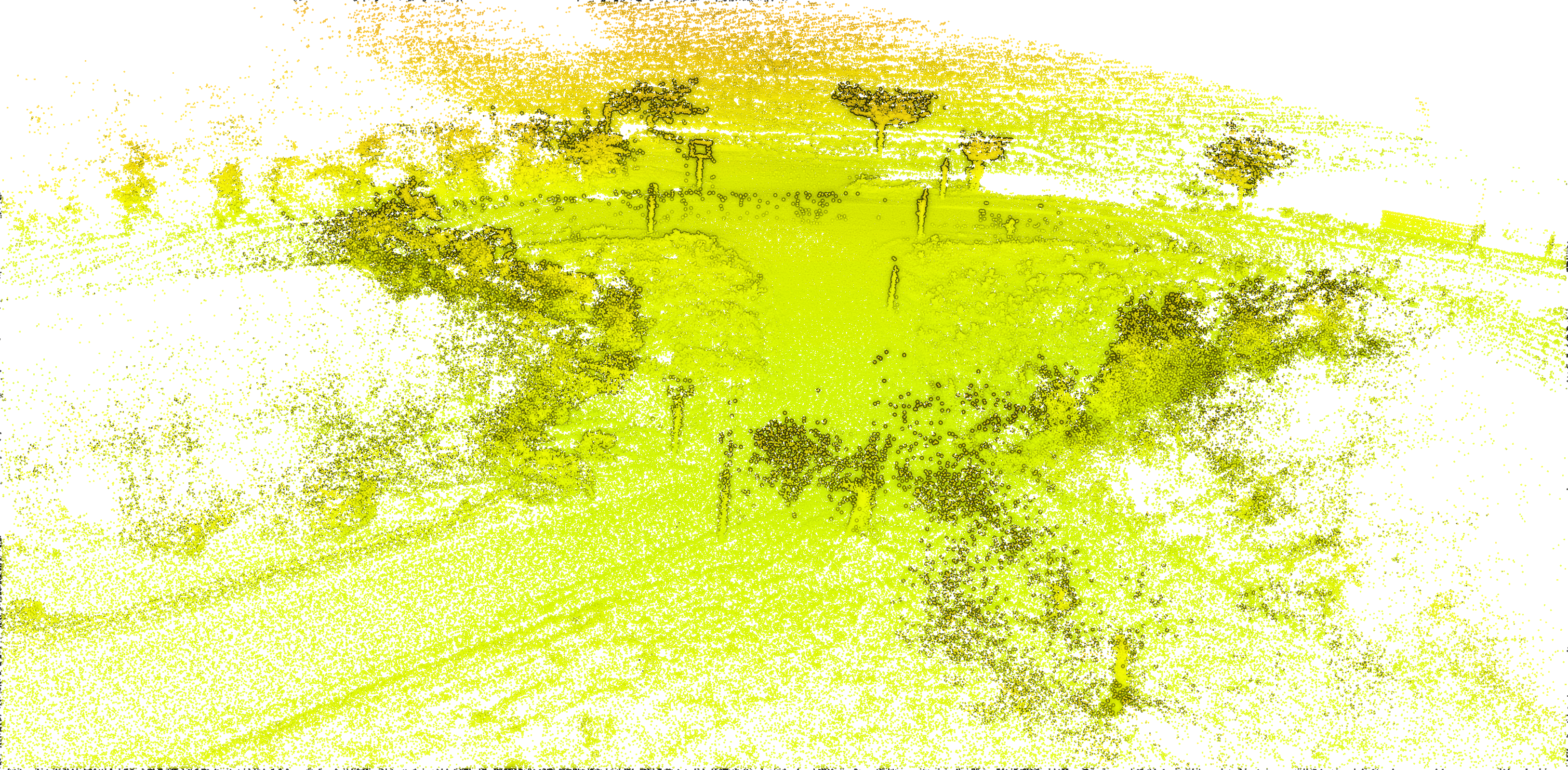}

      \includegraphics[width=\textwidth]{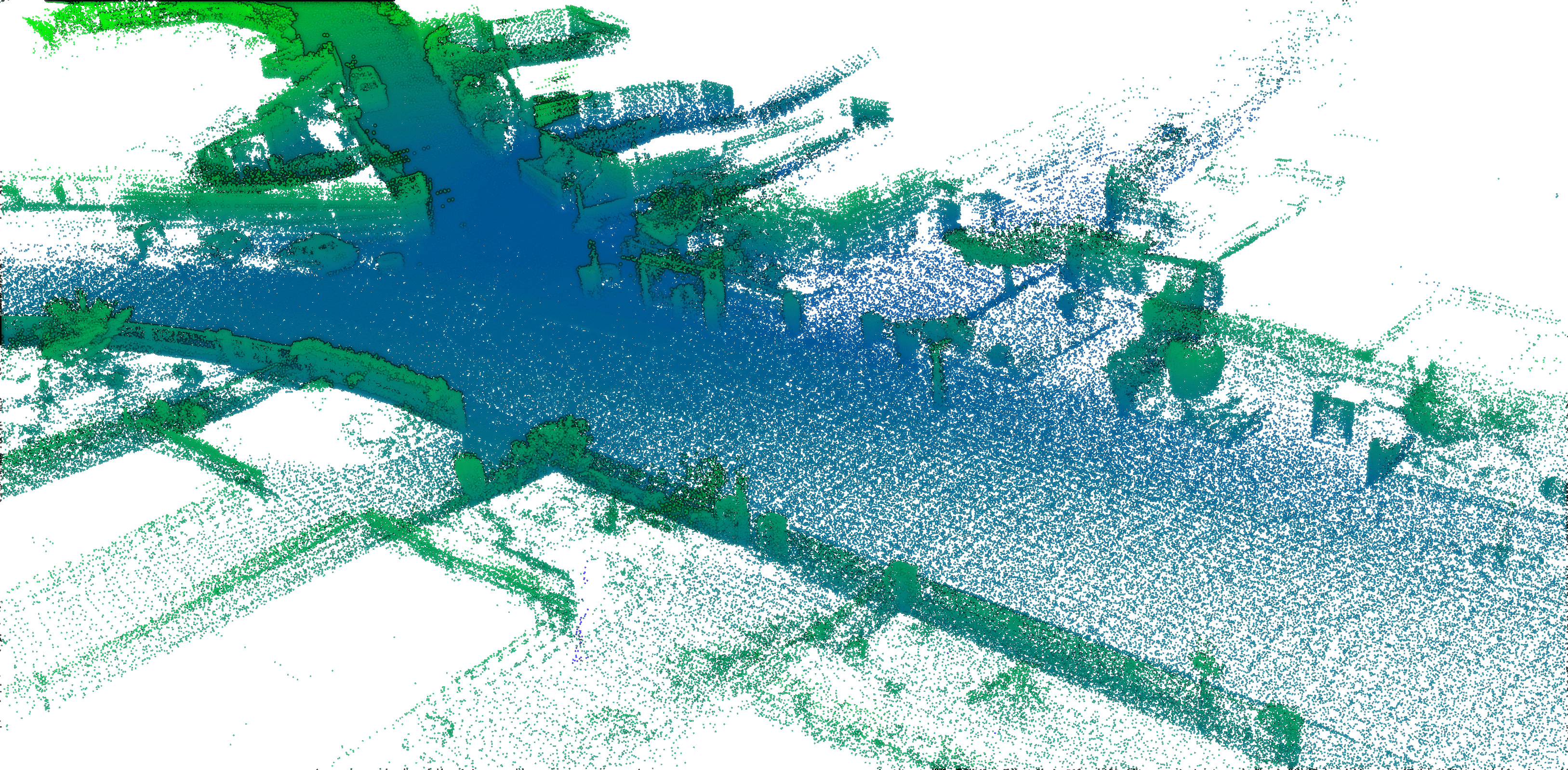}

      \includegraphics[width=\textwidth]{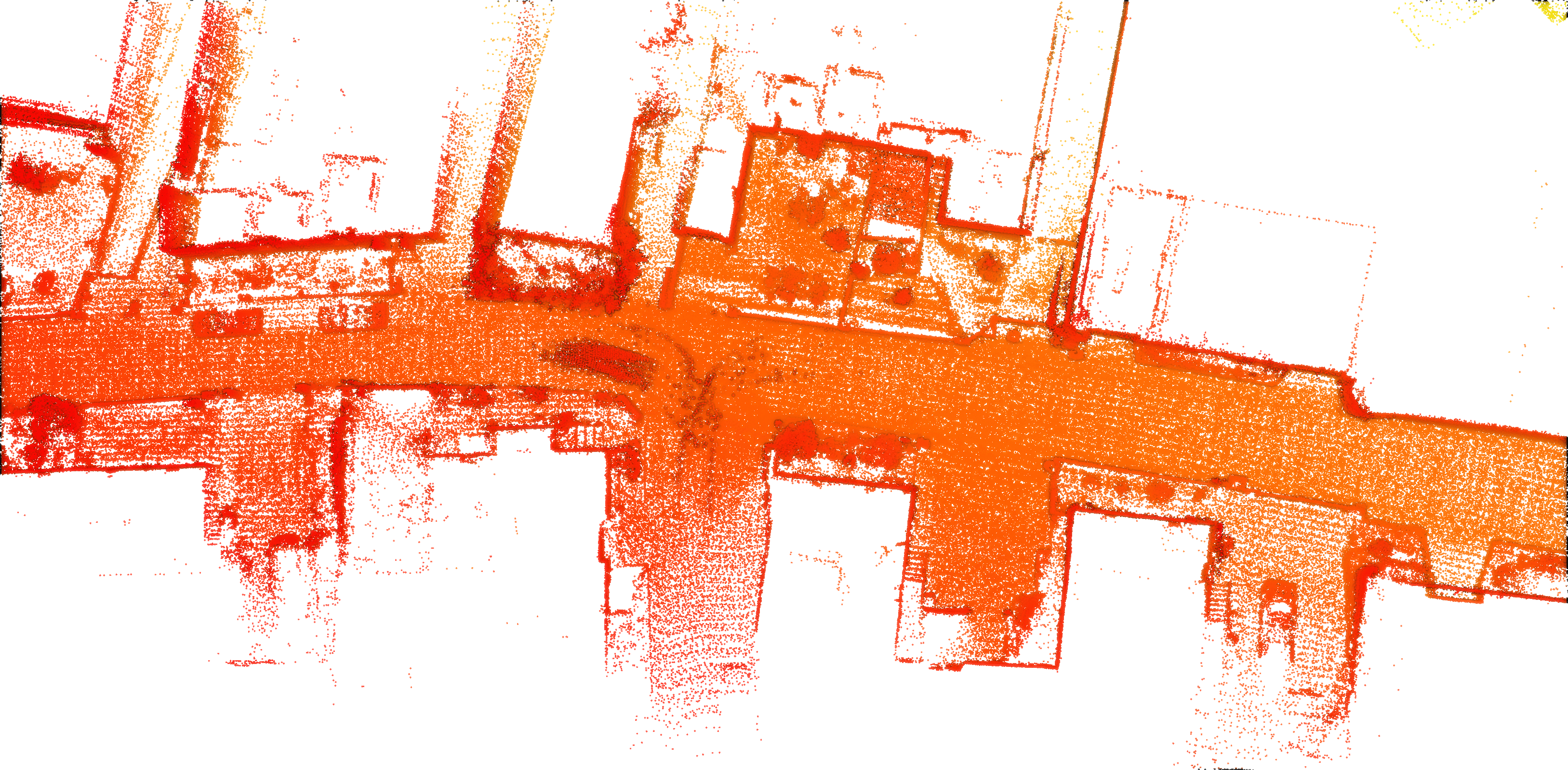}
	  \end{minipage}
  } 
  \subfloat[]{
    \begin{minipage}[b]{0.34\textwidth}
	    \centering
	    \includegraphics[width=\textwidth]{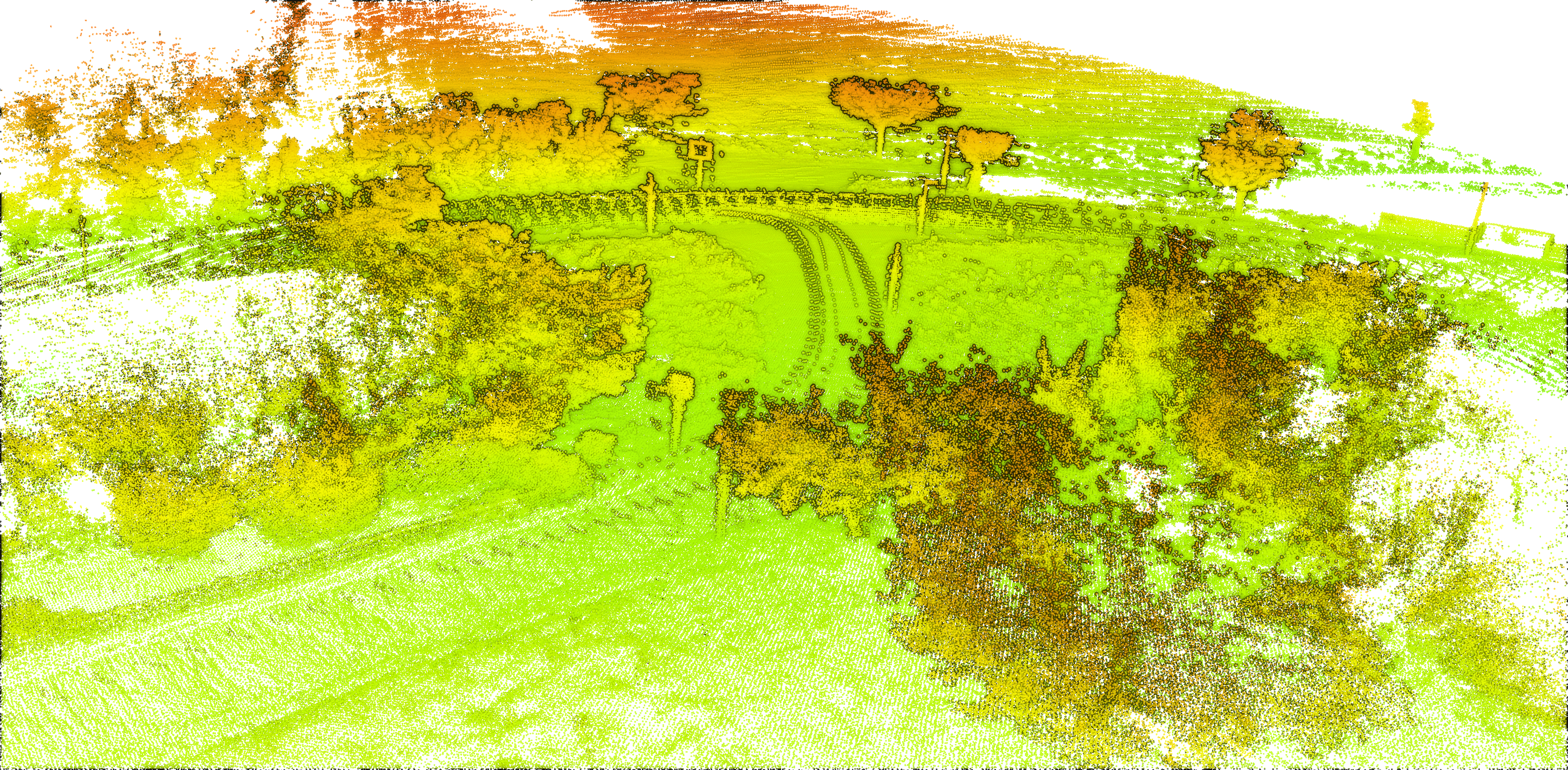}

      \includegraphics[width=\textwidth]{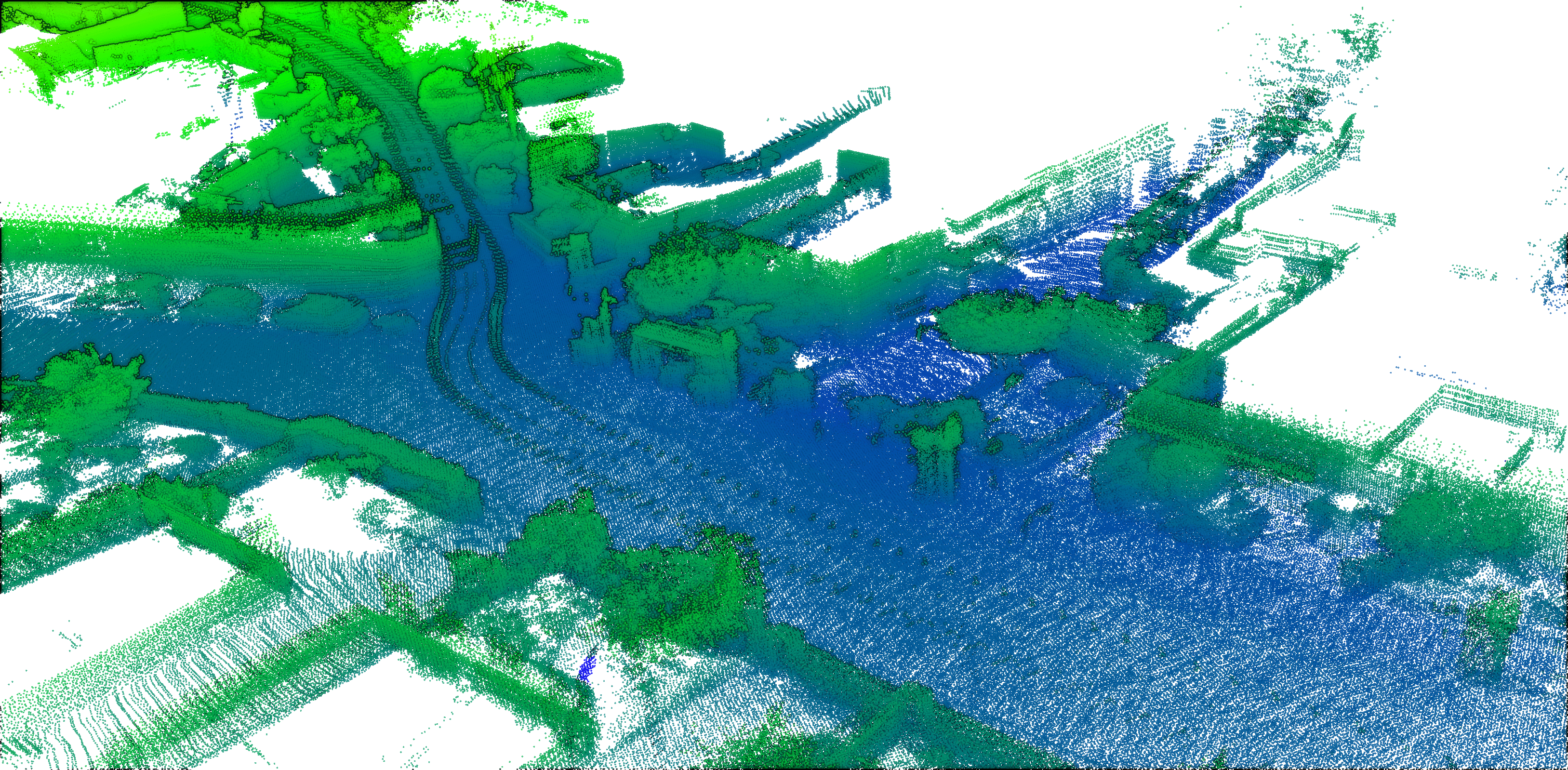}

      \includegraphics[width=\textwidth]{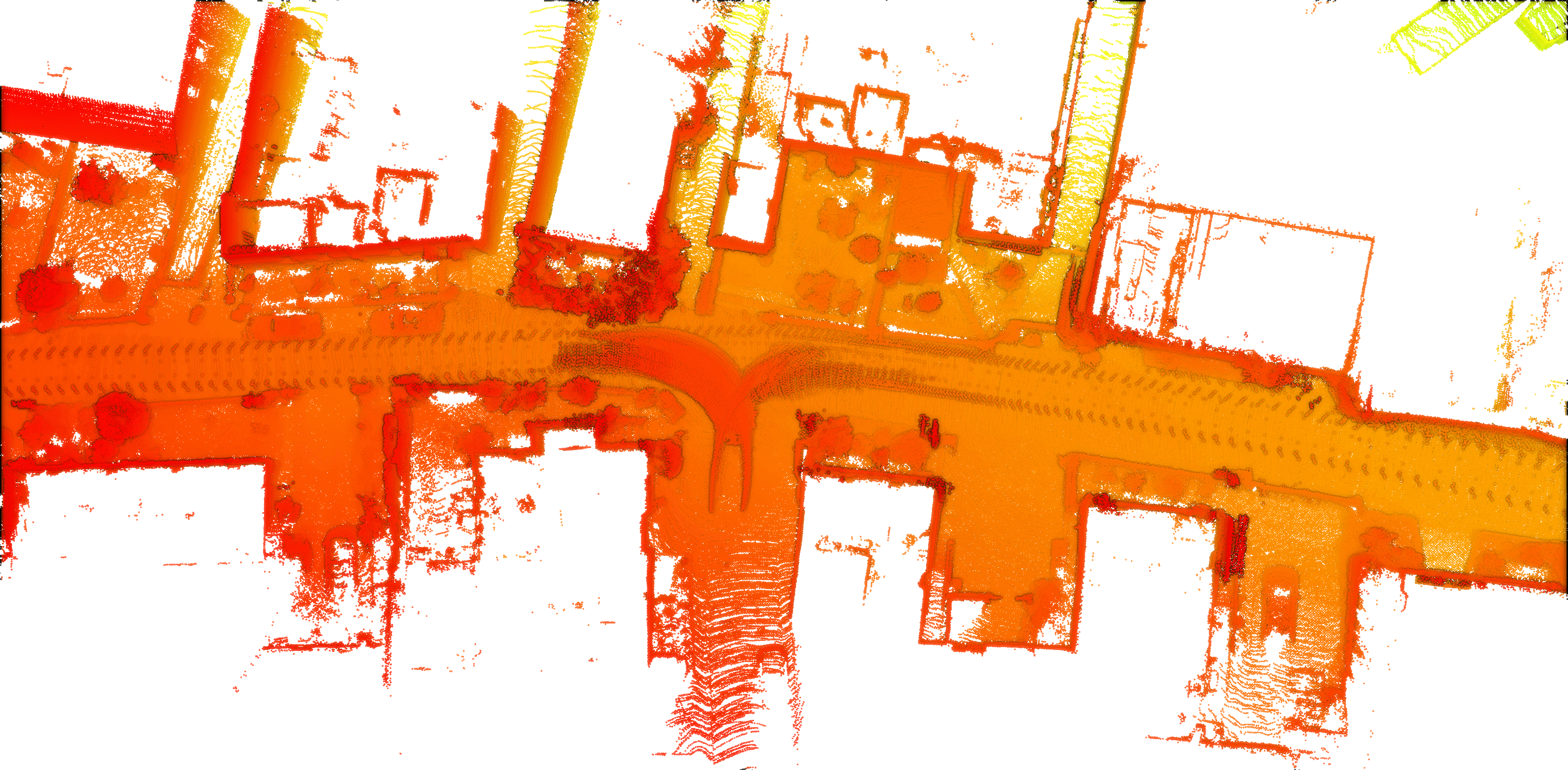}
	  \end{minipage}
  } 
  \caption{InTEn-LOAM's map result on country scenario (KITTI seq.10): (a) overview, 
  (b) map in detail of circled areas, (c) reference map comparison.}\label{FIG:16}
\end{figure}

The qualitative evaluations were conducted by intuitively comparing the constructed map by InTEn-LOAM with the reference map. The reference map is built by merging each frame of laser scan using their groundtruth poses. Maps of Seq.06 and 10 are displayed in Fig.\ref{FIG:15} and Fig.\ref{FIG:16}, which are the urban scenario with trajectory loops and the country road scenario without loop, respectively. 

Although the groundtruth is the post-processing result of POS and its absolute accuracy reaches centimeter-level, the directly merged points map is blurred in the local view. By contrast, maps built by InTEn-LOAM own better local consistency, and various small targets, such as trees, vehicles, and fences, etc., can be clearly distinguished from the points map. The above results prove that the relative accuracy of InTEn-LOAM outperforms that of the GPS/INS post-processing solution, which is very critical for the mapping tasks.

\subsubsection{\bf{Long straight tunnel scenario}}
\begin{figure}[hp] 
  \centering
  \subfloat[]{
    \begin{minipage}[b]{0.49\textwidth}
	    \centering
	    \includegraphics[width=\textwidth]{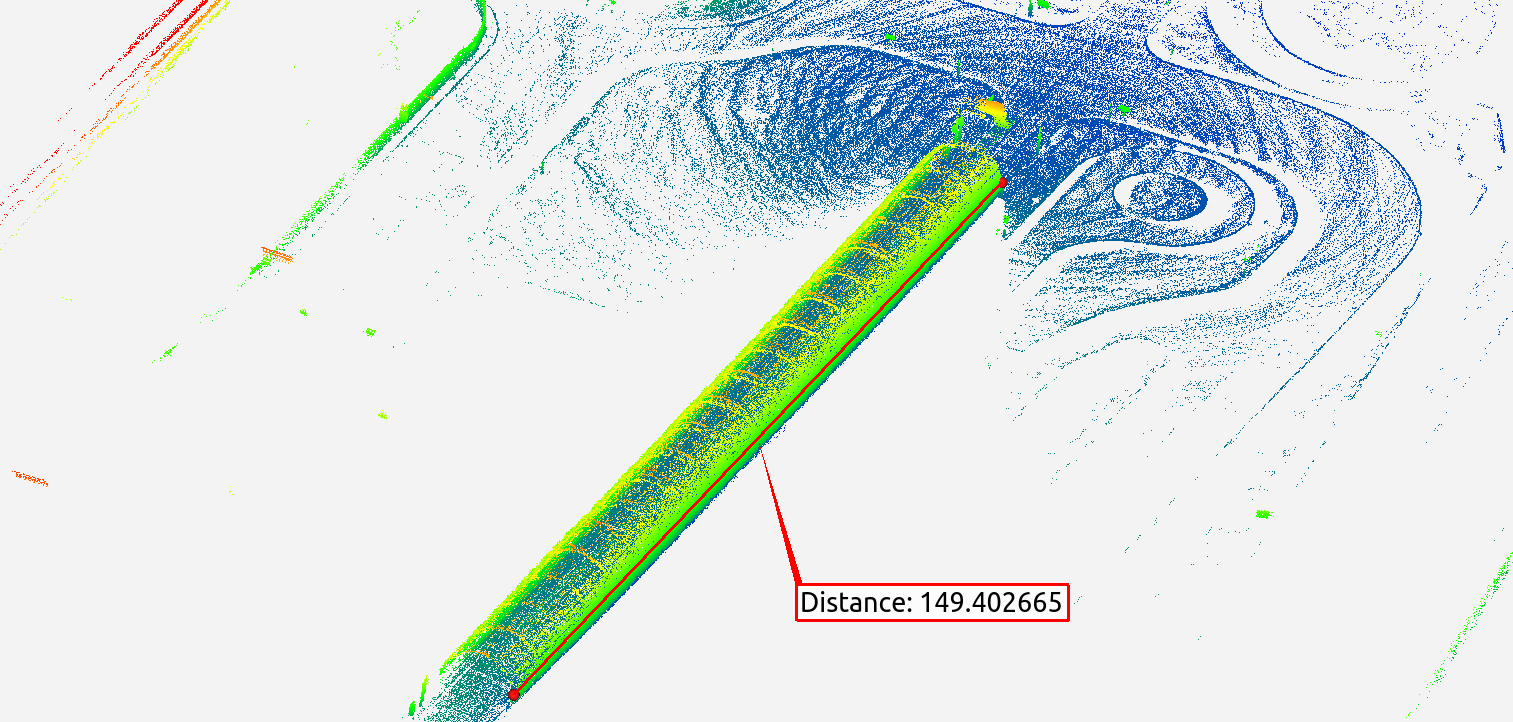}
	  \end{minipage}
  }
  \subfloat[]{
    \begin{minipage}[b]{0.49\textwidth}
	    \centering
	    \includegraphics[width=\textwidth]{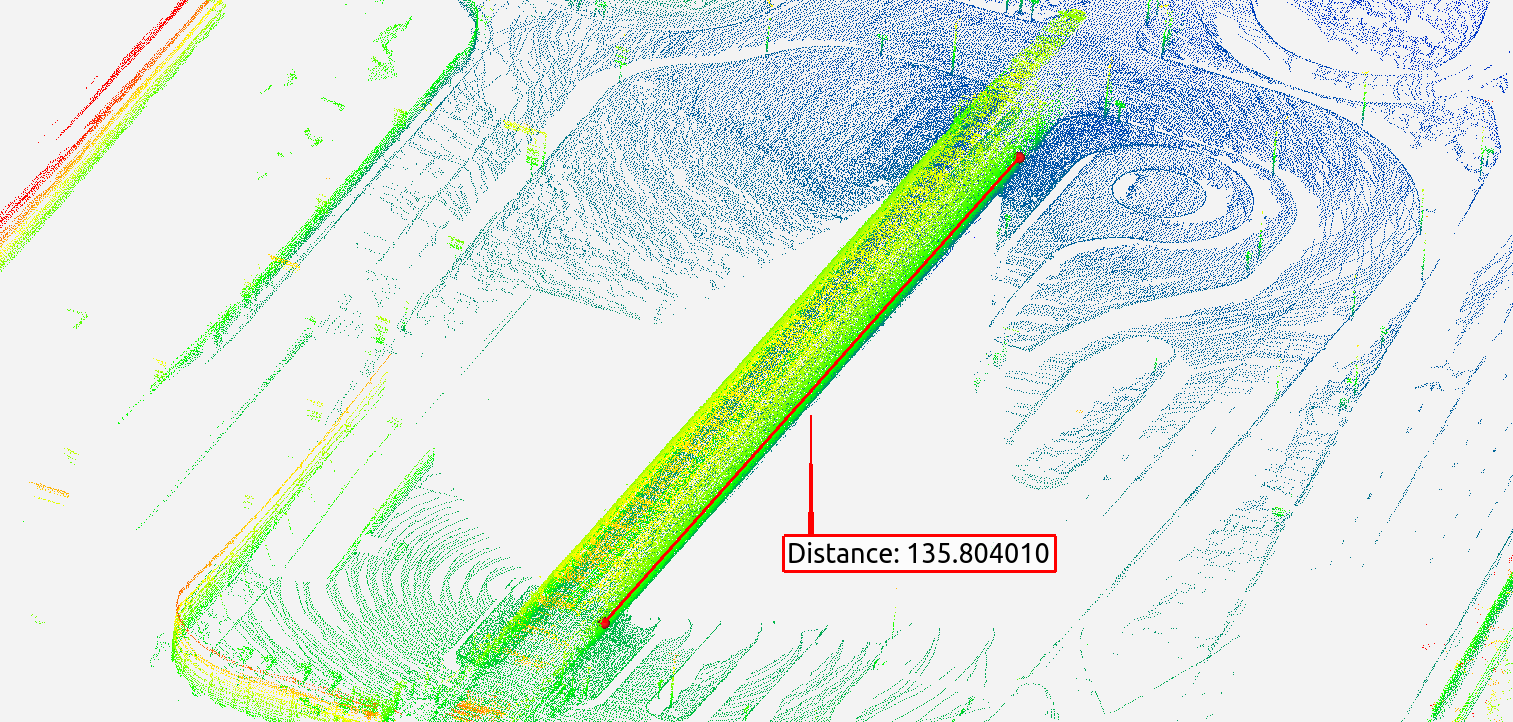}
	  \end{minipage}
  }

  \subfloat[]{
    \begin{minipage}[b]{0.49\textwidth}
	    \centering
	    \includegraphics[width=\textwidth]{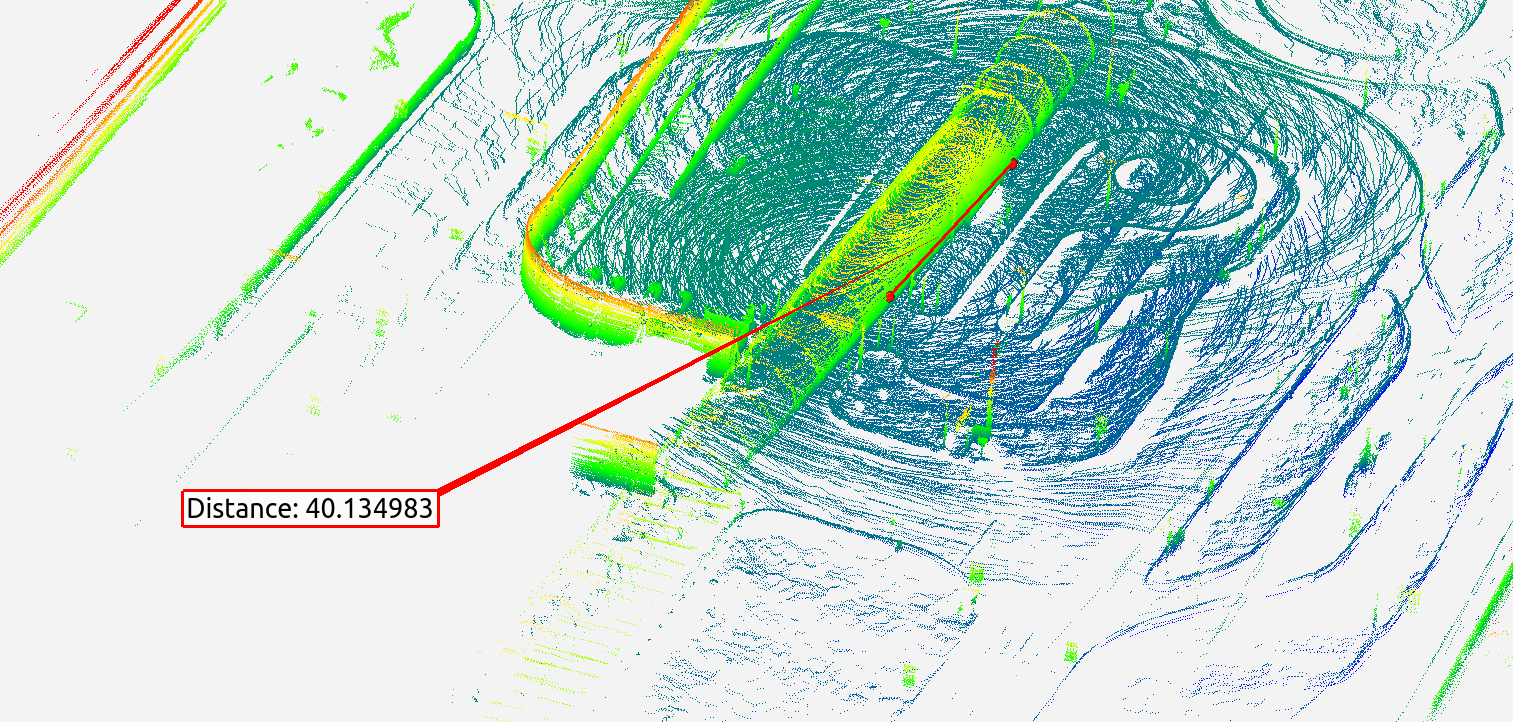}
	  \end{minipage}
  } 
  \subfloat[]{
    \begin{minipage}[b]{0.49\textwidth}
	    \centering
	    \includegraphics[width=\textwidth]{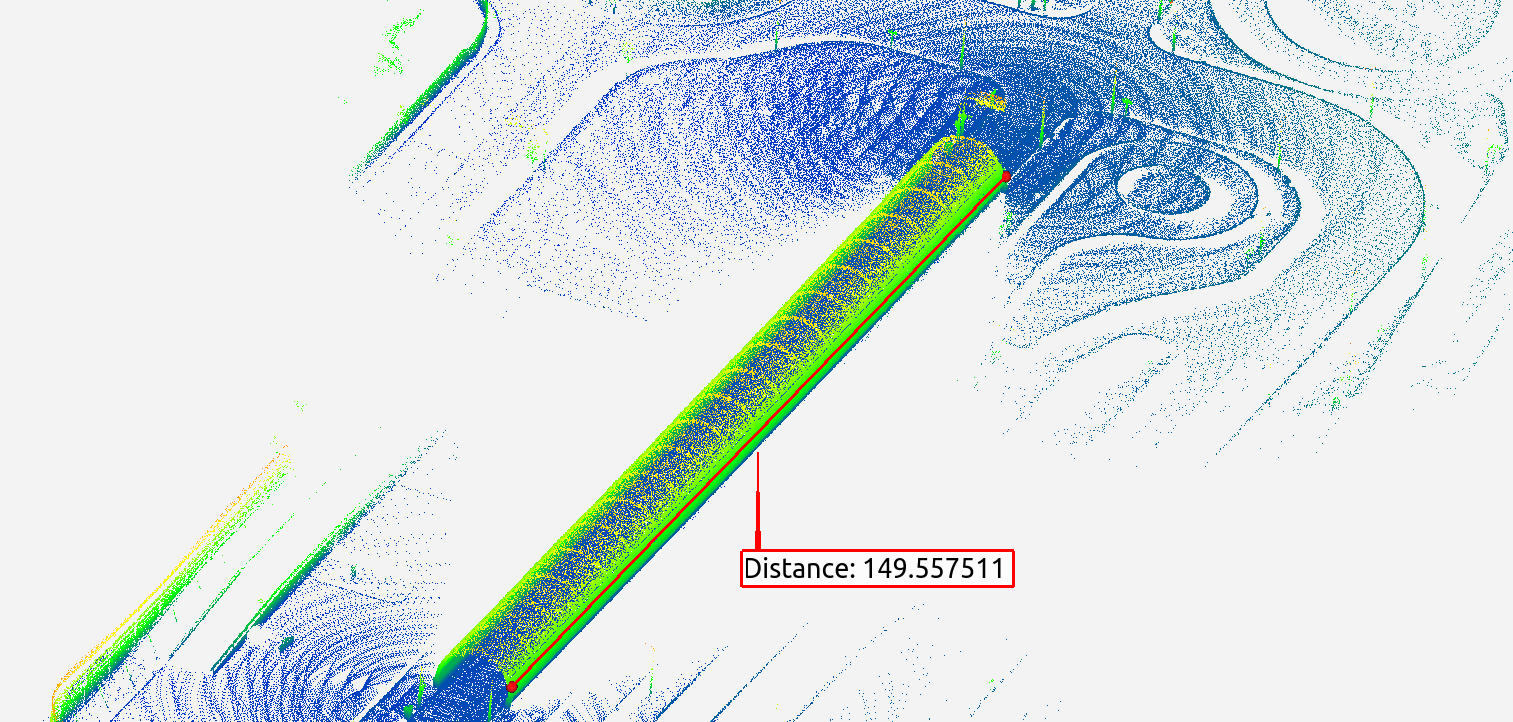}
	  \end{minipage}
  } 
  \caption{LO systems' map results on autonomous driving field dataset in the tunnel 
  region. (a) InTEn-LOAM, (b) LOAM, (c) HDL-Graph-SLAM, (d) MULLS.}\label{FIG:17}
\end{figure}

The second qualitative evaluation test was conducted on the autonomous driving field dataset. There is a $150m$ long straight tunnel in which we alternately posted some reflective signs on sidewalls to manually add some intensity features in such registration-degraded scenario. Maps of InTEn-LOAM, MULLS, LOAM, and HDL-Graph-SLAM are shown in Fig.\ref{FIG:17}. It intuitively shows that both LOAM and HDL-Graph-SLAM present different degrees of scan registration degradation, while the proposed InTEn-LOAM achieves correct motion estimation by jointly utilizing both sparse geometric and intensity features, as shown in Fig.\ref{FIG:17}. Although MULLS is also able to build a correct map since it utilizes intensity information to re-weight geometric feature constraints during the registration iteration, its accuracies of both pose estimation and mapping are inferior to the proposed LO system.

\begin{figure}[hp] 
  \centering
  \subfloat[]{
    \begin{minipage}[b]{0.8\textwidth}
	    \centering
	    \includegraphics[width=\textwidth]{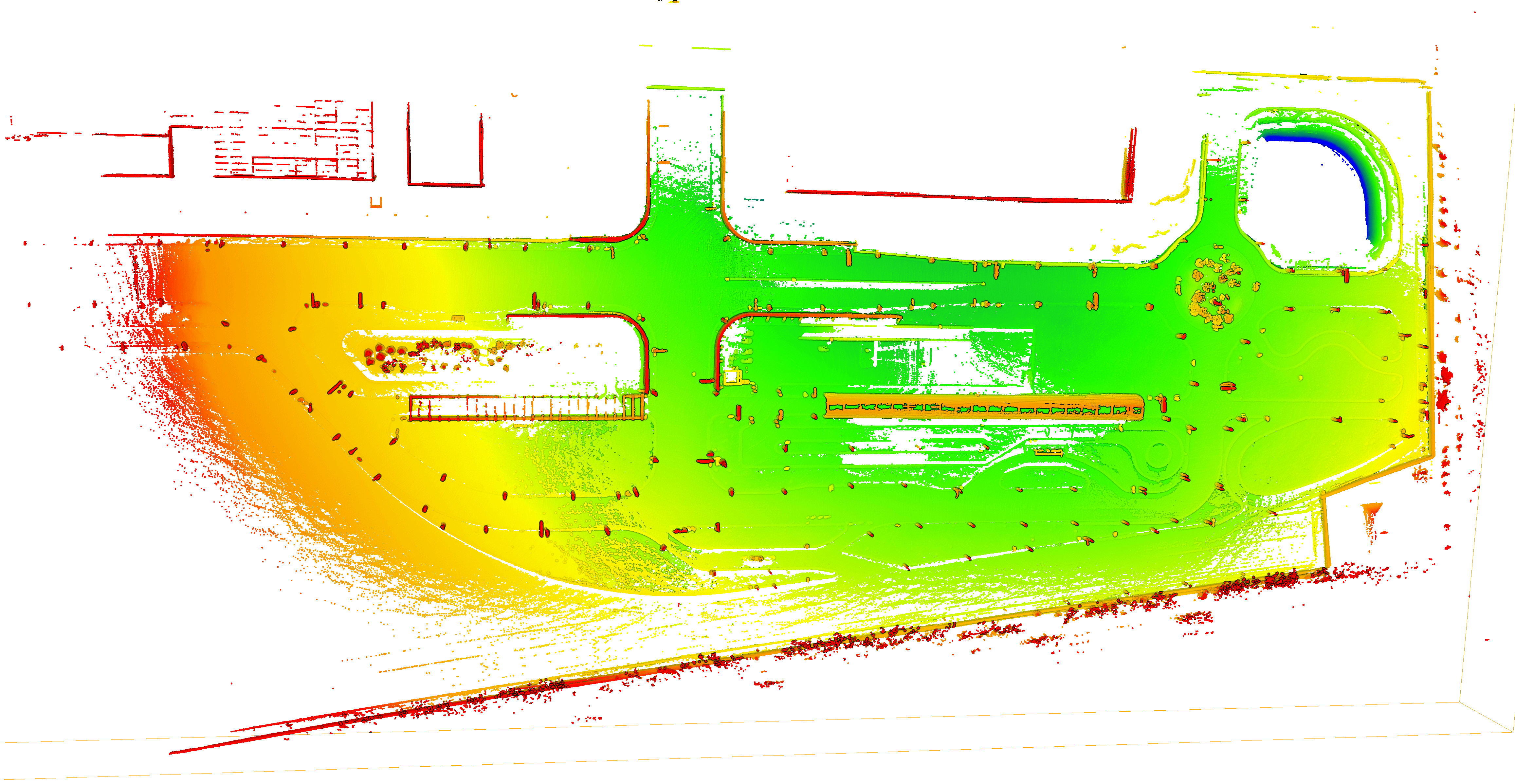}
	  \end{minipage}
  }

  \subfloat[]{
    \begin{minipage}[b]{0.8\textwidth}
	    \centering
	    \includegraphics[width=\textwidth]{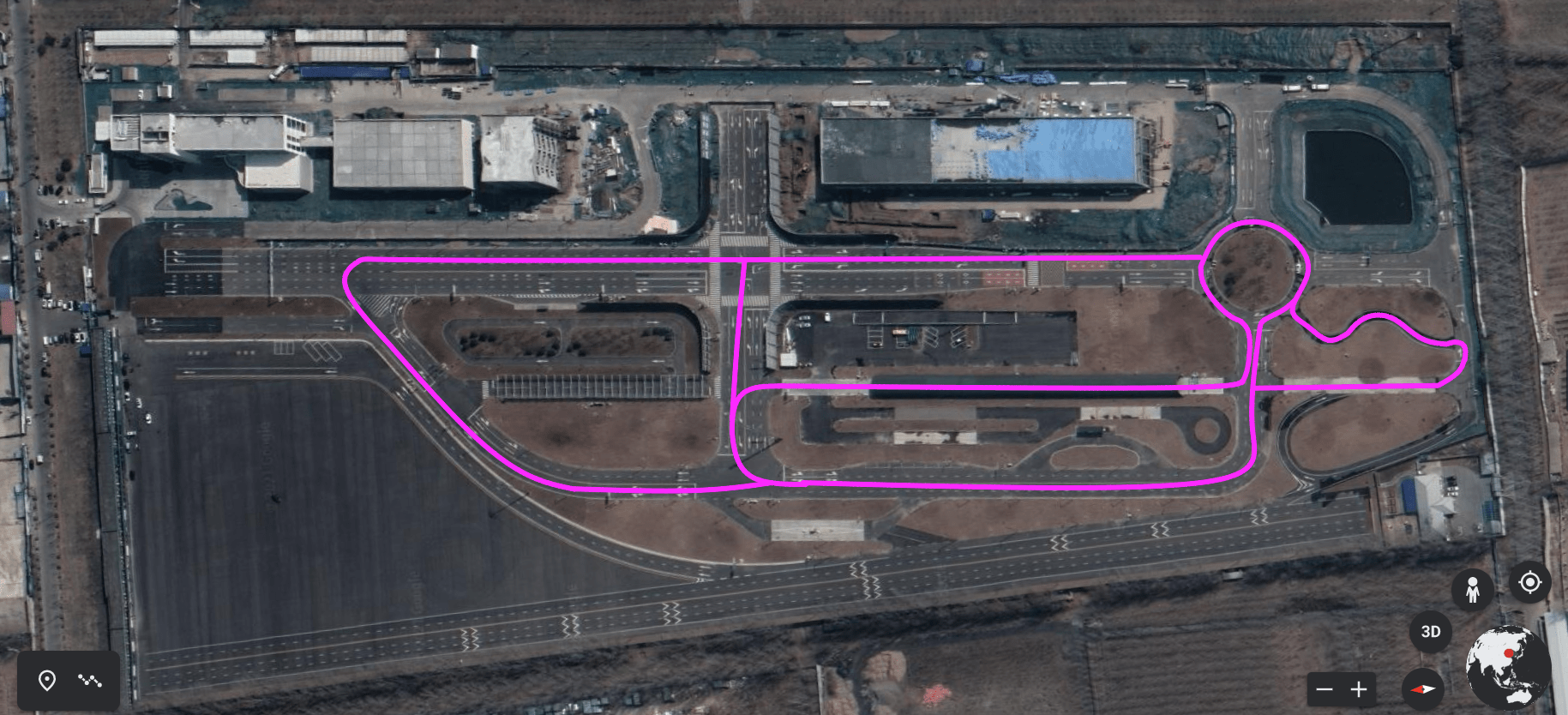}
	  \end{minipage}
  } 
  \caption{InTEn-LOAM's map result on autonomous driving field dataset. (a) the 
  constructed point cloud map, (b) local remote sensing image and estimated trajectory.}\label{FIG:18}
\end{figure}

In addition, we constructed the complete point cloud map for the test field using InTEn-LOAM and compared the result with the local remote sensing image, as shown in Fig.\ref{FIG:18}. It can be seen that the consistency between the constructed point cloud map and regional remote sensing image is good, qualitatively reflecting that the proposed InTEn-LOAM has excellent localization and mapping capability without error accumulation in the around $2km$ long exploration journey.

\section{Conclusions} \label{Sect:5}
In this work, we present a LiDAR-only odometry and mapping solution named InTEn-LOAM to cope with some challenging issues, i.e., dynamic environments, intensity channel incorporation. A temporal-based dynamic removal method and a novel intensity-based scan registration approach are proposed, and both of them are utilized to improve the performance of LOAM. The proposed system is evaluated on both simulated and real-world datasets. Results show that InTEn-LOAM achieves similar or better accuracy in comparison with the state-of-the-art LO solutions in normal environments, and outperforms them in challenging scenarios, such as long straight tunnel. Since the LiDAR-only method cannot adapt to aggressive motion, our future work involves developing a IMU/LiDAR tightly coupled method to escalate the robustness of motion estimation.

\section{Acknowledgement} 


\section{References}\label{REF}

\bibliography{ref}

\end{document}